\documentclass[twoside,11pt,backref=page]{article}
\newcommand{\isAnonymous}{0}
\newcommand{\isPreprint}{1}
\newcommand{\nonAnonymousText}[1]{%
\if\isAnonymous1
\else
	#1
\fi%
}
\newcommand{\ifPreprintElse}[2]{%
\if\isPreprint1
    #1
\else
	#2
\fi%
}

\usepackage{blindtext}

\usepackage{amsthm}
\ifPreprintElse{%
\usepackage[preprint]{jmlr2e}%
}{%
\usepackage{jmlr2e}%
}

\usepackage{lastpage}
\ifPreprintElse{}{%
\jmlrheading{23}{2022}{1-\pageref{LastPage}}{10/22; Revised -/-}{-/-}{21-0000}{Jakob Heiss, Josef Teichmann and Hanna Wutte}%
}

\ShortHeadings{How Infinitely Wide Neural Networks Can Benefit from Multi-task Learning}{Heiss, Teichmann and Wutte}
\firstpageno{1}

\nonAnonymousText{\hypersetup{
pdftitle={How Infinitely Wide Neural Networks Can Benefit from Multi-task Learning -- an Exact Macroscopic Characterization},
    pdfauthor={Jakob Heiss, Hanna Wutte, Josef Teichmann},
    pdfkeywords={Multi-task learning, Machine Learning, Machine learning (artificial intelligence), Machine Learning (stat.ML), Neural Networks, Deep Learning, Deep ReLU neural networks, Deep neural networks, Bayesian Neural Networks, Bayesian neural network, artificial intelligence, machine learning, Artificial neural networks, Regularization, Tikhonov regularization, Generalization, Overparametrized Neural Networks, Large width limit, Regularization, Neural Network Theory, representation learning, feature learning, L2 regularization, weight decay},
}}
\begin{document}

\title{How Infinitely Wide Neural Networks\\ Can Benefit from Multi-task Learning\\ -- an Exact Macroscopic Characterization}

\author{\name Jakob Heiss\,\thanks{Equal contribution.}%
\email jakob.heiss@math.ethz.ch \\
       \addr Department of Mathematics\\
       ETH Zurich
       \AND
       \name  Josef Teichmann \email jteichma@math.ethz.ch \\
       \addr Department of Mathematics\\
       ETH Zurich
       \AND
       \name Hanna Wutte\,$^*$ \email hanna.wutte@math.ethz.ch\\
              \addr Department of Mathematics\\
       ETH Zurich
       }

\editor{My editor}

\maketitle

\begin{abstract}%
In practice, multi-task learning (through learning features shared among tasks) is an essential property of deep neural networks (NNs).
While infinite-width limits of NNs can provide good intuition for their generalization behavior,
the well-known infinite-width limits of NNs in the literature (e.g., neural tangent kernels) assume specific settings in which wide ReLU-NNs behave like shallow Gaussian Processes with a fixed kernel. Consequently, in such settings, these NNs lose their ability to benefit from multi-task learning in the infinite-width limit.
In contrast, we prove that optimizing wide ReLU neural networks with at least one hidden layer using $\ell^2$-regularization on the parameters promotes multi-task learning due to representation-learning -- also in the limiting regime where the network width tends to infinity.
We present an exact quantitative characterization of this infinite width limit in an appropriate function space that neatly describes multi-task learning.

\end{abstract}


\section{Introduction}

One key difference of deep learning models, such as deep \textit{neural networks} (NNs), in contrast to shallow learning models, such as \textit{Gaussian Processes} (GPs)\footnote{Within this paper \enquote{GPs} always refer to shallow GPs where a prior GP has a fixed kernel that is not learned from the data. 
}, is that deep learning methods are capable of benefiting from multi-task learning.
Standard deep NNs with multi-dimensional output share the same weights in the hidden layers, and only the last layer contains different weights for different outputs (or tasks).
This shared representation learned by the hidden layers (representation learning, feature learning or metric learning) can be seen as the source for multi-task learning and transfer learning \citep{MultitaskCaruana1997}.
In various applications, training with high-dimensional output has outperformed separate training of the individuals tasks \citep{MultitaskCaruana1997,ruder2017overviewMultiTask,fifty2021efficiently,tran2021facebook,aribandi2021ext5ExtremeMultiTask}. Especially if the available data for some tasks is limited, these tasks greatly benefit from the other tasks in terms of improved generalization.

Due to the success of deep NNs, there is high interest in studying their generalization behavior (macroscopically, i.e., the inductive bias in function space).
In the case of infinite-width neural networks \cite{jacot2018neural,Neal1996,lee2018deep} have shown that in specific settings, deep NNs are equivalent to shallow GPs. Those however cannot benefit from multi-task learning%
, since their outputs are completely independent.
The authors are aware of this problem and \citet{Neal1996} suggests to define a specific prior for the weights to circumvent it. \citet{roberts2021principles,YaidaBlog2021FirstPrinciples} in contrast suggest to use infinite depth NNs, with a fixed ratio of width and depth to circumvent this problem. %

In contrast, we prove that ReLU-neural networks perfectly optimized under $\ell^2$-regularization can already (also for infinite width and finite depth) benefit from multi-task learning due to representation learning. We deduce this ability by giving an exact quantitative macroscopic characterization of the (generalization) behavior of wide $\ell^2$-regularized NNs.

\subsection{Related Work}
In concurrent work, \citet{YangFeatureLearningInfiniteWidth} show that in specific settings (different from those studied in \citet{jacot2018neural}) gradient descent enables feature learning also in the infinite width limit. This strongly suggests that in these settings, NNs can benefit from multi-task learning. However, \citet{YangFeatureLearningInfiniteWidth} do not provide an easy-to-interpret characterization of the inductive bias in function space for those settings where feature learning is possible \citep[Remark~3.11]{YangFeatureLearningInfiniteWidth}.

\citet{Chizat2020ImplicitBias,NeyshaburImplicitReg2014arXiv1412.6614N,ongie2019function,savarese2019infinite} also provide infinite-width limits of NNs, that do not share the generalization behavior of Gaussian Processes. However, they do not discuss multi-task learning and they do not consider NNs with multiple hidden layers (with nonlinear activation function). Moreover, none of these papers provide a precise theorem showing that the set of finite width networks minimizing the $\ell^2$-regularized loss converge to the set of minimizers of the continuous optimization problem in a certain function space equipped with a certain topology in the limit width to infinity.

\citet{Neal1996,jacot2018neural,implReg1,williams2019gradient,lee2018deep} study settings in which NNs converge to GPs in the infinite width limit. Thus, in these settings the infinitely wide NNs cannot benefit from multi-task learning.

\citet{agrawal2020wide} show that wide \textit{Bayesian Neural Networks} (BNNs) with narrow bottleneck layers are deep Gaussian Processes. However, these deep GPs only deviate from shallow GPs due to narrow bottleneck layers, while our $\ell_2$-regularized wide NNs do not need any finite bottleneck layer in order to deviate from shallow GPs or in order to benefit from multi-task learning (see \Cref{prop:MultiTask}).

Independently and concurrently, \citet{parhi2022kinds} study the macroscopic behavior of deep neural networks with regularization similar to $\ell_2$-regularization. In contrast to their work, we derive an exact characterization of the infinite width-limit of $\ell_2$-regularized NNs (with and without regularization of the biases).  For their choice of regularization on parameter space, they obtain in \citep[Therorem~3.2 and Corollary~4.3]{parhi2022kinds} a similar regularization functional on function space as we do in \Cref{thm:PFunc}. However, their slightly different regularization prevents their model from benefiting from multi-task learning in the case of one hidden layer (as could be proven analogously to \Cref{prop:L1l1NoMultiTask}). 
We think that multi-task learning is one of the most exciting properties of our inductive bias, especially compared to infinite wide Gaussian BNNs. This is a main focus of our paper, which is not treated in  \citet{parhi2022kinds}.
Moreover, we employ different proving techniques and provide different perspectives on the function space characterization:
while their infinite width limit contains fractional derivatives and Radon transforms, our formulation of the infinite width limit is formulated as a natural extension of generalized additive models as we explain on \cpageref{itm:GAMInfiniteDirecteions}.

\subsection{Contribution}
\begin{itemize}
    \item  We give an exact characterization of the (out-of-sample) behavior of wide $\ell^2$-regularized deep ReLU-NNs in function space with arbitrary input and output dimensions. This result also holds for NNs with more than one hidden layer. (\Cref{sec:MainTheorem})
    \item We show that wide $\ell^2$-regularized NNs can benefit from multi-task learning. (\Cref{sec:DeepVsShallow})
    \item We explain a paradoxical contrast between infinitely wide $\ell^2$-regularized NNs and infinitely wide Gaussian Bayesian Neural Networks (BNNs):
    While for finite width, the optimal $\ell^2$-regularized NN is exactly the max a posteriori (MAP) on the parameter space of a Gaussian BNN, this does not hold anymore in the infinite width limit on function space (\Cref{subsec:BNNs}). %
\end{itemize}

\section{Setting and Notation}\label{sec:Setting}
We denote fully connected, shallow NNs (i.e., NNs with one hidden layer) with ReLU activation
as a \enquote{stack}. A fully connected, deep, stacked neural network~$\NN_\theta$ is then given as concatenation of stacks and an element-wise activation function $\sigb$ (not necessarily ReLU as for the hidden layers of a stack),
\begin{equation}\label{def:NN}
    \NN_\theta:=%
    \link^{-1}\circ\NNj{\nStacks}\circ
    \sigb\circ\NNj{\nStacks-1}\circ%
    \dots
    \circ\sigb\circ\NNj{1},
\end{equation}
where $\link^{-1}$ is a Lipschitz continuous function (e.g. identity or soft-max which corresponds to the inverse of a link function in classical statistics) and $\sigb$ is any Lipschitz continuous activation function.
Throughout the paper, we focus on stacks as in \Cref{def:deepNN}; see \Cref{sec:different architectures} for results on different stacks.
\begin{definition}[Deep Stacked Neural Network]\label{def:deepNN}
	A \textit{deep stacked neural network} is defined as in \meqref{def:NN} with stacks $\NNj{j}:\,\R^{d_{j-1}}\to \R^{d_j}$ s.t.
	\begin{equation}\label{eq:RSN}
	\faxd[d_{j-1}] \quad : \qquad \NNj{j}(x)=%
	\sum_{k=1}^{n_j}w_k^{{(j)}}\,\relu[b_{k}^{{(j)}}+\langle v_{k}^{{(j)}}, x\rangle] + c^{{(j)}} 
	\mycomma
	\end{equation}
	with
	\begin{itemize}
		\item\label{itm:as:ReLU} number of hidden neurons~$n_j\in\N$ in the $j$-th stack, not necessarily equal dimensions~$\din=d_0,\dots,d_j,\dots,d_\nStacks=\dout\in\N$ that we call \emph{bottleneck dimensions} and \ReLU{} activation function
		(we collect these numbers in $n=(n_j)_{j\in\fromto{\nStacks}}$ and $d=(d_j)_{j\in\fromto[0]{\nStacks}}$)
	    \item weights~$v_k^{{(j)}} \in\R^{d_{j-1}}$, $w_k^{{(j)}}\in\R^{d_j}$, $k=1,\dots, n_j$ and
		\item biases $c^{{(j)}}\in\R^{d_j}$, $b_k^{{(j)}}\in\R$, $k=1,\dots, n_j$.
		\item Weights and biases are collected in $\theta=\left(\theta^{(j)}\right)_{j\in\fromto{\nStacks}}$ with
		    \[
		\theta^{(j)} :=(v^{{(j)}},b^{{(j)}},w^{{(j)}},c^{{(j)}})\in\Theta^{(j)}:=\R^{n_j\times d_{j-1}}\times\R^{n_j}\times\R^{d_j\times n_j}\times\R^{d_j}.
		    \]
		    \item All network parameters are then collected in
		\[\Theta:=\Thetand:=\Theta^{(1)}\times\dots\times\Theta^{(\nStacks)}.\]
		\end{itemize}
	
\end{definition}

\begin{figure}[htbp]
\centering
\tikzset{every picture/.style={line width=0.75pt}} %

\begin{tikzpicture}[x=0.75pt,y=0.75pt,yscale=-0.8,xscale=1]
\draw   (73.05,219.55) -- (108.15,219.55) -- (108.15,318) -- (73.05,318) -- cycle ;
\draw   (138.96,72) -- (174.49,72) -- (174.49,410.09) -- (138.96,410.09) -- cycle ;
\draw   (209.5,259) -- (246.31,259) -- (246.31,305) -- (209.5,305) -- cycle ;
\draw   (280.42,72) -- (316.33,72) -- (316.33,411.09) -- (280.42,411.09) -- cycle ;
\draw   (353.14,241.36) -- (389.05,241.36) -- (389.05,319.36) -- (353.14,319.36) -- cycle ;
\draw    (108.15,219.55) -- (138.96,72) ;
\draw    (108.15,318) -- (138.96,410.09) ;
\draw    (174.49,72) -- (209.5,259) ;
\draw    (246.31,259) -- (280.42,72) ;
\draw    (174.49,410.09) -- (209.5,305) ;
\draw    (246.31,305) -- (280.42,411.09) ;
\draw    (316.33,72) -- (353.14,241.36) ;
\draw    (316.33,411.09) -- (353.14,319.36) ;
\draw   (423.42,73) -- (459.33,73) -- (459.33,408.09) -- (423.42,408.09) -- cycle ;
\draw   (496.14,261) -- (532.05,261) -- (532.05,316.45) -- (496.14,316.45) -- cycle ;
\draw    (389.05,241.36) -- (423.42,73) ;
\draw    (389.05,319.36) -- (423.42,408.09) ;
\draw    (459.33,73) -- (496.14,261) ;
\draw    (459.33,408.09) -- (496.14,316.45) ;

\draw (140.42,239) node [anchor=north west][inner sep=0.75pt]  [font=\scriptsize] [align=center] {ReLU\\
($n_1$)};
\draw (281.98,241) node [anchor=north west][inner sep=0.75pt]  [font=\scriptsize] [align=center] {ReLU\\
($n_2$)};
\draw (209.24,273) node [anchor=north west][inner sep=0.75pt]  [font=\scriptsize] [align=left] {$\quad\, \sigb$\\
\ \ ($d_1$)};
\draw (351.83,273) node [anchor=north west][inner sep=0.75pt]  [font=\scriptsize] [align=left] {$\quad\ \sigb$\\
\ \ ($d_2$)};
\draw (76.07,261) node [anchor=north west][inner sep=0.75pt]  [font=\scriptsize] [align=center] {Input\\
($d_0$)
};
\draw (424.98,238) node [anchor=north west][inner sep=0.75pt]  [font=\scriptsize] [align=center] {ReLU\\
($n_3$)};
\draw (494.83,270) node [anchor=north west][inner sep=0.75pt]  [font=\scriptsize] [align=center] {\ Output\quad\\
$\link^{-1}$\\
($d_3$)};

\end{tikzpicture} \caption{Schematic representation of a Deep Stacked NN from \Cref{def:deepNN} with $\nStacks=3$ stacks, where we show the activation functions and the dimensions of the layers of this feed forward NN.}
\label{fig:architectureDeepStackBasic}
\end{figure}
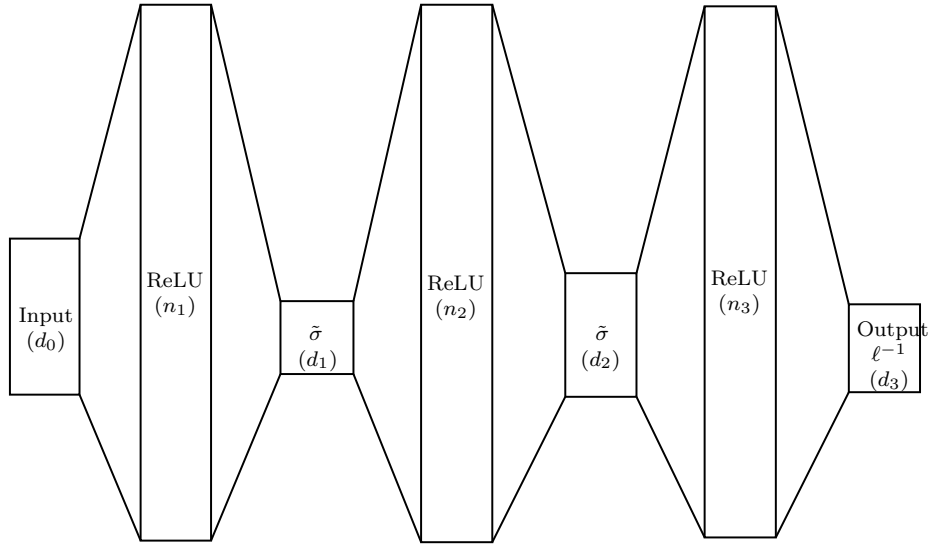

\noindent Within this paper, for simplicity, we assume that NNs are trained to perform $\dout$ one-dimensional regression tasks with quadratic training loss, where the $k\textsuperscript{th}$ task learns from a subset of training data with indices $I_k\subset\set{1,\dots,N}=I_1\dot{\cup}\dots\dot{\cup}I_{\dout}$, i.e.,
\begin{subequations}\label{eq:Ltr}\begin{align}
\Ltrb{\hat{f}}&:=\sum_{k=1}^{\dout}\Ltr_k(\hat{f}),\\
\Ltr_k(\hat{f})&:=\sum_{i\in I_k}(\hat{f}_k(\xtr_i)-\ytr_{i})^2.\label{eq:Lk}%
\end{align}
\end{subequations}
\begin{remark}
Our main \Cref{thm:PFunc} also holds true for other losses of the form $L(\hat{f})=\sum_{i=1}^{N}l_i\left( \hat{f}(\xtr_i),\ytr_i\right)$, with continuous loss-functions $l_i:\ell^{-1}(\R^\dout)\times\ell^{-1}(\R^\dout)\to\R$ (e.g. classification). While in \meqref{eq:Lk}, it is $\ytr_i\in\R$, within this remark, we have $\ytr_i\in\ell^{-1}(\R^\dout)$. Moreover, $l_i$ can ignore different dimensions for different indices $i$, depending on which output-labels are available for certain tasks at certain input points $\xtr_i$. 
\end{remark}

\subsection{Definition of Multi-task Learning}
The idea of multi-task learning \citep{MultitaskCaruana1997} is that when training a model to perform multiple tasks simultaneously, the joint training induces a transfer of knowledge among the model's outputs to improve\footnote{More discussion on the benefits of multi-task learning can be found in \Cref{sec:DiscussBenefitsMultiTask}.} each others generalization.

\begin{definition}
We define a model $\mathcal{M}$ as a map that gets as input a loss functional $\Ltr$ in the form of \meqref{eq:Ltr} (including training data) and gives as output a set of functions $\mathcal{M}(\Ltr)$. (E.g. $\mathcal{M}(\Ltr):=\argmin_f \left(\Ltr(f)+\lambda P(f)\right)$ for some regularization functional $P$)  
\end{definition}

\begin{definition}\label{def:NoMultiTask}
    We say a model $\mathcal{M}$ \emph{cannot benefit from multi-task learning} if for every $k$ and for every loss $\Ltr$ in the form of \meqref{eq:Ltr}, 
    \begin{align*}
        (\mathcal{M}(\Ltr))_k=(\mathcal{M}(\Ltr_k))_k, 
    \end{align*}
    where for any set $A$, we define $(A)_k:=\set{a_k : a\in A}$.
\end{definition}

In simple words, \Cref{def:NoMultiTask} says that a model \emph{cannot benefit from multi-task learning}, if removing all the data for other tasks than task~$k$ from the training set does not influence the behavior of the model on the task~$k$.

\section[Characterizing the Learned Function]{Characterizing the Learned Function}\label{sec:MainTheorem}
In this paper we focus on understanding deep stacked ReLU-NNs (\Cref{def:deepNN}) optimized with $\ell^2$-regularization (weight decay), i.e.,
\begin{equation}\label{eq:optimalParameters}
    \NN_{\theta^{*,\lw}} \text{ with }
    \theta^{*,\lw}\in\argmin_\theta \left( \Ltrb{\NN_{\theta}} + \lw\twonorm[\theta]^2\right),
\end{equation}
for the case of large numbers of $n_j $ for $ j=1\ldots,\nStacks$. First, we formulate a characterization in function space for networks where the width of every second layer stays finite (i.e.,~$d_j\in\N$ stay fixed) and only the width of the other layers goes to infinity (i.e.,~$n\to\infty$). We use $\lim_{n\to\infty}$ as a shorthand notation for 
$\lim_{(n_1,\dots,n_{\nStacks})\to(\infty,\dots,\infty)}$. Then, we proceed to derive the theory for the case where the width of every hidden layer goes to infinity (i.e.,~ $n_j,d_j\to\infty$).

The goal of this section is to formulate $\lim_{n\to\infty}\NN_{\theta^{*,\lw}}$ as the solution of an optimization problem of the form
\begin{equation}\label{eq:optimalFunction}
    f^{*,\lambda}\in\argmin_{f\in\cF} \left(\Ltrb{f}+\lambda\Pfunc(f)\right)
\end{equation}
for an appropriate function space $\mathcal{F}$ and regularization functional $P:\mathcal{F}\to\Rpz$ to better understand how the solution of \meqref{eq:optimalParameters} behaves in function space  in the case of many hidden neurons. %

The form of the regularization functional $P$ depends on the architecture of the network. As we will prove in \Cref{thm:PFunc}, for a wide deep stacked NN as in \Cref{def:deepNN} the corresponding functional is given as
\begin{equation}\label{eq:P}
P(f) = \min_{\substack{%
(h_1,\dots, h_{\nStacks})\text{, s.t. }\\ f=\link^{-1}\circ h_{\nStacks}\circ\dots\circ \sigb\circ h_1%
}} \left(\Pfunc_1(h_1) + \Pfunc_2(h_2) + \dots + \Pfunc_\nStacks(h_\nStacks) \right),
\end{equation}
where $P_j$ (and $\T$) are given with slight abuse of notation (treating distributions as if they were functions)\footnote{To be mathematically precise, the \enquote{function} $\varphi: s\mapsto\varphi_s$ does not have to be a classical function, but can also be a distribution. First derivatives are understood in a weak sense and second derivatives are distributional. See \Cref{sec:EquivalentPFuncs} for different formulations of $P_j$ in precise mathematical notation.} as%

\hypertarget{eq:PgSm}{\begin{equation}\label{eq:Pj}
	\Pfunc_j(h_j):=  \min_{\substack{\varphi\in\T,\ c\in\R^{d_j}\text{ s.t.}\\  h_j=\int_{\Sd[d_{j-1}-1]}\psr[\langle s,\cdot\rangle]{s}\, ds+c}} \left(
	2\int_{\Sd[d_{j-1}-1]}\int_{\R} \frac{\twonorm[ {\psr{s}}^{''} ]}{g(r)} \dx[r]\dx[s] +2\rho(\twonorm[c])		\right),%
\end{equation}}
where

\hypertarget{eq:T}{\begin{align*}
	\T:=\bigg\{ \varphi=(\varphi_s)_{s\in\Sd[d_{j-1}-1]}  &\bigg|\, \forall s\in\Sd[d_{j-1}-1] :\varphi_s:\R\to\R^{d_j},\lim_{r\to -\infty} \psr{s}=0 = \lim_{r\to -\infty} \frac{\partial}{\partial r}\psr{s}%
	\bigg\}
\end{align*}}
handles a boundary condition, $\Sd$ denotes the $(d-1)$-dimensional unit sphere, %
$g(r)=\frac{1}{\sqrt{r^2+1}}$ denotes a weight function, and where $\rho(r)=\begin{cases}r^2/2&\text{, if }|r|\leq1\\
|r|-1/2&\text{, else}
\end{cases}$ is the Huber-loss.

Neither the solutions $\NN_{\theta^{*,\lw}}$ to \cref{eq:optimalParameters} nor the solutions $f^{*,\lambda}$ to \cref{eq:optimalFunction} have to be unique. But even in cases where they are not unique, the set of solutions to \cref{eq:optimalParameters} converges to the set of solutions to \cref{eq:optimalFunction} as $n\to\infty$.

\begin{theorem}\label{thm:PFunc}
Using the definitions from \Cref{sec:Setting,sec:MainTheorem}, it holds that for a sufficiently large\footnote{\label{footnote:boundsonneurons}In \Cref{sec:RequiredNumberofNeurons} we discuss explicit bounds on how many neurons are sufficient for the result to hold. 
}\textsuperscript{,}%
\footnote{\label{footnote:ndependence}%
Note that $\NN_{\theta}$ depends on $n$ since it refers to a network with $n_j$ neurons in the corresponding layers. Moreover, $n\geq\tilde{n}$ is always understood component-wise. By \enquote{sufficiently large $n$}, we mean that every $n_j$ has to be sufficiently large.}
number of neurons $n$ every solution $\NN_{\theta^{*,\lw}}$ to \cref{eq:optimalParameters} is a solution to \cref{eq:optimalFunction} too, i.e.,
\begin{align}\label{eq:Main_Theorem_Subset}
    \NN_{\theta^{*,\lw}} \in \argmin_{f\in \cF} \left(\Ltrb{f}+\lambda\Pfunc(f)\right).
\end{align}
Furthermore, it holds that for every compact $ K\subset\R^{\din},\ \forall \epsilon\in\Rp:$
\begin{multline}\label{eq:continousSolutionAlmostSubsetOfDiscrete}
    \forall f^{*,\lambda} \in \argmin_{f\in \cF} \left(\Ltrb{f}+\lambda\Pfunc(f)\right) : \exists \tilde{n}\in\N^{\nStacks}:
    \forall n\geq\tilde{n}:\\
    \exists \theta^{*,\lw}\in\argmin_{\theta\in\Thetand} \left( \Ltrb{\NN_{\theta}} + \lw\twonorm[\theta]^2\right):
     \sup_{x\in K}\|{f^{*,\lambda}(x)-\NN_{\theta^{*,\lw}}}(x)\|_{\infty} <\epsilon.
\end{multline}
\end{theorem}
\begin{proof}
We prove the theorem in \Cref{subsec:ProofMainTheorem}, while proving all necessary lemmas in \Cref{sec:proofs}. Some basic intuition for the proof can be obtained from \Cref{le:l2equivl1g} for finite-width networks, from \Cref{le:mainThrmFirstStatement} for \eqref{eq:Main_Theorem_Subset}, and from \Cref{le:mainThrmSecondStatement} for \eqref{eq:continousSolutionAlmostSubsetOfDiscrete}.
\end{proof}
Moreover, we can describe the limiting regime $d_j\to\infty$ for $j\in\fromto{\nStacks-1}$, i.e., the regime of letting the width of \emph{all} hidden layers go to infinity.

\begin{corollary}\label{cor:wide_dj}
Let $\sigb$ be ReLU or linear. For every number of training data points $N$, there exist $n^*\in\N^{\nStacks}$ and $d^*\in\N^{\nStacks-1}$,%
\textsuperscript{\ref{footnote:boundsonneurons}} such that in the limit where all layers tend to infinity,
all solutions are characterized as\footnote{Note that $\NN_{\theta}$ depends on $d$ and $n$ since it refers to a network with $d_j$ and $n_j$ neurons in the corresponding layers. Moreover, $d\geq d^*$ is always understood component-wise, i.e., $\forall j\in\{1,\dots, \nStacks-1\}:d_j\geq d_j^*$, while $d_0=\din$ and $d_\nStacks=\dout$ are considered constant throughout the paper.}
\begin{subequations}
\begin{align}\mathcal{S}&:=\overline{\lim_{(n,d)\to\infty}\left\{\NN_{\theta^{*,\lw}}:\theta^{*,\lw}\in\argmin_{\theta\in\Thetand} \left( \Ltrb{\NN_{\theta}} + \lw\twonorm[\theta]^2\right)\right\}}\label{subeq:limitAllLayersWideNetworks}\\
&=
\overline{\bigcup_{(n,d)\geq(n^*,d^*)}\left\{\NN_{\theta^{*,\lw}}:\theta^{*,\lw}\in\argmin_{\theta\in\Thetand} \left( \Ltrb{\NN_{\theta}} + \lw\twonorm[\theta]^2\right)\right\}}\label{subeq:unionAllLayersWideNetworks}\\
&=
\overline{\lim_{d\to\infty}\argmin_{f\in \cF} \left(\Ltrb{f}+\lambda\Pfunc_{(d)}(f)\right)}\label{subeq:limitAllLayersWidePfunc}
\\
&=
\overline{\bigcup_{d\geq d^*}\argmin_{f\in \cF} \left(\Ltrb{f}+\lambda\Pfunc_{(d)}(f)\right)}\label{subeq:unionAllLayersWidePfunc},
\end{align}
\end{subequations}
(where $\Pfunc_{(d)}$ denotes $\Pfunc$ with bottleneck-dimensions $d_j$,\footnote{$\Pfunc$ always depends on the bottleneck-dimensions $d$, but by writing $\Pfunc_{(d)}$ we make the dependency on $d$ more explicit at this point.} and the closure $\overline{(\cdot)}$ is understood with respect to the topology of locally uniform convergence in function space.)

Moreover, there exists a solution in $f^{*,\lambda}\in\mathcal{S}$ that can be represented as a neural network $\NN_{\theta^{*,\lw}}=f^{*,\lambda}$ with dimensions $n^*$ and $d^*$. This network satisfies
\[
    \theta^{*,\lw}\in\argmin_{\theta\in\Thetandstar} \left( \Ltrb{\NN_{\theta}} + \lw\twonorm[\theta]^2\right)%
\]
and 
\[
    \NN_{\theta^{*,\lw}}\in\argmin_{f\in \cF} \left(\Ltrb{f}+\lambda\Pfunc_{(d^*)}(f)\right)%
.\]

\end{corollary}

\begin{proof}
See \Cref{subsec:ProofOfCorollary} for the proof and different points of view on \Cref{cor:wide_dj}.
\end{proof}

\begin{remark}[No regularization on biases]
If one does not regularize the biases but only the weights, one obtains given the same abuse of notation as above (treating distributions as if they were functions)
\hypertarget{eq:PgSm_no_bias_reg}{\begin{equation}\label{eq:Pj_no_bias_reg}
	\Pfunc_j(h_j):=  \min_{\substack{\varphi\in\T,\ c\in\R^{d_j}\text{ s.t.}\\  h_j=\int_{\Sd[d_{j-1}-1]}\psr[\langle s,\cdot\rangle]{s}\, ds+c}} \left(
	2\int_{\Sd[d_{j-1}-1]}\int_{\R} \twonorm[ {\psr{s}}^{''} ] \dx[r]\dx[s]	\right),
\end{equation}}
where $\T$ simplifies to%
\begin{align*}
	\T:=\bigg\{ \varphi=(\varphi_s)_{s\in\Sd[d_{j-1}-1]}  &\bigg|\, \forall s\in\Sd[d_{j-1}-1] :\varphi_s:\R\to\R^{d_j}, \lim_{r\to -\infty} \frac{\partial}{\partial r}\psr{s}=0 %
	\bigg\}.
\end{align*}

In this case, and if $d_j=1$, \citet{savarese2019infinite} provides a simpler reformulation of $\Pfunc_j$ from \cref{eq:Pj_no_bias_reg}, i.e, given the same abuse of notation as above (treating distributions as if they were functions)
{\tiny\begin{equation}\label{eq:Pj_no_bias_reg_simple}
	\Pfunc_j(h_j):=  \min_{\substack{\varphi=(\varphi_s)_{s\in\Sd[d_{j-1}-1]} ,\ c\in\R^{d_j}\text{ s.t.}\\  h_j=\int_{\Sd[d_{j-1}-1]}\psr[\langle s,\cdot\rangle]{s}\, ds+c}} \left(
	2\int_{\Sd[d_{j-1}-1]}\max\left(\int_{\R} \left| {\psr{s}}^{''} \right| \dx[r],\left|\lim_{r\to -\infty} {\psr{s}}^{'} + \lim_{r\to +\infty} {\psr{s}}^{'}\right|\right)\dx[s]	\right).
\end{equation}}%
\Cref{eq:Pj_no_bias_reg_simple} can be particularly intuitively interpreted as a generalized additive model (GAM), where
\begin{enumerate}
    \item\label{itm:GAMInfiniteDirecteions} %
    instead of only using the coordinate directions ($e_1,\dots,e_{d_j}$), all possible directions $s\in\Sd[d_{j-1}-1]$ are used,
\item%
instead of the typical smoothing spline regularization $\int_{\R} \twonorm[ {\psr{s}}^{''} ]^2 \dx[r]$, an $L^1$-regularization $\int_{\R} \twonorm[ {\psr{s}}^{''} ] \dx[r]\,\left(=\int_{\R} \left| {\psr{s}}^{''}\right| \dx[r] \text{ for } d_j=1\right)$ is applied and
\item%
the first derivative additionally gets regularized.
\end{enumerate}
\Cref{eq:Pj_no_bias_reg} is a natural extension of \Cref{eq:Pj_no_bias_reg_simple} to higher dimensional outputs.
\Cref{eq:Pj} imposes a qualitatively similar regularization on the learned function as \cref{eq:Pj_no_bias_reg}, with the exception that also the zeroth derivative gets slightly regularized and the second derivative gets penalized more strongly far away from the origin $0\in\R^\din$ than close to it. %
However, even if one does not regularize the biases explicitly, the obtained functions will in practice have some qualitative aspects of \cref{eq:Pj}, since gradient descent initialized close to zero can implicitly regularize the bias too. 

\end{remark}

\begin{remark}[Random hidden layers ~\cite{implReg1}]\label{rem:RandomHiddenLayers}%
Comparing \cref{eq:Pj} to \cite{implReg1}, where the first-layer weights and biases $v$ and $b$ are not trained but chosen randomly, one can see that the main difference is that the integrand $\frac{\twonorm[ {\psr{s}}^{''} ]}{g(r)}$ replaced  the integrand $\frac{\twonorm[ {\psr{s}}^{''} ]^2}{g(r)}$ (lifted to multi-dimensional in- and output), i.e., the integrand in  \cref{eq:Pj} takes the square root of the numerator and also the weighting function $g$ does not depend on the distribution of $v$ and $b$ anymore (since $v$ and $b$ are trainable now too). If one still sampled $v^{{(j)}}$ and $b^{{(j)}}$ randomly without training them, one could plug in the corresponding regularizing functional %
for $\Pfunc_j$ in \cref{eq:P}. %
\end{remark}

\section{Discussion of Multi-task Learning}\label{sec:DeepVsShallow}

\subsection{Multi-task Learning for Single Stacks and GPs}

Already for neural networks with a single hidden layer one can see that the regularizing functional from the previous section induces multi-task learning. In this case, we have $\nStacks=1$, and $\Pfunc=\Pfunc_1$ from \cref{eq:Pj_no_bias_reg} if the biases are not regularized. (Analogous arguments also hold for the functional in \meqref{eq:Pj} where biases are regularized as well.)
The square-root in the definition of the Euclidean norm $\twonorm$ that appears in \cref{eq:Pj_no_bias_reg} enables multi-task learning: 
If multiple outputs $f_k$ almost only vary\footnote{We say that $f_k$ \enquote{varies} a lot in a direction $s$, when changing the input $x$ in the direction of $s$  the output $f_k(x)$ changes a lot, possibly very non-linearly with very strong second derivative in this direction $s$. We say that $f_k$ does almost not vary in other directions when changing the input in other directions has little influence on the output, i.e., the output in these other directions is mostly linear and very flat (i.e., it has low first derivative and very low second derivative in these directions).} in a certain direction $s\in \Sd[\din-1]$, i.e., if $\twonorm[ {\psr{s}}^{''} ]$ is larger than for other directions, then, due to the concavity of the square root, adjustments in this second derivative ${\psr{s}}^{''}$ have a much lower effect on $\twonorm[ {\psr{s}}^{''} ]$ for this direction $s$ than for others. 
In other words, the marginal regularization costs\footnote{The marginal costs~$\frac{\partial}{\partial a_i}\twonorm[a]=\frac{a_i}{\twonorm[a]}$ of increasing one component of a vector $a$ are inversely proportional to the euclidean norm of all the components of a vector. E.g., if $\psr{s}^{''}=(0,100,100)^\top$ then replacing it by $(1,100,100)^\top$ increases $\twonorm[\psr{s}^{''}]$ by only $\twonorm[(1,100,100)^\top]-\twonorm[(0,100,100)^\top]\href{https://www.wolframalpha.com/input?i=sqrt\%281\%2B100\%5E2\%2B100\%5E2\%29-sqrt\%28100\%5E2\%2B100\%5E2\%29}{\approx0.0035}$, but if $\psr{s}^{''}=(0,0,0)^\top$, replacing it by $(1,0,0)^\top$ increases $\twonorm[\psr{s}^{''}]$ by $\twonorm[(1,0,0)^\top]-\twonorm[(0,0,0)^\top]=1$.
} for second derivative of any other $\tilde{k}$\textsuperscript{th} component of $\psr{s}$ are much smaller for these directions $s$ than for other directions, because of the strict concavity of the square root-function. 

Consequently, the output $f_{\tilde{k}}$ trained to perform the $\tilde{k}$\textsuperscript{th} task will prefer functions which mainly vary in those directions in which also the other outputs that were trained to perform other tasks vary a lot. In this way, different outputs can learn from each other which directions tend to be more important.
Note that the model $\mathcal{M}(\Ltr)=\argmin_{f\in\cF} \Ltr(f)+\lambda P(f)$ is still universal and thus is also able to learn functions $f$ where different components $f_k$ vary in very different directions \emph{if} there is enough evidence in the data (i.e., in $L$) to do so.

Already for one-dimensional input, the square-root can lead to multi-task learning: If some outputs $f_k$ have stronger second derivative $|f_k^{''}(x)|$ or even kinks at some positions $x$, other outputs $f_{\tilde{k}}$ will also prefer to have stronger second derivative or even kinks at these positions $x$.

In line with \Cref{rem:RandomHiddenLayers} and the following proposition, shallow NNs with random first layer, and where only the terminal layer is trained are not capable of multi-task learning.

\begin{proposition}\label{prop:RSNNoMultiTask}
The model $\mathcal{M}^{L^2}(\Ltr):=\argmin_f \left(\Ltr(f) + \lambda \int\twonorm[f''(x)]^2dx\right)$ is not capable of benefiting from multi-task learning (see \Cref{def:NoMultiTask}). 
\end{proposition}
\begin{proof}\label{proof:prop:RSNNoMultiTask}
The main idea of the proof %
is the following:
Squaring the Euclidean norm cancels the square root that connects the outputs to each other. 
Without the square-root, learning a separate function $f_k$ for each task would result exactly in the same functions $f_k$ as training them all together%
, since
\[
\int\twonorm[f^{''}(x)]^2 \,dx
=\int\sum_{k=1}^{\dout}\left(f_k^{''}(x)\right)^2 \,dx
=\sum_{k=1}^{\dout}\int\left(f_k^{''}(x)\right)^2 \,dx.\]
Therefore,
\[\Ltrb{f} + \lambda\int\twonorm[f^{''}(x)]^2 \,dx
=\sum_{k=1}^{\dout}\left(\Ltr_k(f) + \lambda\int\left(f_k^{''}(x)\right)^2 \,dx\right),
\]
\begin{align*}
\left(\mathcal{M}^{L^2}(\Ltr)\right)_k
&=\left(\argmin_f \Big(\Ltrb{f} + \lambda\int\twonorm[f^{''}(x)]^2 \,dx\Big)\right)_k\\
&=\argmin_{f_k}\left(\sum_{i\in I_k}({f}_k(\xtr_i)-\ytr_{i})^2+ \lambda\int\left(f_k^{''}(x)\right)^2 \,dx\right)
=\left(\mathcal{M}^{L^2}(\Ltr_k)\right)_k%
.
\end{align*}
(Note that all other components of $\mathcal{M}^{L^2}(\Ltr_k)$ are the zero function.)
\end{proof}

By contrast,  this is not the case if the square-root appears in the regularizing functional, since \[\int\twonorm[f^{''}(x)] \,dx
=\int\sqrt[2]{\sum_{k=1}^{\dout}\left(f_k^{''}(x)\right)^2} \,dx\]
is in general not equal to
$\sum_{k=1}^{\dout}\int\left|f_k^{''}(x)\right| \,dx$.
Therefore, we see that already a single hidden layer is sufficient to get the effect of multi-task learning for $\lim_{n\to\infty}\NN_{\theta^{*,\lw}}$ when $\NN_{\theta^{*,\lw}}$ are trained with $\ell^2$-regularization. 

\begin{proposition}\label{prop:MultiTask}
The model $\mathcal{M}^{P}(\Ltr):=\argmin_f \left(\Ltr(f) + \lambda P(f)\right)$, with $\Pfunc=\Pfunc_1$ from \cref{eq:Pj}, induces multi-task learning (without the need for any finite bottleneck layer). 
\end{proposition}

\begin{proof}
Consider $\Ltr$ as in \meqref{eq:Ltr} with $\dout=2$ tasks. Let training data be given as
\begin{align*}
    ((\xtr_{1,1},\xtr_{1,2}),\ytr_1) &= ((2,2),2), & ((\xtr_{2,1},\xtr_{2,2}),\ytr_2) &= ((-2,2),2),\\
    ((\xtr_{3,1},\xtr_{3,2}),\ytr_3) &= ((-2,-2),-2), &((\xtr_{4,1},\xtr_{4,2}),\ytr_4) &= ((2,-2),-2),\\
    ((\xtr_{5,1},\xtr_{5,2}),\ytr_5) &= ((0,-3),-3), &((\xtr_{6,1},\xtr_{6,2}),\ytr_6) &= ((0,3),3),
\end{align*}
for $I_1=\set{1,2,3,4,5,6}$ and
\begin{align*}
    ((\xtr_{7,1},\xtr_{7,2}),\ytr_7) &= ((1,2),1), & ((\xtr_{8,1},\xtr_{8,2}),\ytr_8) &= ((-1,-2),-1),\\
    ((\xtr_{9,1},\xtr_{9,2}),\ytr_9) &= ((2,4),2), & ((\xtr_{10,1},\xtr_{10,2}),\ytr_{10}) &= ((-2,-4),-2),
\end{align*}
for $I_2=\set{7,8,9,10}$. Then, one can calculate the sets of minimizers as
\[\Set{\left(f_1:(x_1,x_2)\mapsto x_2, f_2:(x_1,x_2)\mapsto \frac{x_2}{2}\right)}=\lim_{\lambda\to0+}\argmin_{f\in\cF} \left(\Ltr(f) + \lambda P(f)\right),\]
\[\Set{\left(\widetilde{f}_1\equiv0, \widetilde{f}_2:(x_1,x_2)\mapsto \frac{x_1+2x_2}{5}\right)}=\lim_{\lambda\to0+}\argmin_{f\in\cF} \left(\Ltr_2(f) + \lambda P(f)\right).\]
Therefore, (because of continuity arguments) for small values of $\lambda$,
\begin{align*}
        (\mathcal{M}^{P}(\Ltr))_2\neq(\mathcal{M}^{P}(\Ltr_2))_2. 
    \end{align*}
Intuitively speaking, the second output has learned from the training of the first output that $x_2$ is a more important feature than $x_1$. This is one simple example that is sufficient to prove \Cref{prop:MultiTask} in the sense of \Cref{def:NoMultiTask}.
\end{proof}

Crucially however, the limits corresponding to exactly the same NN-architecture that are discussed in \citet{jacot2018neural,Neal1996,lee2018deep} result in a GP-regression with absolutely no multi-task learning benefits, because of the fixed, data-independent kernel.
The prior GPs considered %
there have completely independent outputs. Consequently, calculating the max a posteriori (MAP)\footnote{The MAP of a GP is always interpreted in a Cameron-Martin sense \citep{CameronMartin10.2307/1969276} in this paper. This is equivalent to taking the point-wise MAP for posterior marginals at every single point.} of all outputs jointly results in exactly the same function as if calculating the MAP for every output separately. Thus, it is mathematically impossible that any transfer of knowledge from one task to another can happen, as we note in the following proposition.

\begin{proposition}\label{prop:GPnoMultiTask}
Let $\mathcal{M}^\text{GP}(\Ltr):=\argmin_f \left(\Ltr(f) + \|f\|_{\text{RKHS}}^2\right)$, where $\|\cdot\|_{\text{RKHS}}$ is the reproducing kernel Hilbert space norm of one of the kernels suggested in \citet{jacot2018neural,Neal1996,lee2018deep} or any other kernel with independent outputs, then $\mathcal{M}^\text{GP}$ cannot benefit from multi-task learning according to \Cref{def:NoMultiTask}. ($\mathcal{M}^\text{GP}(\Ltr)$ corresponds to the MAP with respect to the GP prior by interpreting $\Ltr$ as the log-likelihood modulo a constant, which is also equal to the mean a posteriori.)
\end{proposition}
\begin{proof}
Trivial, because then $\|\cdot\|_{\text{RKHS}}$ splits into a sum such as in the \hyperref[proof:prop:RSNNoMultiTask]{proof} 
of \Cref{prop:RSNNoMultiTask}.
\end{proof}

\subsection{Multi-task Learning for Deep Stacked NNs}
Increasing the number of stacks $\nStacks>1$ further strengthens the multi-task learning effects, because not only does the square-root in each $\Pfunc_j$ enforce multi-task learning for each $h_j$, but all the functions $h_j$ for $j\in\fromto{\nStacks-1}$ are shared among all the outputs as well. Thus, $H:=\sigb\circ h_{\nStacks-1}\circ\dots\circ\sigb\circ h_{1}$ has to be learned to transform inputs $x$ into a vector representation $H(x)$ that allows jointly for all functions $(\link\circ f)_k$ to be nicely representable as $h_{\nStacks,k}\circ H$ such that $\Pfunc_\nStacks(h_{\nStacks})$ is not too large. In \Cref{sec:VisualizingMultiTask}, we visualize multi-task learning of a deep stacked NN on two simple examples.

The multi-task learning behavior enforced by the inductive bias $\Pfunc$ of deep NNs (i.e., $\nStacks>1$) is clearly different from the inductive bias $P_1$ of a shallow neural network (i.e., $\nStacks=1$). Firstly, deep NNs allow to share arbitrarily complicated features among different tasks, and the outputs can approximate any continuous task-specific function in terms of these features (see \Cref{sec:UniversalFeaturesUniversalReadout}). Secondly, deep NNs allow to share different levels of abstractions among different subgroups of tasks (see \Cref{sec:LevelsOfAbstraction}).

\section{Connection to Bayesian Neural Networks (BNNs)}\label{subsec:BNNs}
The solution $\theta^{*,\lw}$ of \cref{eq:optimalParameters} is the max a posteriori (MAP) on parameter space of a Gaussian BNN, when the ratio of Gaussian\footnote{In \cref{subsec:BNNs} we assume that $\Ltr$ is the squared loss as given in \cref{eq:Ltr}, while outside \cref{subsec:BNNs}, all results hold for more general losses too.} \iid{} data noise variance and the variance of the Gaussian \iid{} prior of parameters is $\lw$. 

In this paper we study the limit in function space of $\NN_{\theta^{*,\lw}}$ as the number of hidden neurons goes to infinity (see \Cref{thm:PFunc}). We even show that resulting function $\NN_{\theta^{*,\lw}}=\lim_{n\to\infty}\NN_{\theta^{*,\lw}}$ already reaches the limit when $n_j\geq n^*_j$ for all $ j \in \fromto{\nStacks}$, where $n^*_j\leq d_j \cdot N +1$ (see \Cref{cor:wide_dj}). From this we derive that arbitrarily wide $\NN_{\theta^{*,\lw}}$ can benefit from multi-task learning by representation-learning (even if $n_j\gg d_j \cdot N$ and even a single hidden layer is sufficient, i.e., for \ $\nStacks\geq 1$).

In contrast, the MAP of a GP is
not capable of benefiting from multi-task learning or representation learning because of the fixed data-independent kernel (see \Cref{prop:GPnoMultiTask} and its discussion).

This might sound contradicting since an insufficient
summary of \citet{Neal1996} could be naively formulated as \enquote{Infinitely wide Gaussian BNNs are equivalent to shallow GPs}.

The solution to this paradox is that exchanging the order of taking the MAP, going from parameter space to function space, and taking the limit of width~$n$ to infinity
vastly changes the behavior of the obtained function: \citet{Neal1996} %
shows that the prior of a very wide BNN is similar to a GP%
. However, \citet{Neal1996} never claims that the MAP~$\theta^{*,\lambda}$ on the parameter space of a very wide BNN is, as a function~$\NN_{\theta^{*,\lambda}}$, close to the MAP of the corresponding GP%
. In fact, we show in \Cref{thm:PFunc,prop:MultiTask,prop:GPnoMultiTask} that a sufficiently wide~$\NN_{\theta^{*,\lw}}$ as given in \cref{eq:optimalParameters} is typically not close at all to the MAP of the corresponding GP. This result is important since the gradient descent-based algorithms typically used in practice aim to approximate \cref{eq:optimalParameters}, and thus solutions found in practice are not accurately described by the GP's MAP.

\subsection{Is Poor Man's Bayes Better Than Optimal Bayes?}
Mathematically, in a correct Bayesian regression setting, the mean a posteriori on function space is the best possible model in terms of expected mean squared error (MSE) on a test set. The MAP (on the parameter space) is not optimal in this sense, but it can be cheaper to compute and is often a reasonable approximation of the mean a posteriori. Therefore, the MAP is called \enquote{poor man's Bayes}. 

One consequence of our result is the following plot-twist:
One would expect the exact mean a posterioiri of a wide BNN (given by $\mathcal{M}^\text{GP}(\Ltr)$) to perform even better in terms of expected test MSE than its poor approximation~$\NN_{\theta^{*,\lw}}$ (where $\theta^{*,\lw}$ is the MAP on the parameter space).
However, our results suggest that $\NN_{\theta^{*,\lw}}$ can generalize better than $\mathcal{M}^\text{GP}(\Ltr)$:
While \Cref{prop:MultiTask} shows that $\NN_{\theta^{*,\lw}}$ allows for multi-task learning, $\mathcal{M}^\text{GP}(\Ltr)$ does not (\Cref{prop:GPnoMultiTask}).
Paradoxically, this suggests that the poor man's approximation has lower expected test MSE than the theoretical optimum.

\paragraph{The Solution to the Paradox.} When the true prior is a wide Gaussian BNN, the mean a posteriori~$\mathcal{M}^\text{GP}(\Ltr)$ is better than $\NN_{\theta^{*,\lw}}$ in terms of expected MSE. However, if the Gaussian BNN prior is not close enough to the true prior, this statement does not hold anymore.
We conclude that the improved generalization behavior of $\NN_{\theta^{*,\lw}}$ over $\mathcal{M}^\text{GP}(\Ltr)$ does not contradict with Bayesian theory; it simply suggests that Gaussian BNNs are quite far from the true prior and that $\NN_{\theta^{*,\lw}}$ is better in approximating the mean a posterioiri of the unknown true prior than $\mathcal{M}^\text{GP}(\Ltr)$.

\section{Conclusion}

In this paper, we gave an exact characterization for infinite-width deep $\ell^2$-regularized ReLU NNs.  
Further, we gave a mathematical definition of multi-task learning on function space and showed that, based on our characterization, infinitely wide, $\ell^2$-regularized ReLU NNs are capable of multi-task learning.
We highlighted in particular, that many infinite-width limits of NNs previously derived in the literature, like shallow GPs are not capable of multi-task-learning.
These observations enabled us to explain two paradoxical contrasts between infinite-width $\ell^2$-regularized NNs and infinite-width Gaussian BNNs: First, the optimal infinite-width $\ell^2$-regularized NN does not correspond to the MAP of the corresponding GP.
Second, the two errors of poor man's approximation of the mean a posteriori BNN and of the poor choice of the prior partially cancel out and this results in an estimator that better approximates the mean a posteriori of some other unknown prior.
In particular, the latter observation could give further\footnote{\citet{aitchison2020statisticalTheoryColdPosterior} only partially solves the paradox explained in \cite{wenzel2020ColdPosterior}.} insights into the much discussed phenomenon of cold posteriors of Gaussian BNNs \cite{wenzel2020ColdPosterior}. We will investigate this further in future work.

We mathematically proved in \cref{sec:DeepVsShallow} that different outputs of infinitely-wide $\ell^2$-regularized ReLU NNs can influence each other in the infinite width limit.
We hope that our characterization of the infinite-width limit can also provide some intuition on how they influence each other, how this influence can be beneficial in terms of generalization, and how it explains representation learning, feature learning, metric learning or transfer learning.
In future work, we provide empirical evidence and more discussions on the beneficial effects of multi-task learning and on its relation to model uncertainty \cite[Desiderata D4]{NOMUarxiv}.

Our results always consider NNs that are optimal with respect to an $\ell^2$-regularized loss. In future work, we compare them to NNs trained using standard (gradient descent based) optimization algorithms with and without $\ell^2$-regularization, and compare to the results of \cite{YangFeatureLearningInfiniteWidth}.
Finally, we extend our results to other architectures.

\nonAnonymousText{\section*{Acknowledgements}
The authors gratefully acknowledge the support from ETH-foundation. We are very thankful for numerous helpful discussions, feedback and code-implementations -- especially to Alexis Stockinger, Sven Rosenthal and Sebastian Schein. Further we want to thank  Gudmund Pammer, Anastasis Kratsios, Michael Heiss, Kei Ishikawa, Tereza Burgetová, Jakob Weissteiner, Dorothea Heiss, Lukas Fertl and many more for the feedback and discussions. 
} 

\clearpage{}%
\appendix
\section{Network Architectures and Their P-functionals}\label{sec:different architectures}

\begin{figure}[htbp]
\centering
\begin{minipage}[t]{.3\linewidth}
\centering
\begin{adjustbox}{width=\linewidth, totalheight=1.5\linewidth}
\begin{tikzpicture}[x=0.75pt,y=0.75pt,yscale=-1,xscale=1]
\draw   (90.05,586.55) -- (125.15,586.55) -- (125.15,685) -- (90.05,685) -- cycle ;
\draw   (155.96,458) -- (191.49,458) -- (191.49,800.73) -- (155.96,800.73) -- cycle ;
\draw   (226.5,611) -- (263.31,611) -- (263.31,657) -- (226.5,657) -- cycle ;
\draw    (125.15,586.55) -- (155.96,458) ;
\draw    (125.15,685) -- (155.96,800.73) ;
\draw    (191.49,458) -- (226.5,611) ;
\draw    (191.49,800.73) -- (226.5,657) ;

\draw (160.42,626) node [anchor=north west][inner sep=0.75pt]  [font=\scriptsize] [align=left] {\scalebox{1}[1.1]{ReLU}\\ \scalebox{0.9}[1.1]{($n_j$)}};
\draw (229.24,627) node [anchor=north west][inner sep=0.75pt]  [font=\scriptsize] [align=left] {\scalebox{0.9}[1.1]{$\link^{-1}$}\\
\scalebox{0.9}[1.1]{($d_j$)}};
\draw (95.07,626) node [anchor=north west][inner sep=0.75pt]  [font=\scriptsize] [align=left] {\scalebox{1}[1.1]{Input}\\ \scalebox{0.9}[1.1]{($d_{j-1}$)} };

\end{tikzpicture}
\end{adjustbox}
\caption{Schematic representation of a shallow neural network, which corresponds to one stack in the main paper.}
\label{fig:shallow_nn}
\end{minipage}
\hfill
\begin{minipage}[t]{.3\linewidth}
\centering
\begin{adjustbox}{width=\linewidth, totalheight=1.5\linewidth}
\begin{tikzpicture}[x=0.75pt,y=0.75pt,yscale=-1,xscale=1]
\draw   (124.05,689.55) -- (159.15,689.55) -- (159.15,788) -- (124.05,788) -- cycle ;
\draw   (189.96,489.09) -- (225.49,489.09) -- (225.49,732.64) -- (189.96,732.64) -- cycle ;
\draw   (260.5,714) -- (297.31,714) -- (297.31,760) -- (260.5,760) -- cycle ;
\draw    (159.15,689.55) -- (189.96,489.09) ;
\draw    (159.15,788) -- (189.96,732.64) ;
\draw    (225.49,489.09) -- (260.5,714) ;
\draw    (225.49,732.64) -- (260.5,760) ;
\draw    (159.15,689.55) .. controls (207.05,818) and (209.05,784) .. (260.5,714) ;
\draw    (159.15,788) .. controls (217.05,824) and (224.05,824) .. (260.5,760) ;

\draw (194.42,600) node [anchor=north west][inner sep=0.75pt]  [font=\scriptsize] [align=left] {\scalebox{1}[1.1]{ReLU}\\ \scalebox{0.9}[1.1]{($n_j$)}};
\draw (263.24,730) node [anchor=north west][inner sep=0.75pt]  [font=\scriptsize] [align=left] {\scalebox{0.9}[1.1]{$\link^{-1}$}\\
\scalebox{0.9}[1.1]{($d_j$)}};
\draw (129.07,729) node [anchor=north west][inner sep=0.75pt]  [font=\scriptsize] [align=left] {\scalebox{1}[1.1]{Input}\\ \scalebox{0.9}[1.1]{($d_{j-1}$)} };

\end{tikzpicture}
\end{adjustbox}
\captionsetup{skip=5pt} %
\caption{Schematic representation of a shallow neural network with a trainable linear skip connection}
\label{fig:shallow_nn_skip}
\end{minipage}
\hfill
\begin{minipage}[t]{.3\linewidth}
\begin{adjustbox}{width=\linewidth,totalheight=1.5\linewidth}
\begin{tikzpicture}[x=0.75pt,y=0.75pt,yscale=-1,xscale=1]
\draw   (124.05,689.55) -- (159.15,689.55) -- (159.15,788) -- (124.05,788) -- cycle ;
\draw   (189.96,489.09) -- (225.49,489.09) -- (225.49,732.64) -- (189.96,732.64) -- cycle ;
\draw   (260.5,714) -- (297.31,714) -- (297.31,760) -- (260.5,760) -- cycle ;
\draw    (159.15,689.55) -- (189.96,489.09) ;
\draw    (159.15,788) -- (189.96,732.64) ;
\draw    (225.49,489.09) -- (260.5,714) ;
\draw    (225.49,732.64) -- (260.5,760) ;
\draw   (188.05,776) -- (224.05,776) -- (224.05,822) -- (188.05,822) -- cycle ;
\draw    (159.15,689.55) -- (188.05,776) ;
\draw    (159.15,788) -- (188.05,822) ;
\draw    (260.5,714) -- (224.05,776) ;
\draw    (260.5,760) -- (224.05,822) ;

\draw (194.42,600) node [anchor=north west][inner sep=0.75pt]  [font=\scriptsize] [align=left] {\scalebox{1}[1.1]{ReLU}\\ \scalebox{0.9}[1.1]{($n_j$)}};
\draw (263.24,730) node [anchor=north west][inner sep=0.75pt]  [font=\scriptsize] [align=left] {\scalebox{0.9}[1.1]{$\link^{-1}$}\\
\scalebox{0.9}[1.1]{($d_j$)}};
\draw (129.07,729) node [anchor=north west][inner sep=0.75pt]  [font=\scriptsize] [align=left] {\scalebox{1}[1.1]{Input}\\ \scalebox{0.9}[1.1]{($d_{j-1}$)} };
\draw (191.19,790) node [anchor=north west][inner sep=0.75pt]  [font=\scriptsize] [align=left] {\scalebox{1}[1.1]{Linear}\\
\scalebox{0.5}[0.75]{\fontsize{5}{6}\selectfont($\min(d_{j-1},d_j)$)}};

\end{tikzpicture}
\end{adjustbox}
\caption{Schematic representation of a shallow neural network where many ($n_j$) hidden nodes have a ReLU activation and some ($\min(d_{j-1},d_j)$) hidden nodes have a linear (identity) activation function. Compared to \Cref{fig:shallow_nn_skip} the linear skip connection is factorized into two matrices.}
\label{fig:one_stack}
\end{minipage}

\end{figure} %
\begin{figure}[htbp]
\centering
\begin{minipage}{.45\linewidth}
\begin{adjustbox}{width=\linewidth} %
\begin{tikzpicture}[x=0.75pt,y=0.75pt,yscale=-1,xscale=1]
\draw   (73.05,208.55) -- (108.15,208.55) -- (108.15,307) -- (73.05,307) -- cycle ;
\draw   (138.96,72) -- (174.49,72) -- (174.49,277) -- (138.96,277) -- cycle ;
\draw   (209.5,259) -- (246.31,259) -- (246.31,305) -- (209.5,305) -- cycle ;
\draw   (280.42,72) -- (316.33,72) -- (316.33,277.45) -- (280.42,277.45) -- cycle ;
\draw   (353.14,241.36) -- (389.05,241.36) -- (389.05,319.36) -- (353.14,319.36) -- cycle ;
\draw    (108.15,208.55) -- (138.96,72) ;
\draw    (108.15,307) -- (138.96,277) ;
\draw    (174.49,72) -- (209.5,259) ;
\draw    (246.31,259) -- (280.42,72) ;
\draw    (174.49,277) -- (209.5,305) ;
\draw    (246.31,305) -- (280.42,277.45) ;
\draw    (316.33,72) -- (353.14,241.36) ;
\draw    (316.33,277.45) -- (353.14,319.36) ;
\draw   (423.42,73) -- (459.33,73) -- (459.33,278.45) -- (423.42,278.45) -- cycle ;
\draw   (496.14,261) -- (532.05,261) -- (532.05,306.45) -- (496.14,306.45) -- cycle ;
\draw    (389.05,241.36) -- (423.42,73) ;
\draw    (389.05,319.36) -- (423.42,278.45) ;
\draw    (459.33,73) -- (496.14,261) ;
\draw    (459.33,278.45) -- (496.14,306.45) ;
\draw    (108.15,208.55) .. controls (156.05,341.36) and (135.05,339.36) .. (209.5,259) ;
\draw    (108.15,307) .. controls (129.05,372.36) and (181.05,348.36) .. (209.5,305) ;
\draw    (246.31,259) .. controls (277.05,361.36) and (335.05,341.36) .. (353.14,241.36) ;
\draw    (246.31,305) .. controls (266.05,375.36) and (323.05,356.36) .. (353.14,319.36) ;
\draw    (389.05,241.36) .. controls (437.94,373.18) and (459.05,313.36) .. (496.14,261) ;
\draw    (389.05,319.36) .. controls (419.05,361.36) and (487.05,360.36) .. (496.14,306.45) ;

\draw (143.42,166) node [anchor=north west][inner sep=0.75pt]  [font=\scriptsize] [align=left] {ReLU\\
($n_1$)};
\draw (284.98,166) node [anchor=north west][inner sep=0.75pt]  [font=\scriptsize] [align=left] {ReLU\\
($n_2$)};
\draw (212.24,273) node [anchor=north west][inner sep=0.75pt]  [font=\scriptsize] [align=left] {$\sigb$\\
($d_1$)};
\draw (354.83,273) node [anchor=north west][inner sep=0.75pt]  [font=\scriptsize] [align=left] {$\sigb$\\
($d_2$)};
\draw (79.07,247) node [anchor=north west][inner sep=0.75pt]  [font=\scriptsize] [align=left] {Input\\
($d_0$)};
\draw (427.98,167) node [anchor=north west][inner sep=0.75pt]  [font=\scriptsize] [align=left] {ReLU\\
($n_3$)};
\draw (497.83,274) node [anchor=north west][inner sep=0.75pt]  [font=\scriptsize] [align=left] {$\link^{-1}$\\
($d_3$)};

\end{tikzpicture} \end{adjustbox} 
\captionsetup{skip=-5pt}
\caption{Schematic representation of $\nStacks=3$ stacks of \Cref{fig:shallow_nn_skip}.}
\label{fig:simp_architecture}
\end{minipage}
\hfill
\begin{minipage}{.45\linewidth}
\begin{adjustbox}{width=\linewidth} %
\begin{tikzpicture}[x=0.75pt,y=0.75pt,yscale=-1,xscale=1]
\draw   (73.05,208.55) -- (108.15,208.55) -- (108.15,307) -- (73.05,307) -- cycle ;
\draw   (138.96,72) -- (174.49,72) -- (174.49,277) -- (138.96,277) -- cycle ;
\draw   (209.5,259) -- (246.31,259) -- (246.31,305) -- (209.5,305) -- cycle ;
\draw   (280.42,72) -- (316.33,72) -- (316.33,277.45) -- (280.42,277.45) -- cycle ;
\draw   (353.14,241.36) -- (389.05,241.36) -- (389.05,319.36) -- (353.14,319.36) -- cycle ;
\draw    (108.15,208.55) -- (138.96,72) ;
\draw    (108.15,307) -- (138.96,277) ;
\draw    (174.49,72) -- (209.5,259) ;
\draw    (246.31,259) -- (280.42,72) ;
\draw    (174.49,277) -- (209.5,305) ;
\draw    (246.31,305) -- (280.42,277.45) ;
\draw    (316.33,72) -- (353.14,241.36) ;
\draw    (316.33,277.45) -- (353.14,319.36) ;
\draw   (282.22,319.45) -- (317.23,319.45) -- (317.23,364) -- (282.22,364) -- cycle ;
\draw    (246.31,259) -- (282.22,319.45) ;
\draw    (246.31,305) -- (282.22,364) ;
\draw    (317.23,319.45) -- (353.14,241.36) ;
\draw    (317.23,364) -- (353.14,319.36) ;
\draw   (136.78,319.45) -- (171.8,319.45) -- (171.8,364) -- (136.78,364) -- cycle ;
\draw    (108.15,208.55) -- (136.78,319.45) ;
\draw    (108.15,307) -- (136.78,364) ;
\draw    (171.8,319.45) -- (209.5,259) ;
\draw    (171.8,364) -- (209.5,305) ;
\draw   (423.42,73) -- (459.33,73) -- (459.33,278.45) -- (423.42,278.45) -- cycle ;
\draw   (496.14,261) -- (532.05,261) -- (532.05,306.45) -- (496.14,306.45) -- cycle ;
\draw    (389.05,241.36) -- (423.42,73) ;
\draw    (389.05,319.36) -- (423.42,278.45) ;
\draw    (459.33,73) -- (496.14,261) ;
\draw    (459.33,278.45) -- (496.14,306.45) ;
\draw   (425.22,320.45) -- (460.23,320.45) -- (460.23,365) -- (425.22,365) -- cycle ;
\draw    (389.05,241.36) -- (425.22,320.45) ;
\draw    (389.05,319.36) -- (425.22,365) ;
\draw    (460.23,320.45) -- (496.14,261) ;
\draw    (460.23,365) -- (496.14,306.45) ;

\draw (143.42,166) node [anchor=north west][inner sep=0.75pt]  [font=\scriptsize] [align=left] {ReLU\\
($n_1$)};
\draw (284.98,166) node [anchor=north west][inner sep=0.75pt]  [font=\scriptsize] [align=left] {ReLU\\
($n_2$)};
\draw (212.24,273) node [anchor=north west][inner sep=0.75pt]  [font=\scriptsize] [align=left] {$\sigb$\\
($d_1$)};
\draw (354.83,273) node [anchor=north west][inner sep=0.75pt]  [font=\scriptsize] [align=left] {$\sigb$\\
($d_2$)};
\draw (285.63,330) node [anchor=north west][inner sep=0.75pt]  [font=\scriptsize] [align=left] {Linear\\
\scalebox{0.5}{\!\!($\min(d_1,d_2)$)}};
\draw (78.07,247) node [anchor=north west][inner sep=0.75pt]  [font=\scriptsize] [align=left] {Input\\
($d_0$)};
\draw (140.19,330) node [anchor=north west][inner sep=0.75pt]  [font=\scriptsize] [align=left] {Linear\\
\scalebox{0.5}{\!\!($\min(d_0,d_1)$)}};
\draw (427.98,167) node [anchor=north west][inner sep=0.75pt]  [font=\scriptsize] [align=left] {ReLU\\
($n_3$)};
\draw (497.83,274) node [anchor=north west][inner sep=0.75pt]  [font=\scriptsize] [align=left] {$\link^{-1}$\\
($d_3$)};
\draw (428.63,331) node [anchor=north west][inner sep=0.75pt]  [font=\scriptsize] [align=left] {Linear\\
\scalebox{0.5}{\!\!($\min(d_2,d_3)$)}};

\end{tikzpicture} \end{adjustbox} 
\caption{Schematic representation of $\nStacks=3$ stacks of \Cref{fig:one_stack}.}
\label{fig:architecture}
\end{minipage}     

\end{figure}
In a future version of this paper, we will discuss \Pfunc-functionals for modifications of the NN architecture. For architectures of the type of \Cref{fig:shallow_nn_skip,fig:simp_architecture}, for instance, $P_j$ (and $\T$) are given with the same abuse of notation as in the main paper (treating distributions as if they were functions) as
\begin{equation}\label{eq:Pj_simple_skip}
	\Pfunc_j(h_j):=  \min_{\substack{\varphi\in\T,\ c\in\R^{d_j},A\in\R^{d_{j-1}\times d_j}\text{ s.t.}\\  h_j=\int_{\Sd[d_{j-1}-1]}\psr[\langle s,\cdot\rangle]{s}\, ds+c + A(\cdot)}} \left(
	2\int_{\Sd[d_{j-1}-1]}\int_{\R} \frac{\twonorm[ {\psr{s}}^{''} ]}{g(r)} \dx[r]\dx[s] +2\rho(\twonorm[c])	+ \twonorm[A]^2	\right),%
\end{equation}
where $\twonorm[A]$ is the Frobenius norm of $A$ and

\begin{align*}
	\T:=\bigg\{ \varphi=(\varphi_s)_{s\in\Sd[d_{j-1}-1]}  &\bigg|\, \forall s\in\Sd[d_{j-1}-1] :\varphi_s:\R\to\R^{d_j},\lim_{r\to -\infty} \psr{s}=0 \text{ and } \lim_{r\to -\infty} \frac{\partial}{\partial r}\psr{s}=0\bigg\}.
\end{align*}
For an architecture of the type of \Cref{fig:one_stack,fig:architecture} one gets:
\begin{equation}\label{eq:Pj_advanced_skip}
	\Pfunc_j(h_j):=  \min_{\substack{\varphi\in\T,\ c\in\R^{d_j},A\in\R^{d_{j-1}\times d_j}\text{ s.t.}\\  h_j=\int_{\Sd[d_{j-1}-1]}\psr[\langle s,\cdot\rangle]{s}\, ds+c + A(\cdot)}} \left(
	2\int_{\Sd[d_{j-1}-1]}\int_{\R} \frac{\twonorm[ {\psr{s}}^{''} ]}{g(r)} \dx[r]\dx[s] +2\rho(\twonorm[c])	+ 2\left\|A\right\|_{\href{https://en.wikipedia.org/wiki/Schatten_norm}{\text{Schatten}1}}	\right),%
\end{equation}
where $\left\|A\right\|_{\href{https://en.wikipedia.org/wiki/Schatten_norm}{\text{Schatten}1}}$ is the \href{https://en.wikipedia.org/wiki/Schatten_norm}{Schatten 1-norm} (or Schatten–von-Neumann 1-norm) of $A$, i.e., the sum of the absolute values of the singular values of $A$.

\section{Equivalent Formulations of Regularization Functionals \texorpdfstring{$\Pfunc$}{P}}\label{sec:EquivalentPFuncs}

In this section, we state equivalent formulations of regularizing functional $\Pfunc$. See \Cref{subsec:EquivalencesofP} for proofs of their equivalence. %
Moreover, we set $g(r):=\frac{1}{\sqrt{1+r^2}}$ as in the main paper.

We start with some basic definitions.
\begin{definition}\label{def:BV2}
Let $\Om\subseteq\R$ be an open set. We define the Banach-space $\BVTwo(\Om;\R^{d_j})$ as the set of functions $f\in W^{1,1}(\R;\R^{d_j})$ which second distributional derivative $D^2f=\mu$ is a bounded $\R^{d_j}$-valued Radon-measure, i.e. $\int f \phi''=\int \phi d\mu \ \forall \phi\in C_c^2$, with norm

\begin{align*}
    \|f\|_{\BVTwo}=\|f\|_{L^1}+\|f'\|_{L^1}+|D^2f|,
\end{align*}
where\footnote{\label{footnote:measurablePartition}%
\enquote{partition} always refers to measurable partitions within this paper.%
}\textsuperscript{, }%
\footnote{\label{footnote:contextTV}%
Out of context it should be clear if $|\mu|$ refers to the total variation measure $|\mu|:\mathfrak{B}(\Om)\to\Rpz,B\mapsto\sup_{(E_i)_{1,\dots,n}\text{ is partition of } B,\,n\in\N}\sum_i\|\mu(E_i)\|_2$, or if $|\mu|$ is a short notation for $|\mu|(\Om)$.
Technically speaking, $D^2f=\mu$ in \Cref{def:BV2loc} is only defined on $\mathfrak{B}_c$ \citep[defined in][p.~168]{leoni2017firstSobolevBook} because $D^2f=\mu$ can be any \emph{locally} bounded Radon-measure for $f\in\BVTwoLoc$. However, in \Cref{def:BV20-} we only consider \emph{bounded} Radon-measures, thus we can extend $D^2f=\mu$ to $\mathfrak{B}$ for any $f\in\BVTwoZ$.%
}
\[
|\mu|:=|\mu|(\Om):=\sup_{(E_i)_{1,\dots,n}\text{ is partition of } \Om,\,n\in\N}\sum_{i=1}^n\|\mu(E_i)\|_2.%
\]
\end{definition}
Therefore, $\BVTwo(\R;\R^{d_j})$
consists of functions
$f\in W^{1,1}(\R;\R^{d_j})$ which fulfill that
$f'$ is in $\text{BV}(\R;\R^{d_j})$ where $\text{BV}(\R;\R^{d_j})$ is defined as in \citet[Section~7.1]{leoni2017firstSobolevBook}.

\begin{definition}\label{def:BV2loc}
We define $\BVTwoLoc(\R;\R^{d_j})$ as the set of functions $f\in W_\text{loc}^{1,1}(\R;\R^{d_j})$ which second distributional derivative $D^2f=\mu$ is a locally bounded $\R^{d_j}$-valued Radon-measure (i.e. $\int f \phi''=\int \phi d\mu \ \forall \phi\in C_c^2$).
For the remainder of the paper, we will often just write  $\BVTwoLoc$ as a short notation for  $\BVTwoLoc(\R;\R^{d_j})$.
\end{definition}
In other words $\BVTwoLoc(\R;\R^{d_j})$ consists of functions $f\in W_\text{loc}^{1,1}(\R;\R^{d_j})$ such that the weak derivative $f'$ is in $\text{BV}_{\text{loc}}(\R;\R^{d_j})$ where $\text{BV}_{\text{loc}}(\R;\R^{d_j})$ is defined as in \citet[p.~188]{leoni2017firstSobolevBook}.

Throughout the paper, it is $g(r):=\frac{1}{\sqrt{1+r^2}}$ as in \meqref{eq:Pj}, as it is also in the following definition:
\begin{definition}\label{def:BV20-} Based on \Cref{def:BV2loc} we define
\[\BVTwoZ(\R;\R^{d_j}):=\Set{ f\in \BVTwoLoc(\R;\R^{d_j}) | \lim_{r\to -\infty} f(r)=0 = \lim_{r\to -\infty} \frac{\partial}{\partial r}f(r), |D^2f|_{\frac{1}{g}}<\infty }.\]
We equip this space with the norm $|D^2(\cdot)|_{\frac{1}{g}}$,
where\textsuperscript{\ref{footnote:measurablePartition}, \ref{footnote:contextTV}}
\[|\mu|_{\frac{1}{g}}:=|\mu|_{\frac{1}{g}}(\R):=\sup_{(E_i)_{1,\dots,n}\text{ is partition of } \R,\,n\in\N}\sum_{i=1}^n\twonorm[\int_{E_i}\frac{1}{g(r)} d\mu(r)].\]
(This is a norm because of the boundary conditions.) (We will sometimes write $|\mu|_{\frac{1}{g(r)}}$ instead of $|\mu|_{\frac{1}{g}}$.)
For the remainder of the paper, we will often just write  $\BVTwoZ$ as a short notation for  $\BVTwoZ(\R;\R^{d_j})$.
\end{definition}

The following \Cref{def:PjDistribution} is the most straight forward mathematically precise interpretation of \cref{eq:Pj}, where we interpret the \enquote{function} $\varphi$ as a distribution, which is mathematically a $\BVTwoZ$-valued bounded Radon measures~$\mu$ on $\Sd[d_{j-1}-1]$ (for the jth stack). This $\mu$ has a $\BVTwoZ$-valued Radon--Nikodym density $\varphi:=\varphi^\mu:=\frac{d\mu}{d|\mu|}:\Sd[d_{j-1}-1]\to\BVTwoZ,s\mapsto\varphi_s$ which is actually a classical function in the strict mathematical sense. The advantage of the measure theory point of view is that $\mu$ and thus $|\mu|$ can also have \href{https://en.wikipedia.org/w/index.php?title=Atom_(measure_theory)&oldid=1079053554}{atoms} (\href{https://en.wikipedia.org/w/index.php?title=Dirac_delta_function&oldid=1085986916}{Dirac impulses}) and therefore the following mathematical definition can also deal precisely with these atoms without any notational ambiguities:
\begin{definition}\label{def:PjDistribution}
$\mathfrak{M}^{\BVTwoZ}(\Sd[d_{j-1}-1])$ is the set of all $\BVTwoZ$-valued bounded Radon measures on $\Sd[d_{j-1}-1]$ for some stack $j$ for which the $\BVTwoZ$-valued point-wise\footnote{We define a \enquote{point-wise Radon--Nikodym density} analogously to the classical Radon--Nikodym density except replacing the Banach space topology used in the definition of the \href{https://en.wikipedia.org/w/index.php?title=Bochner_integral&oldid=1108046365}{Bochner integral} by the topology of point-wise convergence \textit{tptc}.} Radon--Nikodym density $\varphi:=\varphi^\mu:=\frac{d\mu}{d|\mu|}:\Sd[d_{j-1}-1]\to\BVTwoZ,s\mapsto\varphi_s$ exists.\footnote{%
Out of context, it should be clear that $|\mu|$ refers to the total variation measure $|\mu|:\mathfrak{B}(\Sd[\din-1])\to\Rpz$, while $|D^2\varphi_s|_{\frac{1}{g}}$ is a short notation of $|D^2\varphi_s|_{\frac{1}{g}}(\R)$ (see \Cref{def:BV20-}).%
} Then we define
{\small\[
    \Pfunc_j(h_j):= 
\min_{\substack{\mu_j\in\mathfrak{M}^{\BVTwoZ}(\Sd[d_{j-1}-1]),
\\ c\in\R^{d_j}\text{ s.t. } \forall x \in \R^{d_{j-1}}:\\ h_j(x)=c+
\int_{\Sd[{d_{j-1}}-1]} \varphi_s(\langle s,x\rangle) d|\mu_j|(s)
}}
\left(
2\rho\left(\twonorm[c]\right)
+2\int_{\Sd[{d_{j-1}}-1]}|D^2 \varphi_s|_{\frac{1}{g}} d|\mu_j|(s)
\right).
\]}

Moreover, 
\hypertarget{eq:cF}{\begin{align*}
    \cF:=\bigg\{
    f=\link^{-1}\circ h_{\nStacks}\circ\dots\circ \sigb\circ h_1\mid&
    \allIndi{\nStacks}{j}:\exists \mu_j\in\mathfrak{M}^{\BVTwoZ}(\Sd[d_{j-1}-1]), c^{(j)}\in\R^{d_j}:\\
    &\forall x \in \R^{d_{j-1}}:
     h_j(x)=c^{(j)}+\int_{\Sd[{d_{j-1}}-1]} \varphi_s(\langle s,x\rangle) d|\mu_j|(s)%
    \bigg\}.
\end{align*}}
\end{definition}

Note, that $\cF$ is dense in the continuous functions $\mathcal{C}(\R^\din;\R^\dout)$ with respect to the maximum norm on every compact set, if $d\geq\min(\din,\dout)$. (This can be easily derived from the universal approximation theorem in \citet{HornikUniversalApprox1991251,CybenkoUniversalApprox1989}.)

\begin{definition}\label{def:Pwellemeasure}
For $f\in\cF$ as in \Cref{def:PjDistribution}, we define
\begin{equation}\label{eq:Pwellemeasure}
\Pwellemeasure(f) = \min_{\substack{(h_1,\dots, h_{\nStacks})\text{, s.t. }\\ f=\link^{-1}\circ h_{\nStacks}\circ\dots\circ \sigb\circ h_1}} \left(\Pwellemeasure_1(h_1) + \Pwellemeasure_2(h_2) + \dots + \Pwellemeasure_\nStacks(h_\nStacks) \right),
\end{equation}
with
{\small\begin{align*}
\Pwellemeasure_{j}(h_j)
=\min\bigg\{2|\mu|+\twonorm[c]^2 :&
\mu_j\in\mathfrak{M}(\Sd[{d_{j-1}}]\times\Sd[{d_{j}}-1]),\\
&\forall x\in\R^{d_{j-1}}:h_j(x)=c+\int_{{\Sd[{d_{j-1}}]\times\Sd[{d_{j}}-1]}} w\sigma(\langle v, x\rangle-r)d\mu((v,r),w)\bigg\},
\end{align*}}
where $\mathfrak{M}(\Sd[{d_{j-1}}]\times\Sd[{d_{j}}-1])$ is the set of bounded Radon measures on $\Sd[{d_{j-1}}]\times\Sd[{d_{j}}-1]$.
\end{definition}
Moreover, we state two further (equivalent, as we will see in \Cref{subsec:EquivalencesofP}) reformulations of regularizing functional $P_j$ on the $j$\textsuperscript{th} stack.
\begin{definition}\label{def:Pmeasurevec}
{\small\begin{align*}\Pmeasurevec_j(h_j)=\min\bigg\{2|\mu|_{\frac{1}{g(r)}} +2\rho(\|c\|_2): &\mu\in\mathfrak{M}^{d_j}(\Sd[d_{j-1}-1]\times\R),\\
&\forall x\in\R^{d_{j-1}}: h_j(x)=c+\int_{\Sd[d_{j-1}-1]\times\R} \sigma(\langle v, x\rangle -r)d\mu(v,r)\bigg\},
\end{align*}}
with $\mathfrak{M}^{d_j}(\Sd[d_{j-1}-1]\times\R)$ the set of bounded $\R^{d_j}$-valued Radon measures on $\Sd[d_{j-1}-1]\times\R$ and

\[|\mu|_{\frac{1}{g(r)}}:=|\mu|_{\frac{1}{g(r)}}(\Sd[d_{j-1}-1]\times\R):=\sup_{(E_i)_{1,\dots,n}\text{ is partition of } \Sd[d_{j-1}-1]\times\R,\,n\in\N}\sum_{i=1}^n\twonorm[\int_{E_i}\frac{1}{g(r)} d\mu(v,r)].\]
\end{definition}

\begin{definition}\label{def:Pwellemeasureg}
{\scriptsize\begin{align*}\PwellemeasureG_{j}(h_j)
=\min\bigg\{2\int{\frac{1}{g(r)}}d\mu(v,r,w) +2\rho(\twonorm[c]):&
\mu\in\mathfrak{M}(\Sd[{d_{j-1}}-1]\times\R\times\Sd[d_{j}-1]),\\
&
\forall x\in\R^{d_{j-1}}: h_j(x)=c+\int_{\Sd[{d_{j-1}}-1]\times\R\times\Sd[d_j-1]} w\sigma(\langle v, x\rangle-r)d\mu(v,r,w)\bigg\}\end{align*}}
\end{definition}

\section{Required Number of Neurons}\label{sec:RequiredNumberofNeurons}
In the case of linear $\sigb$, a possible choice of $n^*$ and $d^*$ is always $n^*_j= \sum_{i=0}^{\nStacks-j+1}N^i$ and $d^*_j=\sum_{i=0}^{\nStacks-j}N^i$ independent of $\xtr, \ytr$ and $\lambda$.
A smaller choice can be  made in the case of letting some $d_j$ stay small and only letting $d_j$ tend to infinity for some indices $j$. This smaller choice can be written down recursively: $d^*_\nStacks=d_\nStacks=\dout$, $n^*_{\nStacks}=1+N$, and $d^*_j=\min(d_j,n^*_{j+1})$ and $n^*_j=1+N\cdot d^*_{j}$ for $j\in\fromto{\nStacks-1}$.

In the case of $\sigb=\text{ReLU}$, a possible choice of $n^*$ and $d^*$ is always $n^*_j= \sum_{i=0}^{2(\nStacks-j)+1}N^i$ and $d^*_j= \sum_{i=0}^{2(\nStacks-j)-1}N^i$ independent of $\xtr, \ytr$ and $\lambda$.
A smaller choice can be  made in the case of letting some $d_j$ stay small and only letting $d_j$ tend to infinity for some indices $j$. This smaller choice can be written down recursively: $d^*_\nStacks=d_\nStacks=\dout$, $n^*_{\nStacks}=1+N$, and  $d^*_j=\min(d_j,1+N\cdot n^*_{j+1})$ and $n^*_j=1+N\cdot d^*_{j}$ for $j\in\fromto{\nStacks-1}$.%

See \Cref{subsec:ProofOfCorollary} for proof of these statements.

\section{Visualizing Multi-task Learning}\label{sec:VisualizingMultiTask}
\begin{figure}[htb] %
     \centering
     \begin{subfigure}[t]{0.48\linewidth}
         \centering
         \includegraphics[width=\linewidth]{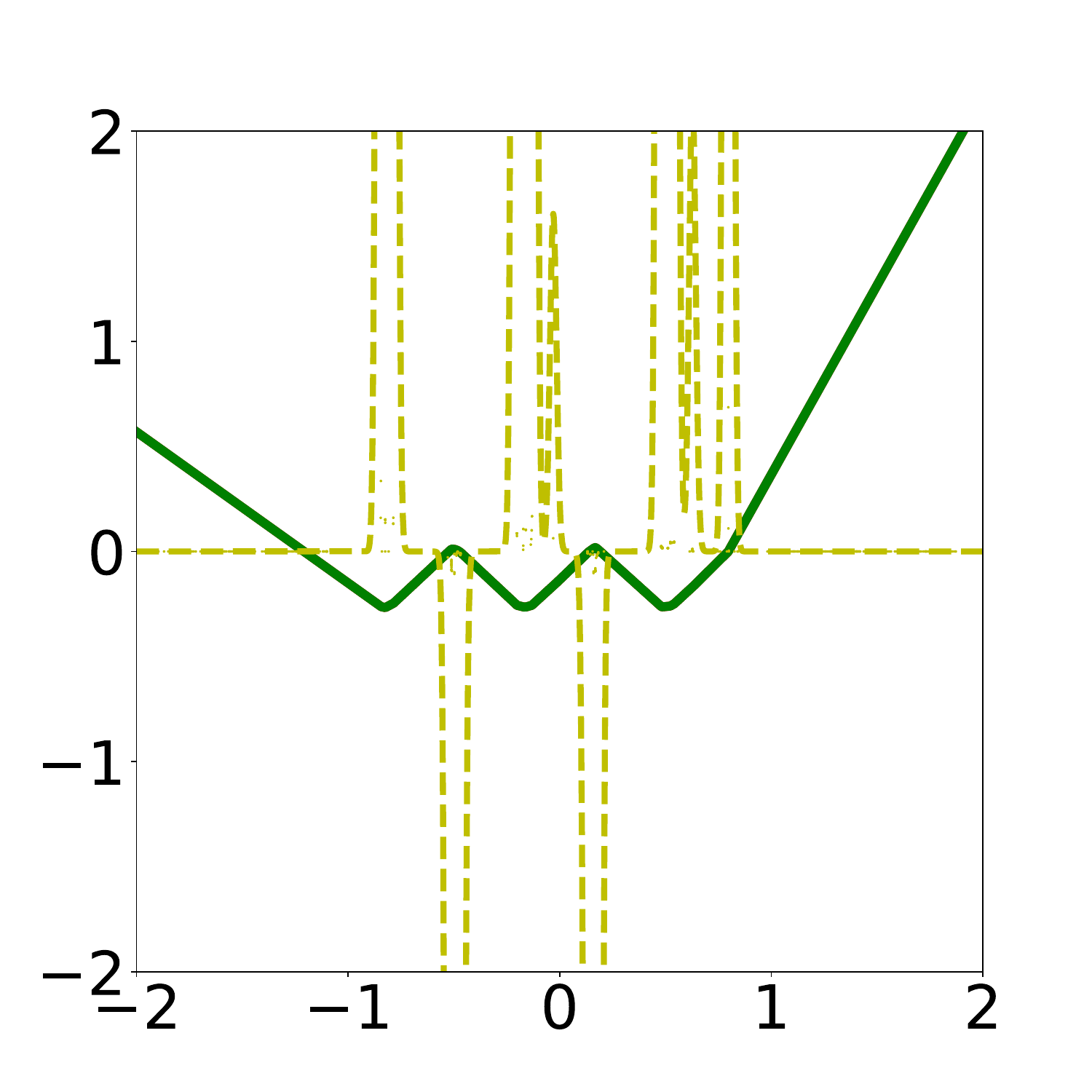}
         \caption{1\textsuperscript{st} stack: $\NNj{1}$ in green \unimportant{and its (distributional) second derivative~$\sum_{i=1}^{n_1} v^{(1)}_i w^{(1)}_i \delta_{\xi^{(1)}_i}$ (visualized by yellow dots $(\xi^{(1)}_i,v^{(1)}_i w^{(1)}_i)$), where $\xi^{(1)}_i=\frac{-b^{(1)}_i}{v^{(1)}_i}$ and a smoothed version of it (yellow dashed line). The smooth second derivative was obtained from a convolution using a Gaussian kernel.%
         }}
         \label{fig:first_stack}
     \end{subfigure}
     \hfill
     \begin{subfigure}[t]{0.48\linewidth}
         \centering
         \includegraphics[width=\linewidth]{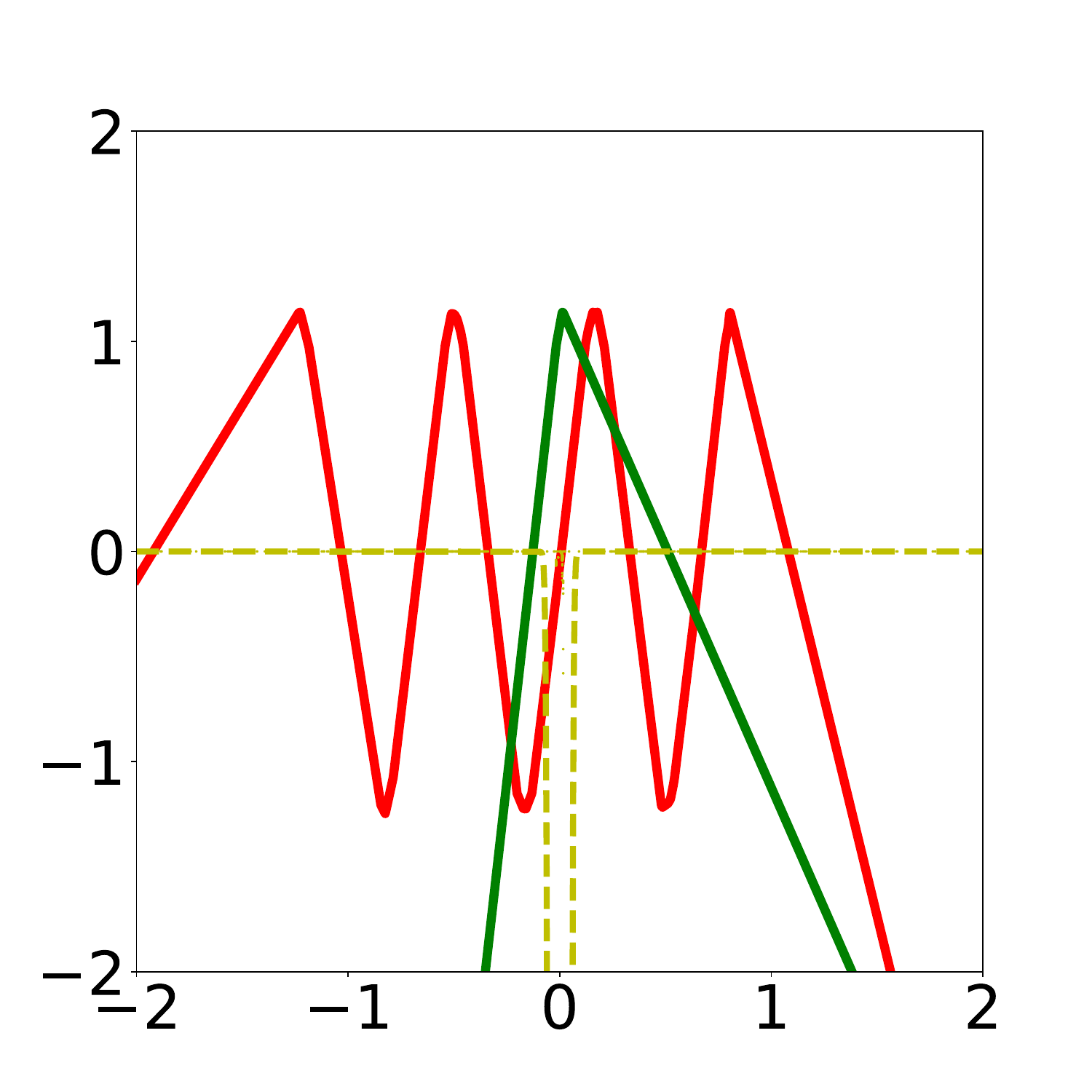}
         \caption{2\textsuperscript{nd} stack: $\NNj{2}$ in green and $H:=\NNj{2}\circ \NNj{1}$ in red
         \unimportant{and the (distributional) second derivative~$h_2^{''}=\sum_{i=1}^{n_2} v^{(2)}_i w^{(2)}_i \delta_{\xi^{(2)}_i}$ (visualized by yellow dots $(\xi^{(2)}_i,v^{(2)}_i w^{(2)}_i)$), where $\xi^{(2)}_i=\frac{-b^{(2)}_i}{v^{(2)}_i}$ and a smoothed version of it (yellow dashed line) as in \Cref{fig:first_stack}.%
         }}
         \label{fig:second_stack}
     \end{subfigure}
     \caption{The hidden stacks learn a periodic shared representation $H=\NNj{2}\circ \NNj{1}$}\label{fig:VisualizePeriodicH}
\end{figure}

\begin{figure}[htbp] %
     \centering
     \begin{subfigure}[b]{0.24\linewidth}
         \centering
         \includegraphics[width=\linewidth]{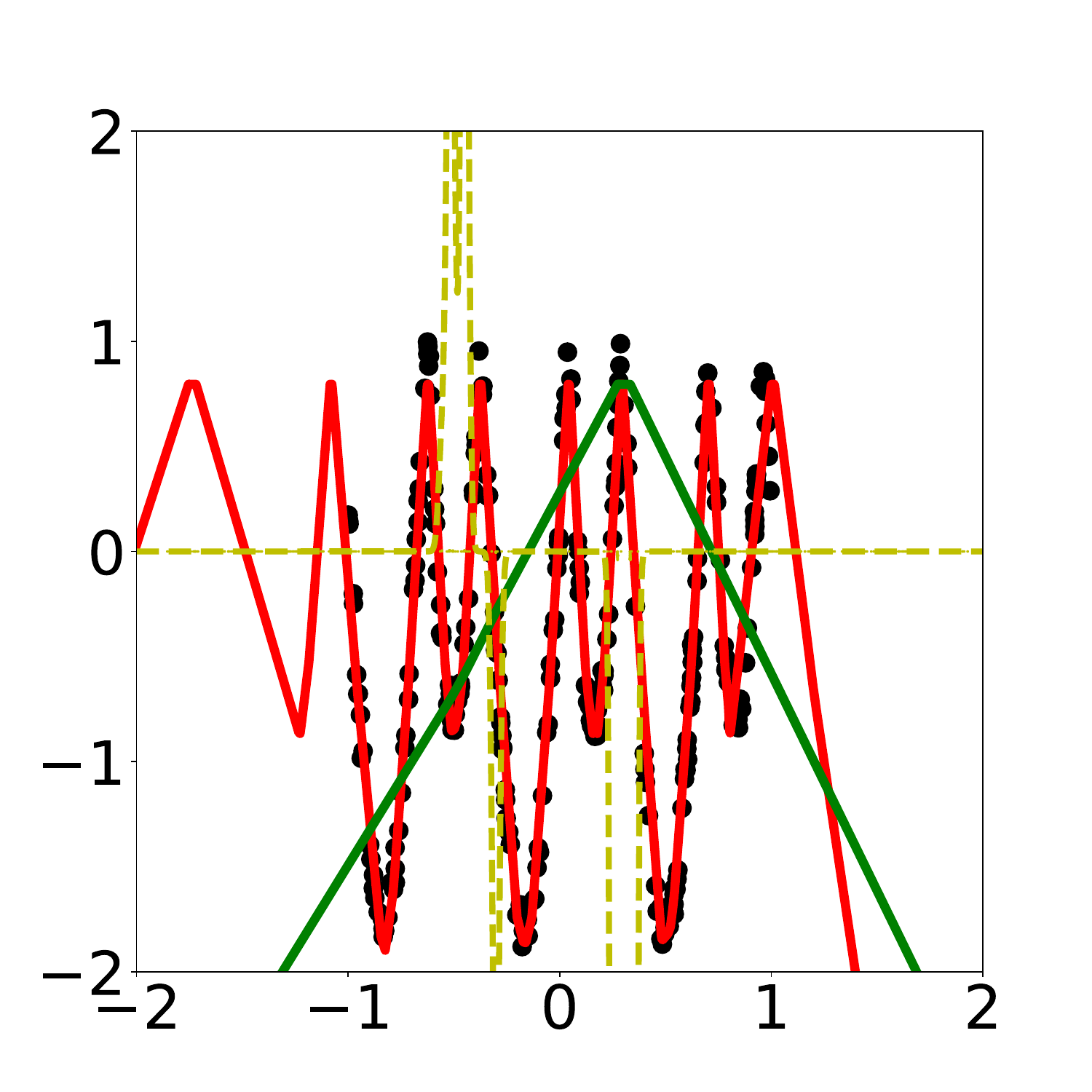}
         \caption{1\textsuperscript{st} component of third stack: ${(\NNj{3})}_1$ in green and $\hat{f}_1:={(\NN_\theta)}_1:={(\NNj{3})}_1\circ\NNj{2}\circ \NNj{1}$ in red and the training data points %
         $(\xtr_i,\ytr_{i})$ for $i\in I_1$
         as black dots.}
         \label{fig:one_kink}
     \end{subfigure}
     \hfill
     \begin{subfigure}[b]{0.24\linewidth}
         \centering
         \includegraphics[width=\linewidth]{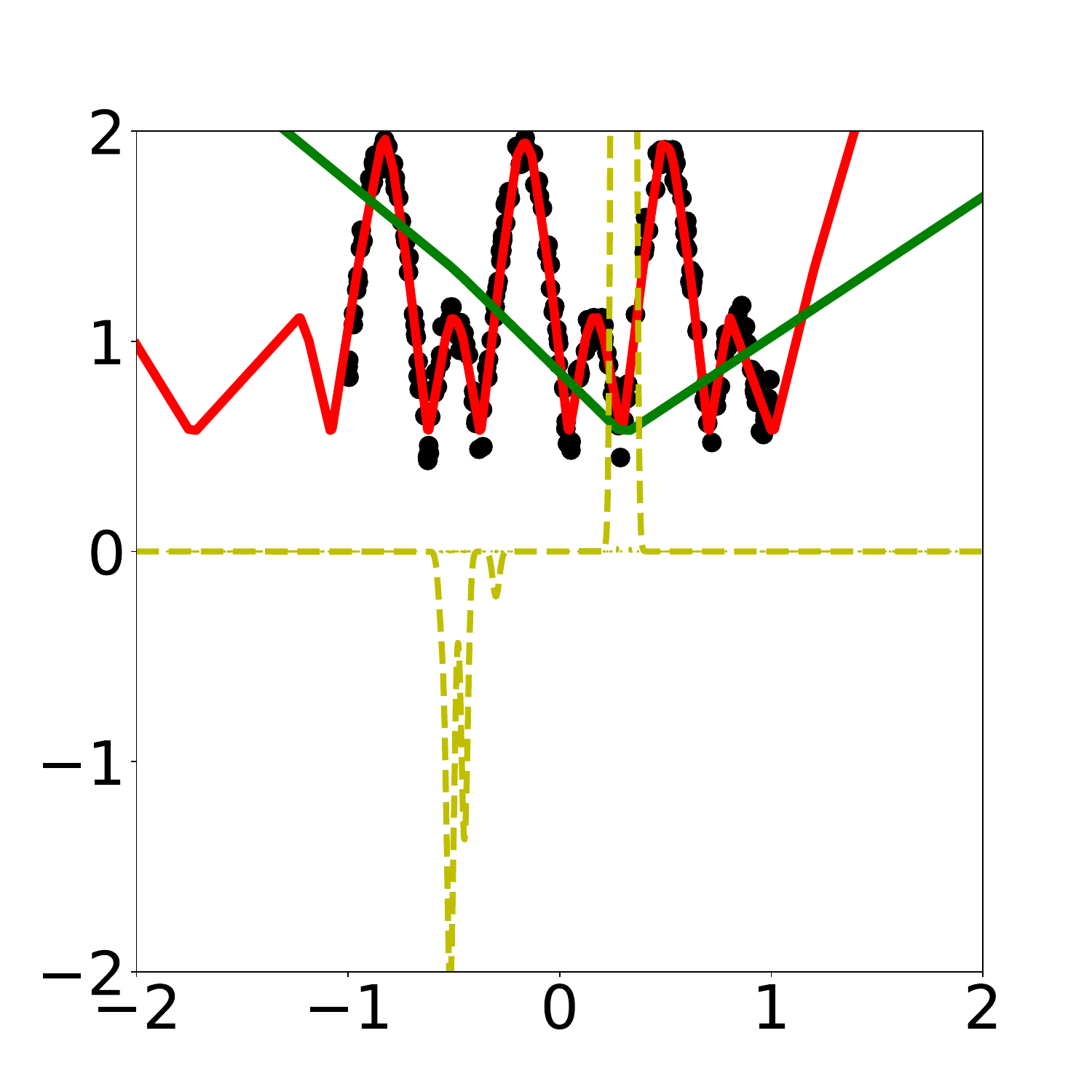}
         \caption{2\textsuperscript{nd} component of third stack: ${(\NNj{3})}_2$ in green and $\hat{f}_2:={(\NN_\theta)}_2:={(\NNj{3})}_2\circ\NNj{2}\circ \NNj{1}$ in red and the training data points %
         $(\xtr_i,\ytr_{i})$ for $i\in I_2$
         as black dots.}
         \label{fig:absolute_value}
     \end{subfigure}
     \hfill
     \begin{subfigure}[b]{0.24\textwidth}
         \centering
         \includegraphics[width=\textwidth]{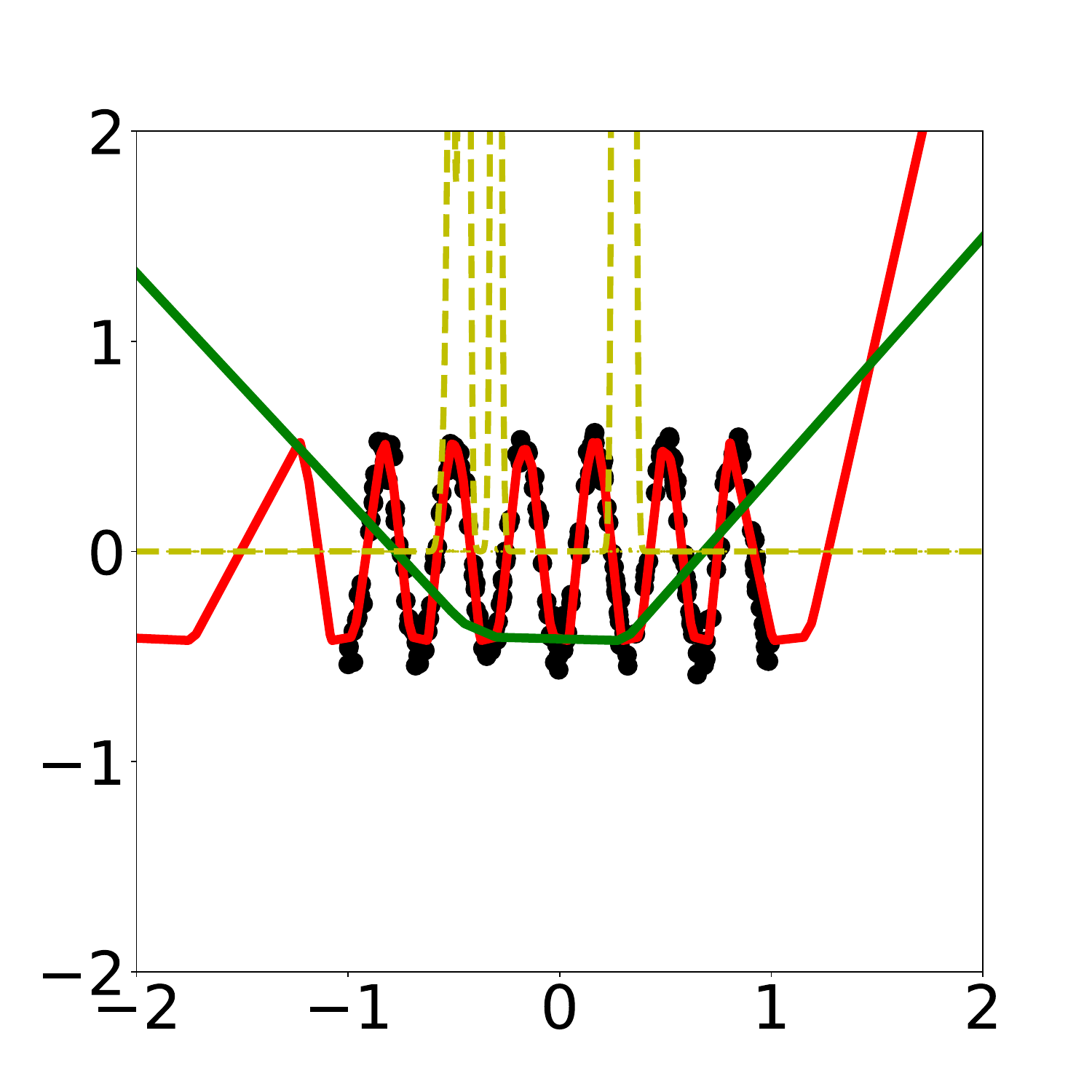}
         \caption{3\textsuperscript{rd} component of third stack: ${(\NNj{3})}_3$ in green and $\hat{f}_3:={(\NN_\theta)}_3:={(\NNj{3})}_3\circ\NNj{2}\circ \NNj{1}$ in red and the training data points %
         $(\xtr_i,\ytr_{i})$ for $i\in I_3$
         as black dots.}
         \label{fig:square}
     \end{subfigure}
     \begin{subfigure}[b]{0.24\textwidth}
         \centering
         \includegraphics[width=\textwidth]{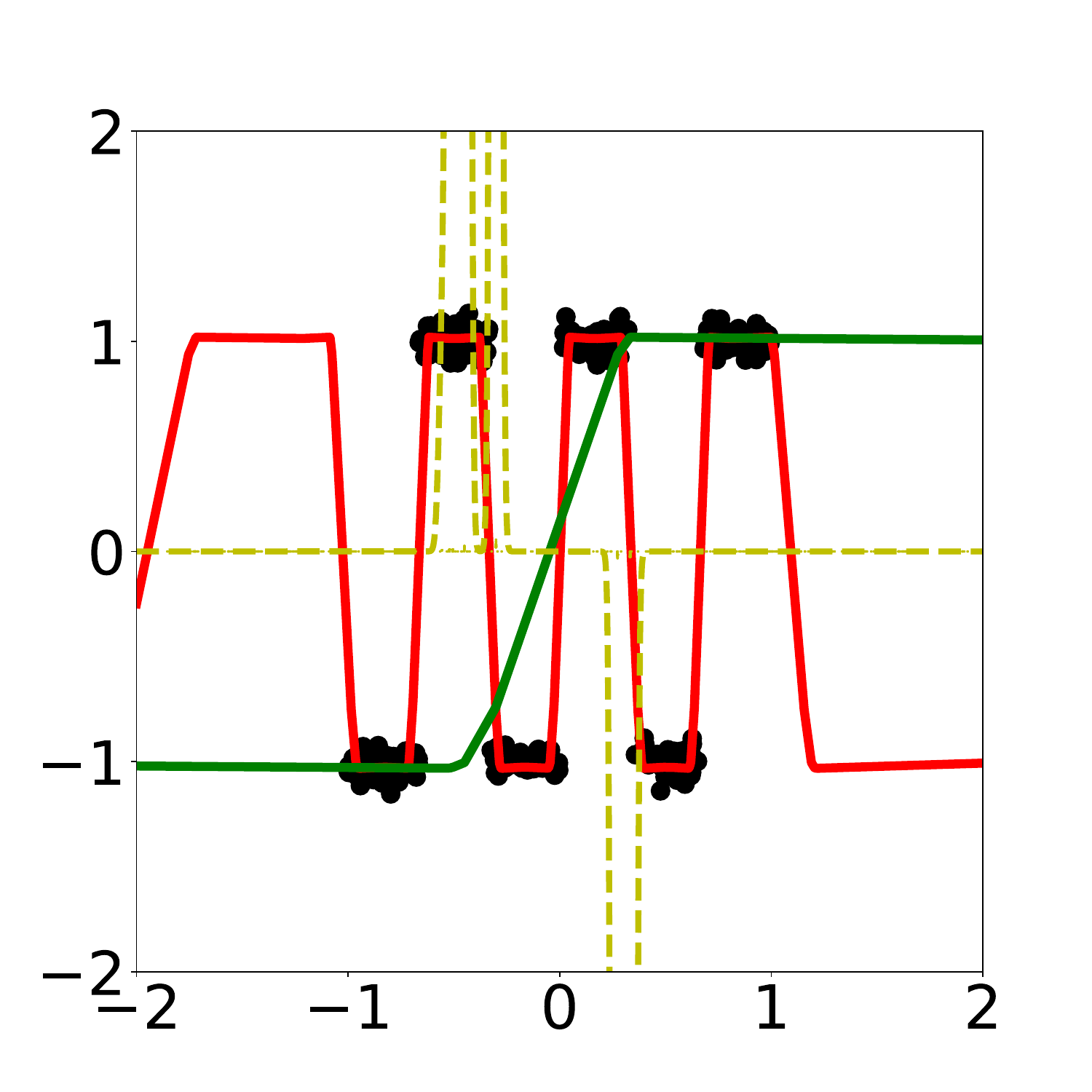}
         \caption{4\textsuperscript{th} component of third stack: ${(\NNj{3})}_4$ in green and $\hat{f}_4:={(\NN_\theta)}_4:={(\NNj{3})}_4\circ\NNj{2}\circ \NNj{1}$ in red and the training data points %
         $(\xtr_i,\ytr_{i})$ for $i\in I_4$
         as black dots.}
         \label{fig:sign}
     \end{subfigure}
     
     \begin{subfigure}[b]{0.12\linewidth}
     \end{subfigure}
     \begin{subfigure}[b]{0.24\linewidth}
         \centering
         \includegraphics[width=\linewidth]{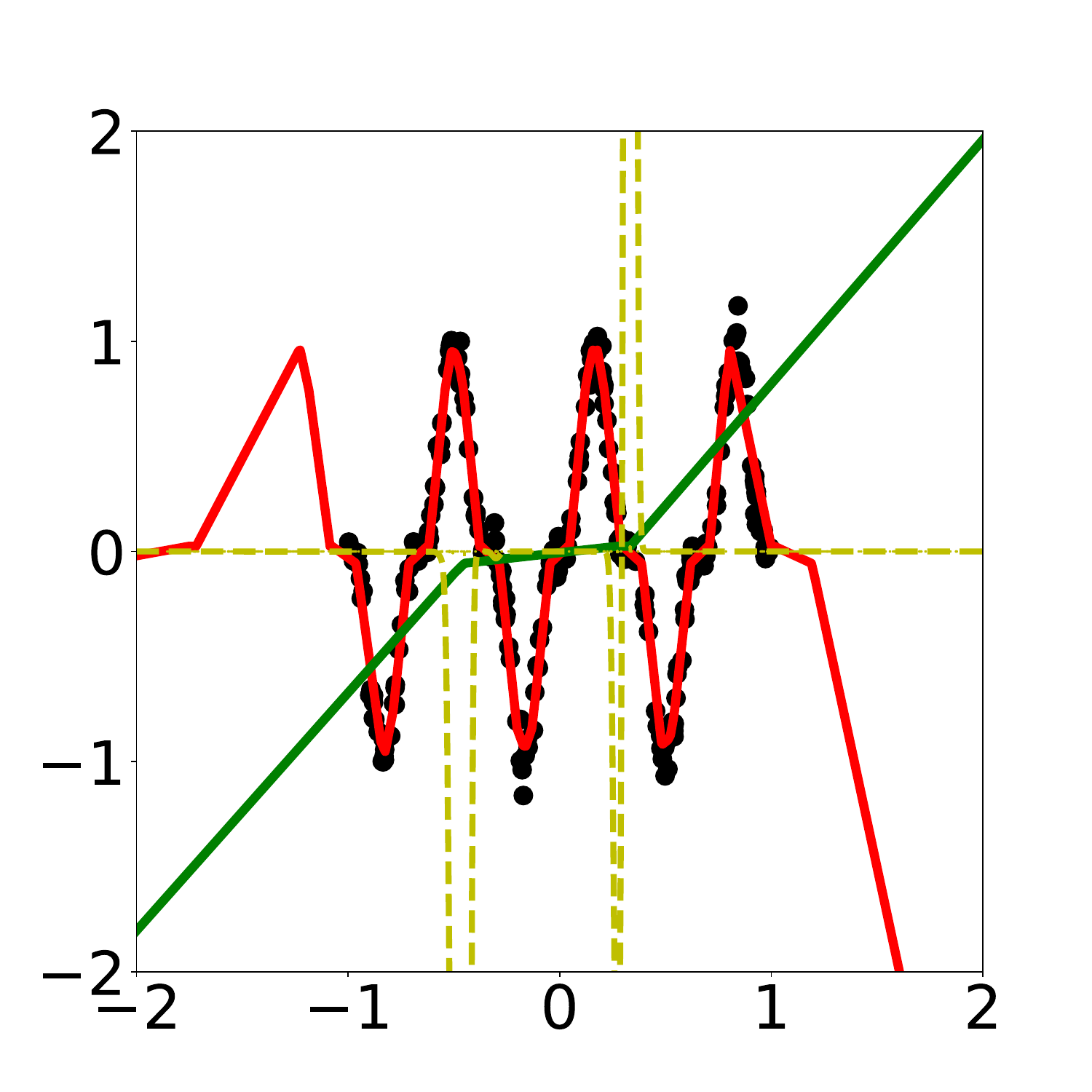}
         \caption{5\textsuperscript{th} component of third stack: ${(\NNj{3})}_5$ in green and $\hat{f}_5:={(\NN_\theta)}_5:={(\NNj{3})}_5\circ\NNj{2}\circ \NNj{1}$ in red and the training data points %
         $(\xtr_i,\ytr_{i})$ for $i\in I_5$
         as black dots.}
         \label{fig:cubic}
     \end{subfigure}
     \begin{subfigure}[b]{0.24\linewidth}
         \centering
         \includegraphics[width=\linewidth]{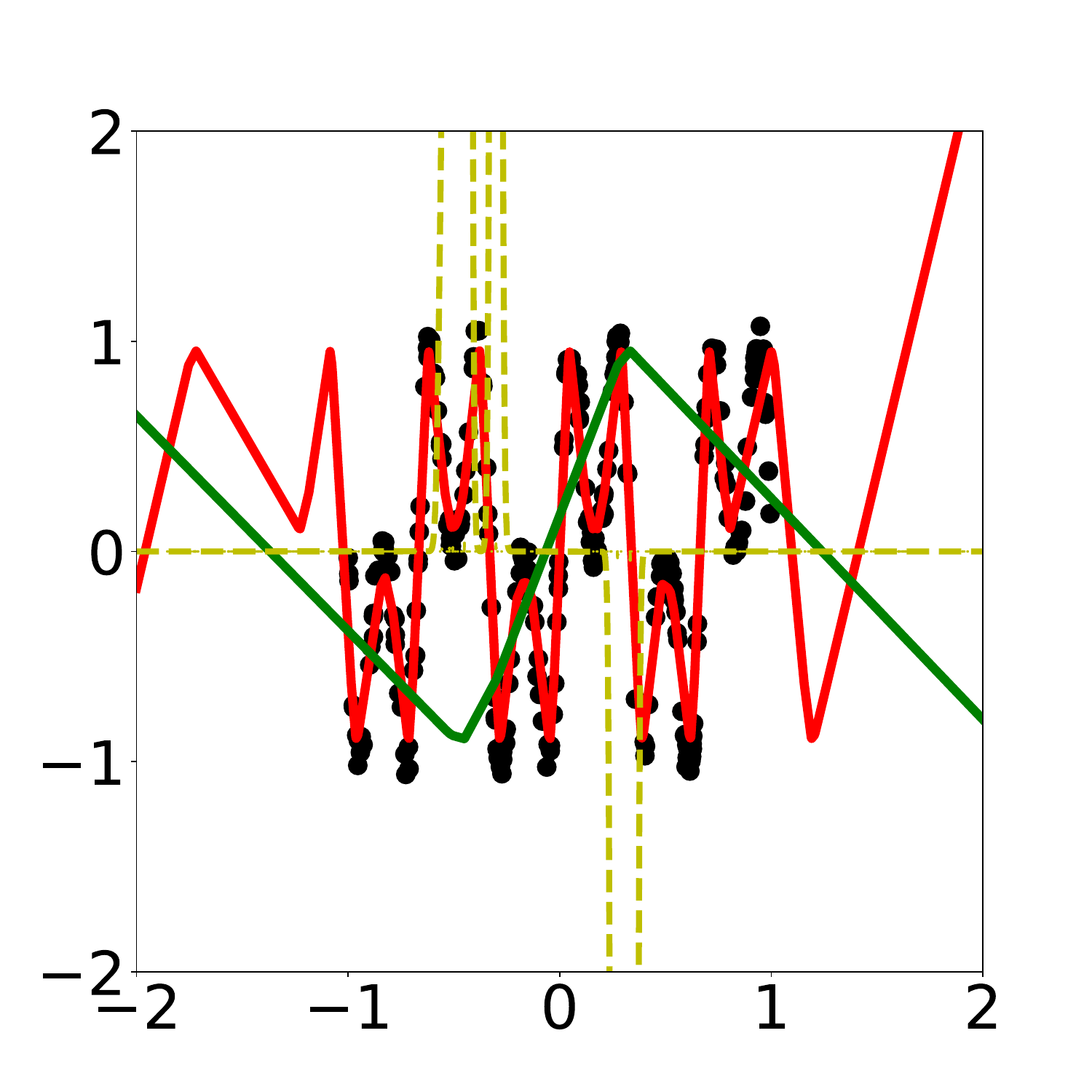}
         \caption{6\textsuperscript{th} component of third stack: ${(\NNj{3})}_6$ in green and $\hat{f}_6:={(\NN_\theta)}_6:={(\NNj{3})}_6\circ\NNj{2}\circ \NNj{1}$ in red and the training data points %
         $(\xtr_i,\ytr_{i})$ for $i\in I_6$
         as black dots.}
         \label{fig:sine}
     \end{subfigure}
     \begin{subfigure}[b]{0.24\textwidth}
         \centering
         \includegraphics[width=\textwidth]{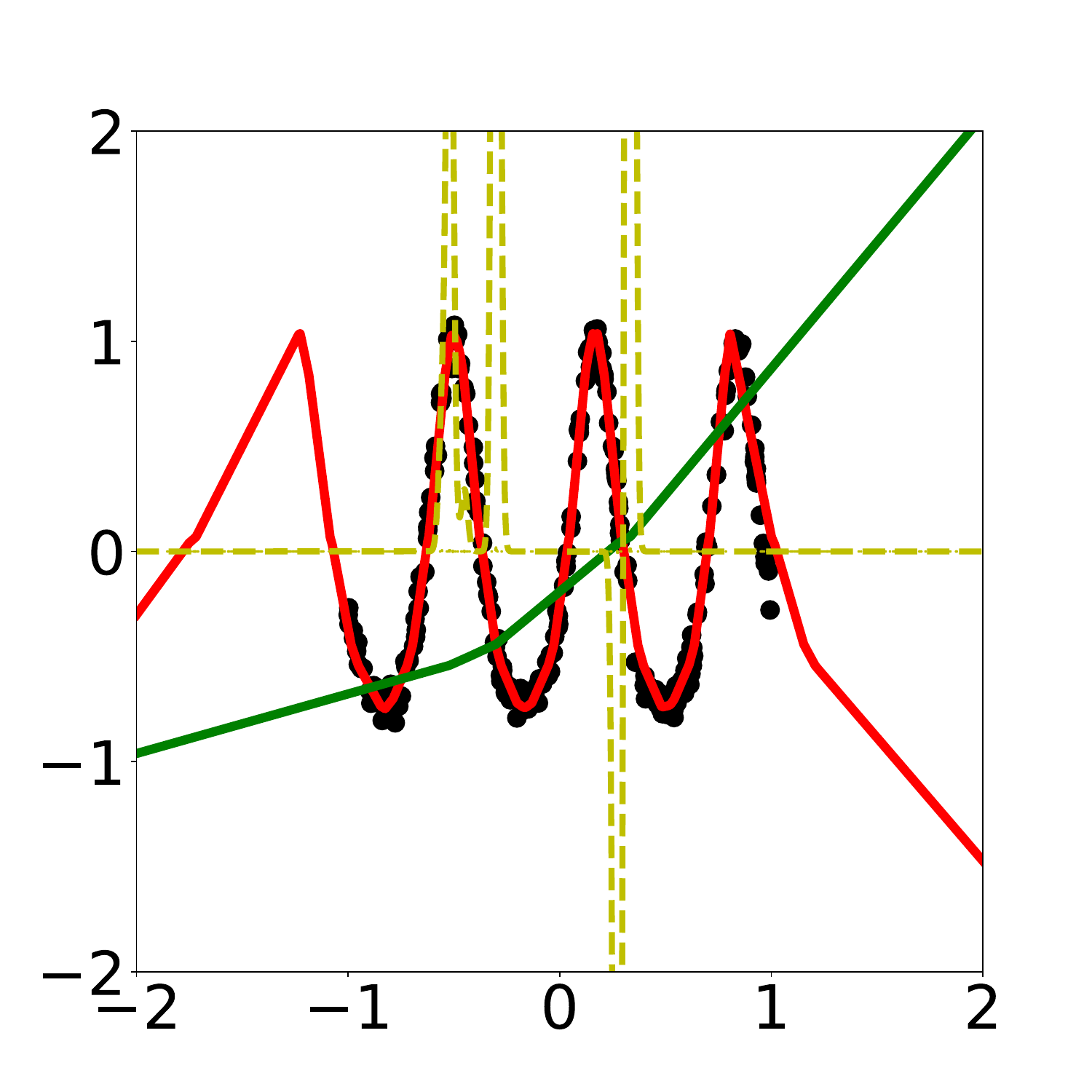}
         \caption{7\textsuperscript{th} component of third stack: ${(\NNj{3})}_7$ in green and $\hat{f}_7:={(\NN_\theta)}_7:={(\NNj{3})}_7\circ\NNj{2}\circ \NNj{1}$ in red and the training data points %
         $(\xtr_i,\ytr_{i})$ for $i\in I_7$
         as black dots.}
         \label{fig:exponential}
     \end{subfigure}
     \begin{subfigure}[b]{0.12\linewidth}
     \end{subfigure}
     \caption{The last stack can easily learn the different tasks based on the representation from the hidden representation. The hidden stacks learn a periodic representation $H=\NNj{2}\circ \NNj{1}$ (see \Cref{fig:VisualizePeriodicH}). The (distributional) second derivatives are visualized in yellow as in \Cref{fig:VisualizePeriodicH}.}\label{fig:VisualizePeriodicHOutput}
\end{figure}

In this section, we present a simple example of a data-generating function $f$ with $\dout=7$ outputs which are all periodic with the same periodicity $\frac{2}{3}$ and input dimension $\din=1$. The knowledge that the outputs are periodic is not given to the network a priori and we hope that the network with $\nStacks=3, n\gg N, d_1=d_2=1, \sigb={id}, \link^{-1}={id}$ (with architecture as in \Cref{fig:architectureDeepStackBasic}) is able to find a periodic representation $H$ by itself, as this would be helpful for all 7 outputs.
In all figures below, we show $\NNj{j}$ instead of $h_j$, since we have obtained $\theta$ from actually training a NN with gradient descent (that can get stuck in local minima) for a a finite time instead of calculating the perfect solution.
In \Cref{fig:VisualizePeriodicH,fig:VisualizePeriodicHOutput}, we visualize what each stack has learned: \Cref{fig:VisualizePeriodicH} visualizes how the hidden stacks learn a shared representation $H=\NNj{2}\circ \NNj{1}$ and \Cref{fig:VisualizePeriodicHOutput} visualizes how this can be useful for the different tasks.

In this case the benefits from multi-task learning are not too big, since all tasks share exactly the same input training data points. Nonetheless, there are still small benefits from multi-task learning in that the model can better filter out data noise.

\begin{figure}
    \centering
    \includegraphics[width=0.95\textwidth,trim={0, 0, 0, 1.36cm},clip]{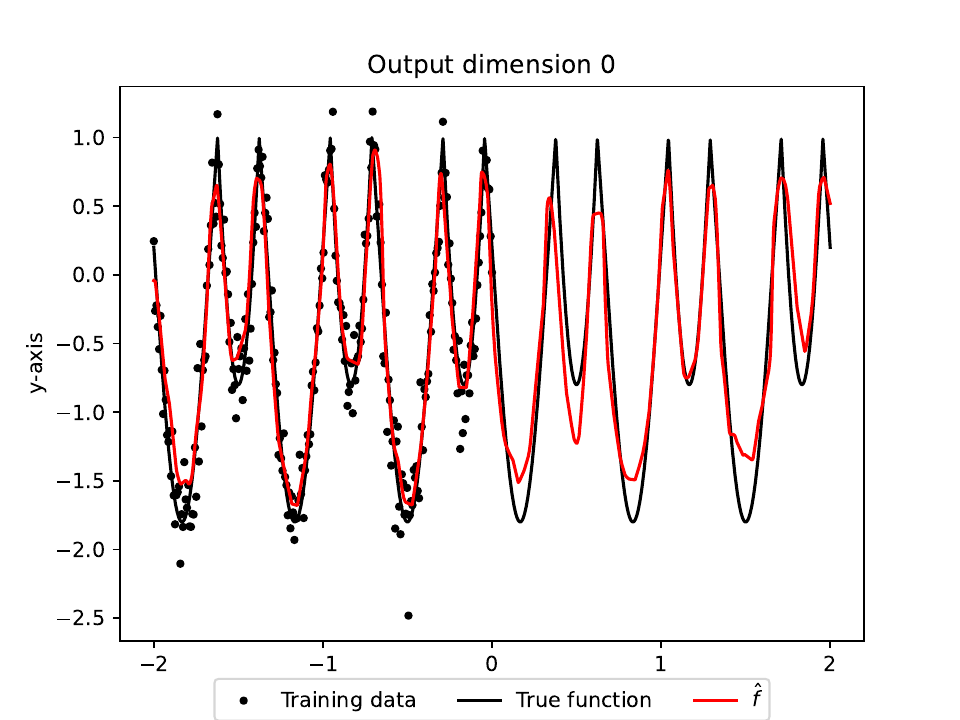}
    \caption{For the first task we only observe training input data $x\leq0$, but $\hat{f}_1:=(\NN_\theta)_1$ learns to extrapolate well to $x>0$, because of the multi-task benefits of being trained together with 6 other tasks (see \Cref{fig:amazingMultiTaskLearningByPOtherTasks}).}
    \label{fig:amazingMultiTaskLearningByP}
\end{figure}

\begin{figure}
     \centering
     \begin{subfigure}[b]{0.3\linewidth}
         \centering
         \includegraphics[width=\textwidth, trim={0, 0, 0, 1.36cm},clip]{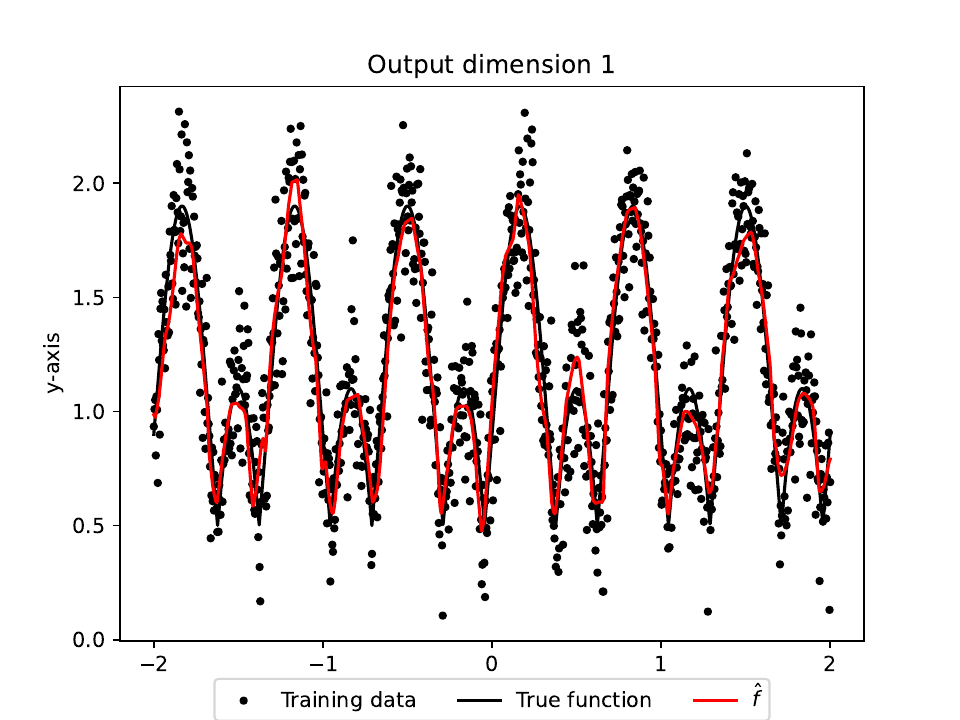}
         \caption{$\hat{f}_2:={(\NN_\theta)}_2$}
         \label{fig:absolute_valueHelpAmazing}
     \end{subfigure}
     \hfill
     \begin{subfigure}[b]{0.3\linewidth}
         \centering
         \includegraphics[width=\textwidth, trim={0, 0, 0, 1.36cm},clip]{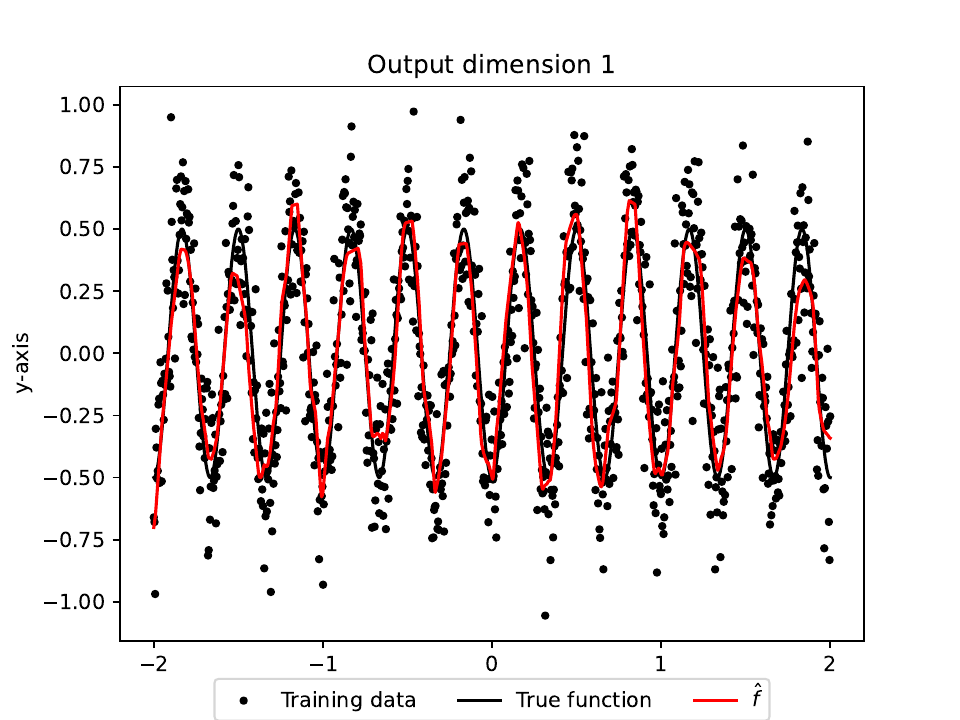}
         \caption{$\hat{f}_3:={(\NN_\theta)}_3$}
         \label{fig:squareHelpAmazing}
     \end{subfigure}
     \hfill
     \begin{subfigure}[b]{0.3\linewidth}
         \centering
         \includegraphics[width=\textwidth, trim={0, 0, 0, 1.36cm},clip]{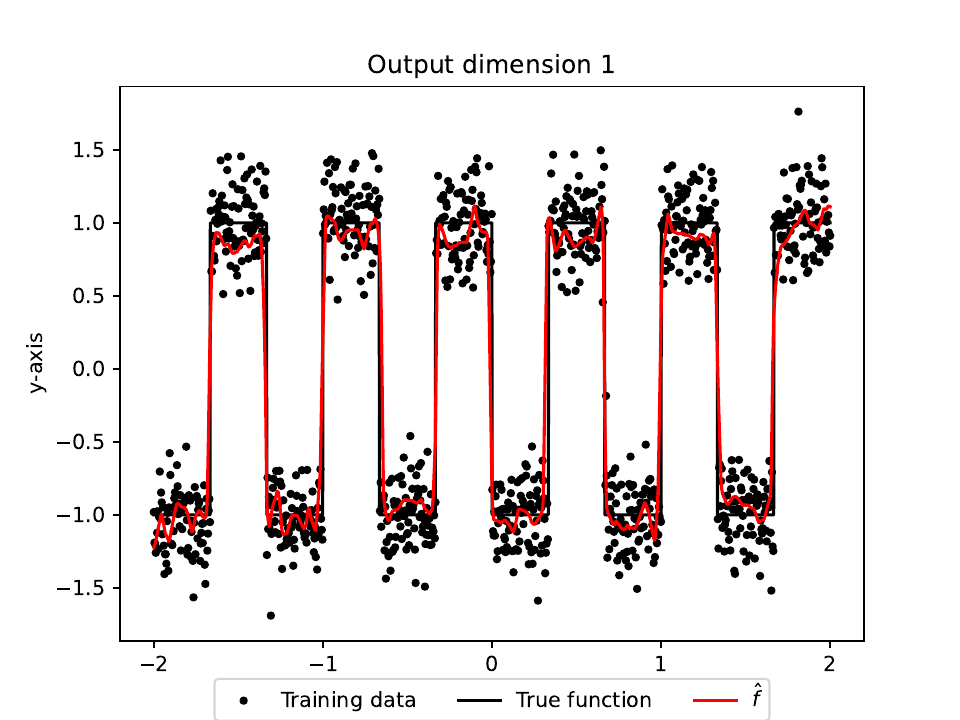}
         \caption{$\hat{f}_4:={(\NN_\theta)}_4$}
         \label{fig:signHelpAmazing}
     \end{subfigure}

     \begin{subfigure}[b]{0.32\linewidth}
         \centering
         \includegraphics[width=\textwidth, trim={0, 0, 0, 1.36cm},clip]{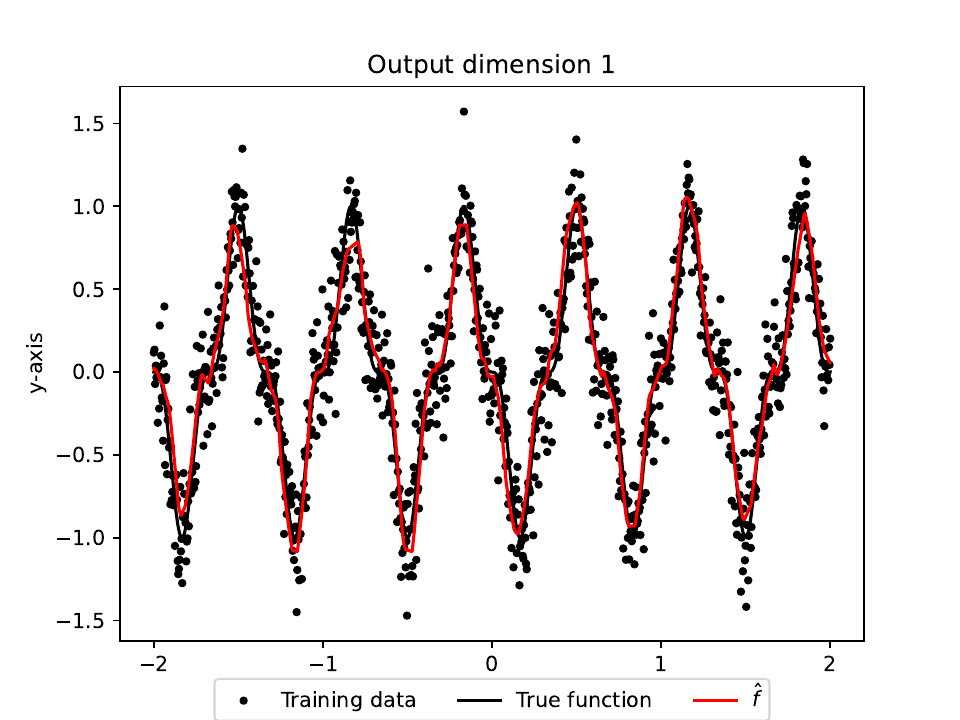}
         \caption{$\hat{f}_5:={(\NN_\theta)}_5$}
         \label{fig:cubicHelpAmazing}
     \end{subfigure}
     \hfill
     \begin{subfigure}[b]{0.32\linewidth}
         \centering
         \includegraphics[width=\textwidth, trim={0, 0, 0, 1.36cm},clip]{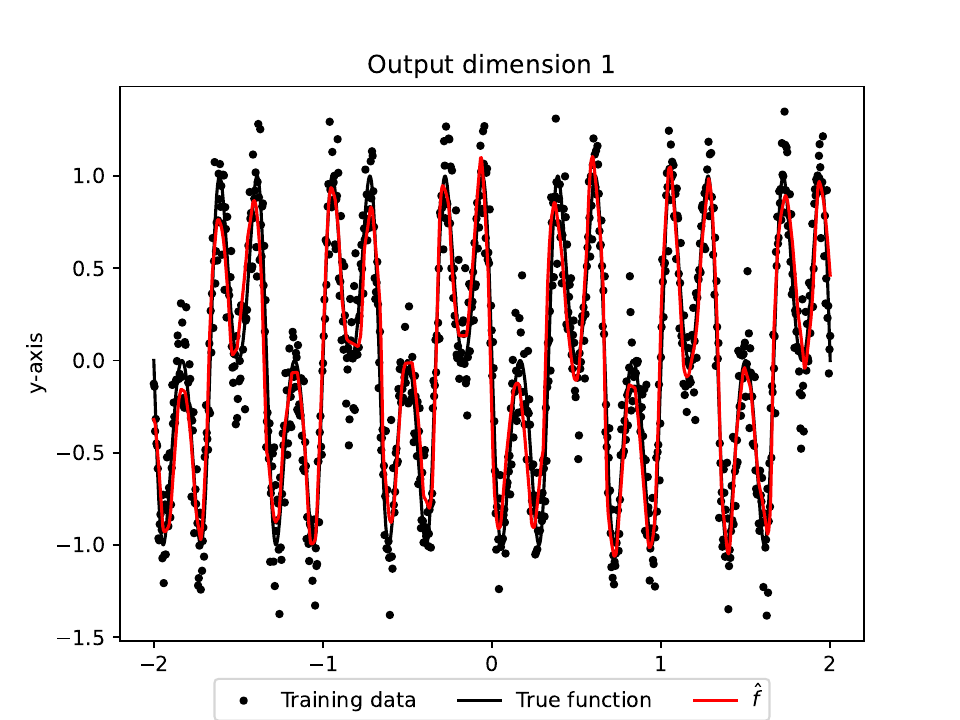}
         \caption{$\hat{f}_6:={(\NN_\theta)}_6$}
         \label{fig:sineHelpAmazing}
     \end{subfigure}
     \hfill
     \begin{subfigure}[b]{0.32\linewidth}
         \centering
         \includegraphics[width=\textwidth, trim={0, 0, 0, 1.36cm},clip]{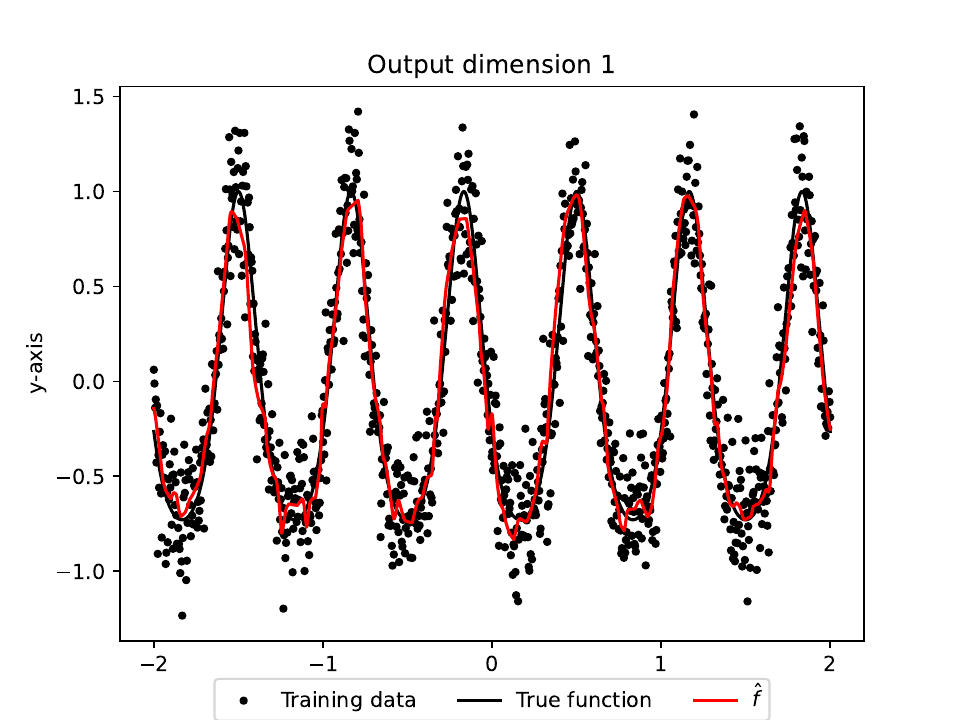}
         \caption{$\hat{f}_7:={(\NN_\theta)}_7$}
         \label{fig:exponentialHelpAmazing}
     \end{subfigure}
     \begin{subfigure}[b]{0.12\linewidth}
     \end{subfigure}
     \caption{The last stack can easily learn the different tasks based on the representation from the hidden representation $H$. The hidden stacks learn a periodic representation $H=\NNj{2}\circ \NNj{1}$.}\label{fig:amazingMultiTaskLearningByPOtherTasks}
\end{figure}

Next we will show an example for the same 7 tasks but with a different training set. This time we have less data for the first task -- in particular we only have training data for $x\in[-2,0]$ for the first task, while for the other 6 tasks we also have $y$ values corresponding to input training data points $x\in[-2,2]$. If one were to train the first task separately only on the input training data $x<0$, it would be extremely hard for the neural network to extrapolate to input testing data $x>0$, but with the help of the multi-task learning induced by $P$, the network can extrapolate very well to input testing data $x>0$ as shown in \Cref{fig:amazingMultiTaskLearningByP}. Note that for this experiment we use wide bottlenecks, i.e.~$d_1=d_2=2048$, to also experimentally validate \Cref{cor:wide_dj} by showing that multi-task learning does not vanish for large bottleneck dimensions.
The detailed experimental setup is $\nStacks=3$, $n_1=n_2=n_3=2048,\lw=0.0005, \sigb=\text{ReLU}, \link^{-1}={id}$. Note that  The training data and the learned functions for the 6 tasks that have training input data $x\in[-2,2]$ are shown in \Cref{fig:amazingMultiTaskLearningByPOtherTasks}. For the plots we used a neural network $\NN_\theta$ which was trained with gradient descent on the objective~\eqref{eq:optimalParameters}, which is probably not a perfect approximation of $\NN_{\theta^{*,\lw}}$, but good enough to show the qualitative behavior of $f^{*,\lambda}$. (For high dimensional $d_1=d_2=2048$ the learned hidden representation cannot be visualized as easily as for the previous experiment. This is why we do not have an analogous plot to \Cref{fig:VisualizePeriodicH} for this experiment.)


\begin{figure}[htb]
    \centering
    \includegraphics[width=0.95\textwidth, trim={0, 0, 0, 1.36cm},clip]{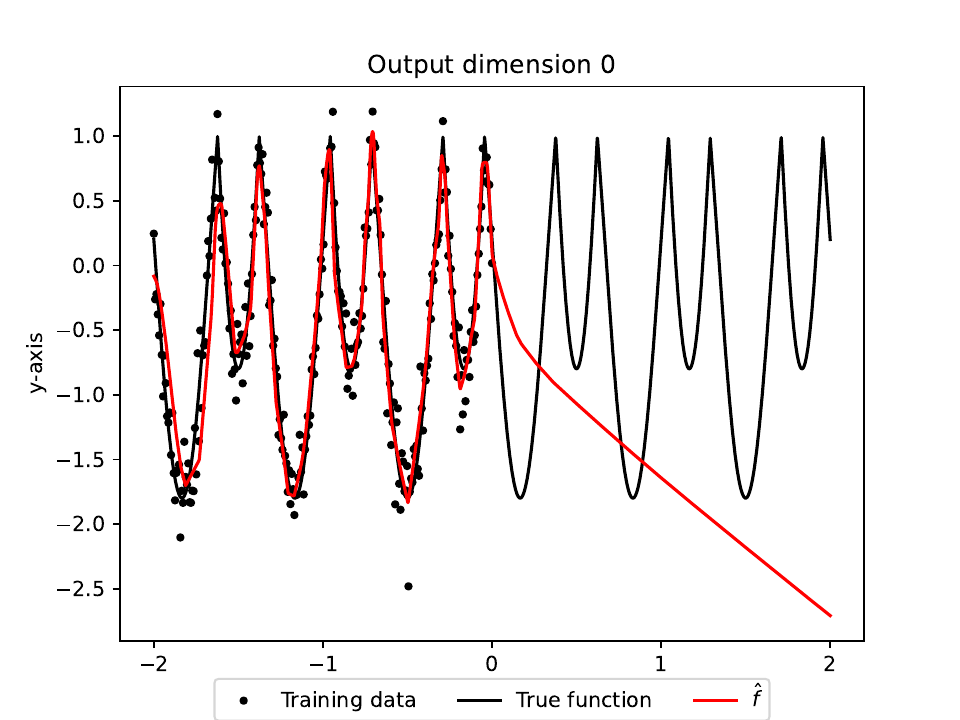}
    \caption{$\hat{f}_1:=(\NN_\theta)_1$ without joint training data for the other 6 outputs cannot extrapolate very well. (Also for $x\leq0$ it is harder to filter out the noise without the other tasks.)}
    \label{fig:trainedAlone}
\end{figure}

Without multi-task learning neural networks would not extrapolate the function periodically. If one would train the same model only on the first task without any additional task, the model would just extrapolate with as little second derivative as possible as can be seen (approximately again because of gradient descent) in \Cref{fig:trainedAlone}.

\section{Discussion of Multi-Task Learning and \Cref{def:NoMultiTask}}\label{sec:DiscussBenefitsMultiTask}
Mathematically one cannot judge a priori if multi-task learning is beneficial or not, because this depends on whether or not one's prior belief (or inductive bias) matches with the true prior.\footnote{If the data comes from a function that was actually sampled from a GP with independent outputs, then multi-task-learning will not be beneficial.}

One can however very precisely define a notion capturing that learning multiple tasks at once does not result in different estimators than when learning them separately (see \Cref{def:NoMultiTask}). And if learning different tasks jointly does not make any difference (in terms of generalization), one can obviously not benefit (in terms of generalization) from learning them jointly.
If however, the generalization behaviours of different outputs trained jointly to perform different tasks influence each other, these outputs can also benefit from it depending on one's prior belief.

Some intuition for what kind of influence among tasks is desirable for real-world applications according to a common sense prior is given in \citet{MultitaskCaruana1997}.
Intuitively, it is not beneficial if different outputs just influence each other randomly. With our formulation of $\Pfunc$, the discussion in \cref{sec:DeepVsShallow} and the visualizations in \Cref{sec:VisualizingMultiTask}, we see that this influence is \emph{not} random for $\ell^2$-regularized NNs (or $\mathcal{M}^{\Pfunc}$). Instead, training NNs with weight-decay enforces multi-task learning due to the shared representation $H$, which corresponds to the intuitively beneficial multi-task learning described in \citet{MultitaskCaruana1997}.
Moreover, from an empirical point of view, in various applications, training multiple tasks jointly has outperformed separate training of the individuals tasks \citep{MultitaskCaruana1997,ruder2017overviewMultiTask,fifty2021efficiently,tran2021facebook,aribandi2021ext5ExtremeMultiTask}.

\subsection{Connection to Representation Learning, Feature Learning, Metric Learning and Transfer Learning}
In this subsection we will explain that multi-task learning, representation learning, feature learning, metric learning and transfer learning are highly related phenomena.
In this paper we focused on multi-task learning, because the absence of multi-task learning can be rigorously defined for models~$\mathcal{M}$ in functions space (see \Cref{def:NoMultiTask}).

But $\Pfunc$ defined in \cref{eq:P,eq:Pj} also intuitively reveals the ability of wide deep $\ell^2$-regularized NNs to perform representation learning, feature learning, metric learning and transfer learning.
For example\footnote{%
A reasonable alternative point of view would be to see $\tilde{H}(x)=\left(\max(0,b_k^{(\nStacks)}+\langle v_k^{(\nStacks)},H(x)\rangle\right)_{k\in\fromto{n_\nStacks}}$ as the learned representation of $x$. The representation $\tilde{H}(x)$ captures the learned hidden representation up to the last hidden layer, but $H$ can be studied more conveniently on function space.%
}, we can interpret $H(x)=\sigb\circ h_{\nStacks-1}\circ\dots\circ\sigb\circ h_{1} (x)$ as the learned representation of $x$. Then we can interpret the coordinates $H_k(x)$ as the learned features of $x$. Further we can interpret $d_H(x,\tilde{x}):=\twonorm[H(x)-H(\tilde{x})]$ as the learned metric on the input space $\R^\din$.
Since the hidden representation $H$ is shared among all tasks, it enforces multi-tasking.
If one does not train all task simultaneously, one can first train some tasks to obtain $H$ and then afterwards train further tasks with fixed $H$ (or by only slightly varying $H$) to achieve transfer learning.

Multi-task learning, representation learning, feature learning, metric learning and transfer learning can not be achieved via Gaussian process regression with a fixed kernel.

\subsection{Adaptivity vs Multi-task Learning}\label{sec:AdaptivityVsMTL}
\citet{Bach2017BreakingCurseDimensionalityConvexNN,Chizat2020ImplicitBias} have shown that wide $\ell^2$-regularized NNs with one hidden layer are adaptive to hidden linear low-dimensional structures in terms of generalization. This means: If the unknown true function $f$ can be expressed as $f(x)=\tilde{f}(Vx)$, where $V\in\R^{r\times\din}$ is a projection matrix on a $r$-dimensional subspace, then the generalization bound of the NN asymptotically only depends on $r$ instead of $\din$ under mild smoothness conditions on $f$. In other words, this means that the curse of dimensionality can be avoided (asymptotically) if $r\ll \din$.
The same holds true, if they chose the first layer weights~$v$ and biases~$b$ randomly\footnote{We assume that the support of the distribution of $(v,b)$ is $\Sd[\din]$, e.g., $(v,b)\sim\Unif[{\Sd[\din]}]$ can be distributed uniformly on the sphere $\Sd[\din]$.} and only train the output weights~$w$ and biases~$c$ with $\ell^1$-regularization. In function space training both layers with $\ell^2$-regularization corresponds to $\Pfunc=P_1$ introduced in \cref{eq:Pj}\footnote{%
If we have only 1 hidden layer (i.e.,  $\nStacks=1$) and the identity as final non-linearity~$\ell^{-1}=\text{id}$, then $\Pfunc$ from \cref{eq:P} is obviously equal to $P_1$ from \cref{eq:Pj}.%
}, while only training the output layer with $\ell^1$-regularization corresponds to\footnote{For \cref{eq:PL1l1} we use the same abuse of notation as for \cref{eq:Pj} (treating distributions as if they were functions).}
\begin{equation}\label{eq:PL1l1}
\tilde{P}^{\ell^1}(f):=\min_{\substack{\varphi\in\T[1],\ c\in\R^{\dout}\text{ s.t.}\\  f=\int_{\Sd[\din-1]}\psr[\langle s,\cdot\rangle]{s}\, ds+c}} \left(
	2\int_{\Sd[\din-1]}\int_{\R} \frac{\left\| {\psr{s}}^{''} \right\|_1}{g(r)} \dx[r]\dx[s] +2\left\|c\right\|_1		\right),%
\end{equation}
where the main difference between $\tilde{P}^{\ell^1}$ and $\Pfunc$ is that the norm inside the inner integral is an $\ell^1$-norm instead of an Euclidean $\ell^2$-norm. In other words, we have an $L^1\left(\R ; (\R^\dout,\|\cdot\|_1\right)$-norm for $\tilde{P}^{\ell^1}$ instead of the $L^1\left(\R ; (\R^\dout,\|\cdot\|_2\right)$-norm that we had for $\Pfunc$.
As we have an $L^1$-norm in both cases, the models
$\mathcal{M}^{\tilde{P}^{\ell^1}},\Ltr\mapsto\mathcal{M}^{\tilde{P}^{\ell^1}}(\Ltr):=\argmin_f \left(\Ltr(f) +\lambda\tilde{P}^{\ell^1}(f)\right)$
and $\mathcal{M}^{\Pfunc}$ both are adaptive to hidden linear low-dimensional structures in terms of generalization.

However, while $\mathcal{M}^{\Pfunc}$ can benefit from multi-task learning (see \Cref{prop:MultiTask}), $\mathcal{M}^{\tilde{P}^{\ell^1}}$ cannot benefit from multi-task learning as the following proposition shows.

\begin{proposition}\label{prop:L1l1NoMultiTask}
The model $\mathcal{M}^{\tilde{P}^{\ell^1}}$ (where $\mathcal{M}^{\tilde{P}^{\ell^1}}(\Ltr):=\argmin_f \left(\Ltr(f) + \lambda \tilde{P}^{\ell^1}(f)\right)$ is defined as above based on \cref{eq:PL1l1}) is not capable of benefiting from multi-task learning (see \Cref{def:NoMultiTask}). 
\end{proposition}
\begin{proof}\label{proof:prop:L1l1NoMultiTask}
The proof is trivial, since $\tilde{P}^{\ell^1}$ splits into a sum such as in the \hyperref[proof:prop:RSNNoMultiTask]{proof} of \Cref{prop:RSNNoMultiTask}:
 Without the square-root, learning a separate function $f_k$ for each task would result exactly in the same functions $f_k$ as training them all together%
, since \[\int\onenorm[f^{''}(x)] \,dx
=\int\sum_{k=1}^{\dout}\left|f_k^{''}(x)\right| \,dx
=\sum_{k=1}^{\dout}\int\left|f_k^{''}(x)\right| \,dx.\]
Therefore,
\begin{multline*}
\Ltrb{f} + \lambda \tilde{P}^{\ell^1}(f)=\\
=\sum_{k=1}^{\dout}\left(\Ltr_k(f) + 
\lambda\min_{\substack{\varphi\in\hyperlink{eq:Tdin1}{\mathcal{T}^{\din;1}},\ c\in\R^{1}\text{ s.t.}\\  f_k=\int_{\Sd[\din-1]}\psr[\langle s,\cdot\rangle]{s}\, ds+c}} \left(
	2\int_{\Sd[\din-1]}\int_{\R} \frac{\left| {\psr{s}}^{''} \right|}{g(r)} \dx[r]\dx[s] +2\left|c\right|		\right)
\right),
\end{multline*}
where $\mathcal{T}^{\din;1}$ is defined as $\T[1]$ with the only difference that $d_1$ is replaced by 1.\footnote{%
Following this notation $\T=\hyperlink{eq:Tdin1}{\mathcal{T}^{d_{j-1};d_j}}$ and $\hyperlink{eq:Tdin1}{\mathcal{T}^{\din;1}}$ is defined as
\hypertarget{eq:Tdin1}{\begin{align*}
	\hyperlink{eq:Tdin1}{\mathcal{T}^{\din;1}}:=\bigg\{ \varphi=(\varphi_s)_{s\in\Sd[\din-1]}  &\bigg|\, \forall s\in\Sd[d_{j-1}-1] :\varphi_s:\R\to\R^1,\lim_{r\to -\infty} \psr{s}=0 = \lim_{r\to -\infty} \frac{\partial}{\partial r}\psr{s}%
	\bigg\}.
\end{align*}}%
}
\[
\left(\mathcal{M}^{\tilde{P}^{\ell^1}}(\Ltr)\right)_k
=\left(\argmin_f \Big(\Ltrb{f} + \lambda \tilde{P}^{\ell^1}(f)\Big)\right)_k
=\left(\mathcal{M}^{\tilde{P}^{\ell^1}}(\Ltr_k)\right)_k.
\]
(Note that all its other components of $\mathcal{M}^{\tilde{P}^{\ell^1}}(\Ltr_k)$ are the zero function.)
\end{proof}
So we can see that adaptivity to hidden linear low-dimensional structures in terms of generalization (as introduced by \citet{QuestAdaptivityBlogPostBach,Bach2017BreakingCurseDimensionalityConvexNN,Chizat2020ImplicitBias}) does not imply the ability for benefiting from multi-task learning, representation learning, feature learning, metric learning or transfer learning.

Therefore, the ability to benefit from multi-task learning can be seen as an extension/continuation of \enquote{quest for adaptivity} by \citet{QuestAdaptivityBlogPostBach}.

\subsection{Universal Features and Universal \enquote{Readout}}\label{sec:UniversalFeaturesUniversalReadout} 
In terms of universality, there are two aspects in which deep stacked NNs (i.e., $\nStacks>1$) differs from the shallow NNs (i.e., $\nStacks=1$).

First, for $\nStacks>1$ the representation $H(x)$ can approximate %
any (continuous) function in $x$ by universal approximation theorems \citep{HornikUniversalApprox1991251,CybenkoUniversalApprox1989}.
This is important since in applications of modern deep learning, representations $H(x)$ shared among different tasks can be a highly non-linear functions of $x$. In \Cref{fig:second_stack}, we see such a non-trivial representation $H$ learned by a neural network. A shallow ReLU NN however cannot learn such a representation within its \emph{shared} hidden layer.

Second, for $\nStacks>1$ the last stack $h_{\nStacks}$ can be seen as a universal non-linear readout. A shallow NN only consists of one hidden layer that learns the shared representation, and one linear task-specific readout layer. In contrast, deep NNs are able to learn non-linear task-specific readout functions $h_{\nStacks},k$ for the different tasks (see \Cref{fig:VisualizePeriodicHOutput}). This is not possible for methods such as linear group-Lasso regression (on non-linear fixed features). The latter we refer to as \enquote{shallow multi-task learning}. 

For the generalization behavior shown in \Cref{fig:signHelpAmazing} of \Cref{sec:VisualizingMultiTask} both universal representation and universal readout is needed, since seven components of the function $f$ are very different highly non-linear functions composed with the highly non-linear sine function. Thus, in this example, the most important shared feature should be somehow related to a sine function and different non-linear readouts are needed for the different tasks in order to obtain such strong benefits from multi-task learning as in \Cref{fig:signHelpAmazing}.
There is no single shared feature, such that all functions could be represented as linear functions of this shared feature.

\subsection{Different Levels of Abstraction}\label{sec:LevelsOfAbstraction} In practice it is often observed that low level features (such as edges for images) are important across many different tasks (for example even tasks based on different types of images, such as natural images and X-ray images share edges as important features). According to our formulation of $\Pfunc$ for $\nStacks>1$, a first stack $h_1$ is preferred that mainly learns features that can be shared among as many tasks as possible instead of learning different features for different tasks in order to keep $\Pfunc_1(h_1)$ as low as possible even when $d_1\gg1$ is very large. Then medium-low level features $h_2\circ h_1$ share the property that they can be represented by a rather \enquote{flat} (in terms of $\Pfunc_2$) function $h_2$ on top of the shared low-level features $h_1$. $\Pfunc$ still prefers to learn features $h_2\circ h_1$ which are helpful\footnote{Mathematically speaking a feature $ (h_j\circ\dots\circ\sigb\circ h_1)_k$ is not helpful (or irrelevant) for the $l$-th task if the function $(\link^{-1}\circ h_{\nStacks}\circ\dots\circ\sigb\circ h_{j+1})_l$ is constant with respect to its $k$-th input dimension. (In practice it is more a continuum of helpfulness/relevance if the function is almost constant with respect to certain input dimensions.)} for all tasks. However, depending on how unrelated the tasks are, also some task-specific features can be learned for the cost of of higher regularisation costs. And as the stack-index $j$ increases, more of the higher-level features $(h_j\circ\dots\circ\sigb\circ h_1)_k$ become mainly relevant for smaller subsets of tasks and only f of the high level features are relevant for all tasks (depending on how related they are). A similar effect has been studied empirically in \citet{NEURIPS2020NeyshaburTransferLearning}, where the performance on one task was improved by reusing features learned from a different task, where especially in the first hidden layer a lot of features could be reused and the number of features that could be reused decreased from low level layers to high level layers.

See \Cref{tab:MTLdifferentMethods} in \Cref{sec:DifferentMTLRegularization} for an overview which methods allows for \enquote{deep} multi-task learning and which do not.

\subsection{On Multi-task Learning Induced by Regularization}\label{sec:DifferentMTLRegularization}

In this paper, we show that the function space regularization $\Pfunc$ induced by $\ell_2$-regularization on the parameters of 
NNs enables them to benefit from multi-task learning. We like to highlight that the induced regularization does not involve the $L^2$-norm in function space. We show that the $\ell_2$-regularization in parameter space induces an $L^1(\ell_2)$-regularization $P_1$ of the second derivative in function space in the case of 1 hidden layer. (Here, we call a norm $L^p(\ell_q)$-typed if inside the integral there is a $\lVert\cdot\rVert_q^p$-term. This can be seen as a $L^p\left(\R ; (\R^\dout,\|\cdot\|_q)\right)$-norm as in \Cref{sec:AdaptivityVsMTL}.)

The $L^1(\ell_2)$-regularization is able to benefit from multi-task learning (see \Cref{prop:MultiTask}). This is in contrast to the regularizations
$L^1(\ell_1)$ (see eq. \eqref{eq:PL1l1}) and $L^2(\ell_2)$, which are both not able to benefit from multi-task learning (see \Cref{prop:RSNNoMultiTask,prop:L1l1NoMultiTask}). Note also that for deep NNs, $P$ (see eq. \eqref{eq:P}) is more intricate than $P_j$ (see eq. \eqref{eq:Pj}). In contrast to $P_j$, $P$ is not a norm anymore, and thus $P$ promotes much more interesting forms of \enquote{deep} multi-task learning (see \Cref{sec:UniversalFeaturesUniversalReadout,sec:LevelsOfAbstraction}).

To highlight once more the differences between various models in terms of multi-task learning and regularization in function space, we have compiled \Cref{tab:MTLdifferentMethods}.

\begin{table}[]
    \centering
    \begin{large}
    \resizebox{1\textwidth}{!}{
    \begin{tabular}{|l|l|l|l|}
    \hline
    
        \multicolumn{2}{|c|}{\bf Model} &  \multicolumn{1}{|c|}{\bf Multi-task Learning} &  \multicolumn{1}{|c|}{\bf Inductive Bias in Function Space} \\
            \hline
          \multicolumn{2}{|l|}{\makecell{ Kernel ridge regression with fixed kernel \\(e.g., NTK, Gaussian BNN kernel, random feature kernel)} }&  No (\Cref{prop:GPnoMultiTask,prop:RSNNoMultiTask}) & $L^2(\ell_2)$-typed regularization\\
        \hline
                 \multicolumn{2}{|l|}{ \makecell{ Kernel lasso regression with fixed kernel \\(i.e., linear $\ell_1$ regression)}} &  No (\Cref{prop:L1l1NoMultiTask}) & $L^1(\ell_1)$-typed regularization\\
        \hline
                          \multicolumn{2}{|l|}{\makecell{Kernel group-lasso regression with fixed kernel \\(i.e., linear group-lasso regression)}} &  ``shallow''  & $L^1(\ell_2)$-typed regularization\\
        \hline
        
                        \multirow{3}{*}{\makecell{
                                $\NN_{\theta^{*,\lw}}$  \\
                               (i.e., our setting) 
                            }}& 
                               \makecell[l]{ one hidden layer \\(i.e., $\nStacks=1$)} 
                                
                            & ``shallow'' & \makecell[l]{Regularization functional $P_1$ \\given in eq. \eqref{eq:Pj}
                            ($L^1(\ell_2)$-typed)
}\\
                            &\makecell[l]{multiple hidden layers \\(i.e., $\nStacks>1$)} &``deep'' &  \makecell[l]{Regularization functional $P$ \\
                            given in eq. \eqref{eq:P}}\\
        \hline
        \multicolumn{2}{|l|}{ \makecell[l]{Multiplying a hard-coded, non-trainable matrix
        on the\\ outputs of a model (that itself is not able to benefit from \\ multi-task learning)}} &\makecell[l]{Weak form of ``shallow''\\ multi-task learning}&  \makecell[l]{Handcrafted inductive bias\\ favoring a certain sign of\\ correlation}\\
        \hline
    \end{tabular}
    }
    \end{large}
    \caption{Various infinite with limits, their inductive bias in function space and their ability to benefit from multi-task learning.}
    \label{tab:MTLdifferentMethods}
\end{table}

\section{Proofs}\label{sec:proofs}

See \Cref{def:BV2,def:BV2loc,def:BV20-} for definitions of the function spaces (such as $\BVTwoZ$) that we consider.

\begin{lemma}
The normed vector space $\BVTwoZ(\R;\R^{d_j})$ is a Banach space.
\end{lemma}
\begin{proof}
 It suffices to consider the case $d=1$. %
 Completeness of $\BVTwoZ(\R;\R)$ can be seen through the following inequalities: firstly, we observe that
\begin{equation}\label{eq:poincare4firstderiv}
\sup_{r \in \R} | f'(r) | \leq |D^2 f |_{\frac{1}{g}}
\end{equation}
by the fact that $ \frac{1}{g} \geq 1$ and the $f'(-\infty)=0$.

Secondly, it follows from integration by parts that
\begin{subequations}\label{eq_PoicareTypedBV20-}
\begin{align}
|f(c)|= \left|\int_{-\infty}^c f'(a) da\right| &=\left|\int_{-\infty}^c D^2f((-\infty,a]) da\right|
\\
&=\left|\int_{-\infty}^c (c-a)dD^2f(a)\right| \\
&\leq\Bigg| D^2f|_{(-\infty,c]} \Bigg|_{|c-(\cdot)|}\\
&\leq
\Bigg| D^2f|_{(-\infty,c]} \Bigg|_{\frac{1+\max(0,c)}{g}}
\leq
\Bigg| D^2f \Bigg|_{\frac{1}{g}}(1+\max(0,c)),
\end{align}
\end{subequations}
where we used that $|c-(\cdot)|\leq\frac{1+\max(0,c)}{g}$.
Analogously, one can derive the same bound for $\left\|f'|_{(-\infty,c]}\right\|_{L^1}$.
Obviously, we have for $ a \leq b \leq 0$ that
\[
|f(b)-f(a)| \leq \int_{a}^b |f'(r)|dr .
\]
Therefore, we can conclude that a Cauchy sequence $(f_n)$ in $ \BVTwoZ(\R;\R) $ defines a limiting measure of bounded variation $ \mu$, a limiting function $u$ being $L^1$ limit of $(f'_n)$ on any interval $(-\infty,a]$ for any real $a$ and a function $h$ being the limit of $(f_n)$ uniformly on $(-\infty,a]$ for any real $a$. Whence there exists $f \in W_\text{loc}^{1,1}(\R;\R)$ such that $f=h$, $f'=u$ and $D^2f = \mu$. Additionally, $h(-\infty)=0$ (because of \meqref{eq_PoicareTypedBV20-}) and $f'(-\infty)=0$ (by \meqref{eq:poincare4firstderiv}) hold true as well as $ |D^2 f |_{\frac{1}{g}} < \infty$.
\end{proof}

\begin{definition}\label{def:FunctionMasureSets}
Let $j\in\{1,\dots,\nStacks\}$, $\mathfrak{M}_j:=\mathfrak{M}(K_j)\times\R^{d_j}$ with $K_j:=\Sd[d_{j-1}]\times\Sd[d_{j}-1]$. We then define
\begin{enumerate}
    \item the set of functions
    \begin{align*}
    \mathcal{F}_{K_j}:=\bigg\{&f:\R^{d_{j-1}}\rightarrow\R^{d_j}\,\Big|\,\exists\,(\nu,c)\in\mathfrak{M}_j\ \forall\,x\in\R^{d_{j-1}}:\\ &f(x)=c+\int_{K_j}w\,\sigma(\langle v, x\rangle-r)\,d\nu((v,r),w)\bigg\},
    \end{align*}
    \item and the set of function representations
    \begin{align*}
    \mathfrak{C}_{K_j}^{f}:=\bigg\{ &(\mu,c)\in \mathfrak{M}_j\,\Big|\,\forall\, x\in\R^{d_{j-1}}:\\ &f(x)=c+\int_{K_j} w\,\sigma(\langle v, x\rangle-r)\,d\mu((v,r),w)\bigg\}
    \end{align*}
    of a given $f\in\mathcal{F}_{K_j}$.
\end{enumerate}
\end{definition}

\subsection{Proofs of Equivalences of P}\label{subsec:EquivalencesofP}
\begin{lemma}\label{le:MeasureToSecondDeriv}
Let $f:\R\to\bar{\R}^{d_{j}}$ be any function and $g(r):=\frac{1}{\sqrt{1+r^2}}$ as in the main paper. Then, the following equivalence holds
\begin{enumerate}
    \item\label{itm:measureExists} $\PmeasurevecOne_j(f)<\infty$
    \item\label{itm:BV2andBoundaryConditions}  $f\in \BVTwoZ$ %
\end{enumerate}
with
\[\PmeasurevecOne_j(f)
:=\inf\left\{|\mu|_{\frac{1}{g}} :
\mu\in\mathfrak{M}^{d_{j}}(\R),
\forall r\in\R: f(r)=\int_{\R} \sigma(r-\xi)d\mu(\xi)\right\}.\]
In this case (if \ref{itm:measureExists} or \ref{itm:BV2andBoundaryConditions} holds):
\[\PmeasurevecOne_j(f)=
|D^2f|_{\frac{1}{g}}\]
\end{lemma}
\begin{proof}
First, assume $\exists\mu\in\mathfrak{M}^{d_{j}}(\R):\forall r\in\R: f(r)=\int_{\R} \sigma(r-\xi)d\mu(\xi)$. Then:
\begin{align}\label{eq:uniqueMu}
    \forall \mu \in \mathfrak{M}^{d_{j}}(\R):
    D^2\int_{\R} \sigma(r-\xi)d\mu(\xi) &=\mu\\
    \implies D^2 f &=\mu \text{, if } \forall r\in\R: f(r)=\int_{\R} \sigma(r-\xi)d\mu(\xi)\notag
\end{align}
This implies that $|\mu|_{\frac{1}{g}}=|D^2 f|_{\frac{1}{g}}<\infty$,  for all $\mu \in \mathfrak{M}^{d_{j}}(\R)$, that fulfill the integral equality $f(r)=\int_{\R} \sigma(r-\xi)d\mu(\xi)$. 
Moreover, by Monotone Convergence Theorem, we have
\begin{align*}
    \lim_{r\to -\infty} f(r)=\lim_{r\to -\infty} \int_{\R} \sigma(r-\xi)d\mu(\xi)&=0\\
    \lim_{r\to -\infty} \frac{\partial}{\partial r}f(r)=\lim_{r\to -\infty} \int_{\R} \ind_{r-\xi>0}d\mu(\xi)&=0.
\end{align*}
Now, assume on the other hand that we have \meqref{itm:BV2andBoundaryConditions}. We set $u:=\frac{\partial}{\partial r}f$ as the weak derivative of $f$ and $\mu:=Du=D^2f$. Then, using the boundary conditions of \meqref{itm:BV2andBoundaryConditions}, we get  %

\[\int_\R \sigma(r-\xi) d\mu(\xi)=-\int_\R \ind_{[0,\infty)}(r-\xi) u(\xi)d\xi=\int_\R \delta_0(r-\xi) f(\xi)d\xi=f(r).\]
Thus, $\mu$ fulfills all conditions of $\PmeasurevecOne_j$, and since \cref{eq:uniqueMu} shows the uniqueness of such $\mu$, we get $\PmeasurevecOne_j(f)=|D^2 f|_{\frac{1}{g}}<\infty$.

\end{proof}

\begin{lemma}\label{le:PsecondderivequivPmeasurevec}
\[P_j=\Pmeasurevec_j\]
\end{lemma}
\begin{proof}

\begin{enumerate}
    \item $P_j\leq \Pmeasurevec_j$\\
    Let $\mu\in\mathfrak{M}^{d_{j}}(\Sd[{d_{j-1}}-1]\times\R)$, $c\in\R^{d_j}$ such that $\forall x\in\R^{d_{j-1}}: f(x)=c+\int_{\Sd[{d_{j-1}}-1]\times\R} \sigma(\langle v, x\rangle -r)d\mu(v,r)$ and $|\mu|_{\frac{1}{g(r)}}<\infty$.
    Then we define $\nu\in\mathfrak{M}^{\BVTwoZ}(\Sd[{d_{j-1}}-1])$ as
    \[\nu(B)(x):=\int_{B\times\R} \sigma(%
    x
    -r)d\mu(v,r), \quad \forall B\in\mathfrak{B}(\Sd[{d_{j-1}}-1]), \ \forall x \in \R.\]
Let us decompose $|\mu|(dv,dr) = k(v,dr)a(dv)$, where $a$ is the marginal measure of $|\mu|$ on $S^{d_{j-1}-1}$. With this in mind we can calculate the total variation measure of $|\nu|$, namely
$$
|\nu|(B):=\int_B \int_{\mathbb{R}} \frac{1}{g(r)} k(v,dr) a(dv) , \quad \forall B\in\mathfrak{B}(\Sd[{d_{j-1}}-1]) \, .
$$
Denote additionally $\tilde k(v,dr):=\frac{d\mu}{d|\mu|}(v,r)k(v,dr)$ the $\R^{d_j}$-valued kernel representing $\mu$ via $a$, i.e.
\[\mu(dv,dr)=\frac{d\mu}{d|\mu|}(v,r)|\mu|(dv,dr)=\frac{d\mu}{d|\mu|}(v,r)k(v,dr)a(dv)=\tilde k(v,dr)a(dv).\] This then allows to calculate the point-wise Radon Nikodym derivative of $\nu$ with respect to $|\nu| $ and introduce the notation $\varphi=\frac{d\nu}{d|\nu|}:\Sd[{d_{j-1}}-1]\to\BVTwoZ,v\mapsto\varphi_v $, such that
\[
\varphi_v(x) = \frac{\int_\R \sigma(x-r)\tilde{k}(v,dr)}{\int_\R \frac{1}{g(r)}k(v,dr)},\quad\forall x \in \R.\]

Then we get that
\begin{itemize}
        \item for every $ x\in\R^{d_{j-1}}$:
    \begin{align*}
        f(x)&=c+\int_{{\Sd[{d_{j-1}}-1]}\times\R} \sigma(\langle v, x\rangle
    -r)\mu(dv,dr)\\
    &=c+\int_{\Sd[{d_{j-1}}-1]}\int_\R\sigma(\langle v, x\rangle
    -r)\,\tilde{k}(v,dr)\,a(dv)\\
    &=c+\int_{\Sd[{d_{j-1}}-1]}\left(\frac{\int_\R \sigma(\langle v, x\rangle-r)\tilde{k}(v,dr)}{\int_\R \frac{1}{g(r)}k(v,dr)}\right)\int_\R \frac{1}{g(r)}k(v,dr)\,a(dv)\\
    &=c+ \int_{\Sd[{d_{j-1}}-1]}\varphi_v(\langle v,x\rangle)\,|\nu|(dv).
    \end{align*}

    \item Finally, we obtain $P_j\leq \Pmeasurevec_j$ from
    \begin{align*}
    \int_{\Sd[{d_{j-1}}-1]}|D^2 \varphi_v|_{\frac{1}{g(r)}} |\nu|(dv)%
    &=\int_{\Sd[{d_{j-1}}-1]}\frac{\left|\tilde{k}(v,\cdot)\right|_{\frac{1}{g(r)}}}{\int_\R \frac{1}{g(r)}k(v,dr)} |\nu|(dv)\\
    &= \int_{\Sd[{d_{j-1}}-1]}\frac{\int_\R \frac{1}{g(r)}k(v,dr)}{\int_\R \frac{1}{g(r)}k(v,dr)} |\nu|(dv)\\
    &= \int_{\Sd[{d_{j-1}}-1]}\int_{\R}\frac{1}{g(r)} k(v,dr)a(dv)\\
    &=\int_{\Sd[{d_{j-1}}-1]\times\R}\frac{1}{g(r)}\,|\mu|(dv,dr)= |\mu|_{\frac{1}{g(r)}}.
\end{align*}
\end{itemize}
    \item $P_j\geq \Pmeasurevec_j$\\
    Let $\nu\in\mathfrak{M}^{\BVTwoZ}(\Sd[{d_{j-1}}-1])$, $c\in\R^{d_j}$. 
    Then, we define $\mu\in\mathfrak{M}^{d_{j}}(\Sd[{d_{j-1}}-1]\times\R)$ as
    \[
    \mu(B,dr):= D^2 (\nu(B))(dr) \, .
    \]
    Analogously to before, we have $\forall x\in\R^{d_{j-1}}: f(x):=c+
\int_{\Sd[{d_{j-1}}-1]} \varphi_v(\langle v,x\rangle) |\nu|(dv)=c+\int_{\Sd[{d_{j-1}}-1]\times\R} \sigma(\langle v, x\rangle -r)\mu(dv,dr)$
    and
\begin{align*}
    |\mu|_{\frac{1}{g}}=\int_{\Sd[{d_{j-1}}-1]}|D^2 \varphi |_{\frac{1}{g(r)}} d|\nu|(s).
\end{align*}
\end{enumerate}

\end{proof}

\begin{lemma}\label{le:PmeasurevecEquivPwellemeasureG}
\[\Pmeasurevec_j=\PwellemeasureG_j\]

\end{lemma}
\begin{proof}
Similarly to \cite{chizat2018global}, we define an operator $T$ mapping a bounded Radon measure on $\Sd[{d_{j-1}}-1]\times\mathbb{R}\times\Sd[{d_{j}}-1]$ onto an $\R^{d_{j}}$-valued, bounded Radon measure on $\Sd[{d_{j-1}}-1]\times\mathbb{R}$ as
\begin{align*}
    T: \mathfrak{M}(\Sd[{d_{j-1}}-1]\times\mathbb{R}\times\Sd[{d_{j}}-1])&\to\mathfrak{M}^{{d_{j}}}(\Sd[{d_{j-1}}-1]\times\mathbb{R})\\
    \nu&\mapsto T(\nu), T(\nu)(B):=\int_{B\times\Sd[{d_{j}}-1]}{w}\,d\nu((v,r),w), \quad \forall B\in\mathfrak{B}(\Sd[{d_{j-1}}-1]\times\R).
\end{align*}
\begin{itemize}
    \item $\PwellemeasureG_j\ge \Pmeasurevec_j$\newline 
    Then, for every $\nu\in\mathfrak{M}(\Sd[{d_{j-1}}-1]\times\mathbb{R}\times\Sd[{d_{j}}-1])$ s.t.
\begin{align*}
    f(x)=c+\int_{\Sd[{d_{j-1}}-1]\times\R\times\Sd[{d_{j}}-1]} w\sigma(\langle v, x\rangle-r)d\nu(v,r,w),
\end{align*}
the above defined $T(\nu)$ fulfills
\begin{align*}
    f(x)&=c+\int_{\Sd[{d_{j-1}}-1]\times\R\times\Sd[{d_{j}}-1]} w\sigma(\langle v, x\rangle-r)d\nu(v,r,w)\\
    &=c+\int_{\Sd[{d_{j-1}}-1]\times\R} \sigma(\langle v, x\rangle-r)dT(\nu)(v,r),
\end{align*}
and
\begin{align*}
    |T(\nu)|_{\frac{1}{g(r)}}&=\sup_{(E_i)_{1,\dots,n}\text{ is partition of } \Sd[{d_{j-1}}-1]\times\R,n\in\N}\sum_i\|\int_{E_i}\frac{1}{g(r)} dT(\nu)(v,r)\|_2\\
    &=\sup_{(E_i)_{1,\dots,n}\text{ is partition of } \Sd[{d_{j-1}}-1]\times\R,n\in\N}\sum_i\|\int_{E_i\times\Sd[{d_{j}}-1]}w\frac{1}{g(r)} d\nu(v,r,w)\|_2\\
     &\leq\int_{\Sd[{d_{j-1}}-1]\times\R\times\Sd[{d_{j}}-1]}\|w\|_2\frac{1}{g(r)} d\nu(v,r,w)\\
     &=\int_{\Sd[{d_{j-1}}-1]\times\R\times\Sd[{d_{j}}-1]}\frac{1}{g(r)} d\nu(v,r,w)
\end{align*}
    \item $\Pmeasurevec_j\ge\PwellemeasureG_j$\newline 
Conversely, for every $\mu\in\mathfrak{M}^{{d_{j}}}(\Sd[{d_{j-1}}-1]\times\mathbb{R})$ of bounded variation, there exists a $h:\Sd[{d_{j-1}}-1]\times\mathbb{R}\to\R^{d_{j}}$ such that

\begin{align*}
    \mu(B)=\int_{B}h d|\mu|=\int_{B\times\Sd[{d_{j}}-1]}w d\underbrace{|\mu|_{\#}(id\times h)}_{=:\nu}(v,r,w),
\end{align*}
where the last inequality takes the pushforward measure\footnote{Within this argument $|\mu|$ denotes the total variation measure, whereas most of the time we use $|\mu|$ as a short notation of $|\mu|(\Om)$}
of $|\mu|$ under $id\times h:\Sd[{d_{j-1}}-1]\times\mathbb{R}\to\Sd[{d_{j-1}}-1]\times\mathbb{R}\times\Sd[{d_{j}}-1]$.

One can pull the norm inside the integral for sufficiently fine measurable partition $(E_i)$ since $h$ is measurable:
\begin{align*}
 |\mu|_{\frac{1}{g(r)}}&=\sup_{(E_i)_{1,\dots,n}\text{ is partition of } \Sd[{d_{j-1}}-1]\times\R,n\in\N}\sum_i\|\int_{E_i}\frac{1}{g(r)} d\mu(v,r)\|_2\\
&=\sup_{(E_i)_{1,\dots,n}\text{ is partition of } \Sd[{d_{j-1}}-1]\times\R,n\in\N}\sum_i\|\int_{E_i}\frac{1}{g(r)} h d|\mu|(v,r)\|_2\\
&=\sup_{(E_i)_{1,\dots,n}\text{ is partition of } \Sd[{d_{j-1}}-1]\times\R,n\in\N}\sum_i\int_{E_i}\frac{1}{g(r)} \|h\|_2 d|\mu|(v,r)\\
&=\sup_{(E_i)_{1,\dots,n}\text{ is partition of } \Sd[{d_{j-1}}-1]\times\R,n\in\N}\sum_i\int_{E_i\times\Sd[{d_{j}}-1]}\frac{1}{g(r)} \|w\|_2 d\nu(v,r,w)\\
&=\int_{\Sd[{d_{j-1}}-1]\times\R\times\Sd[{d_{j}}-1]}\frac{1}{g(r)} \|w\|_2 d\nu(v,r,w)\\
&=\int_{\Sd[{d_{j-1}}-1]\times\R\times\Sd[{d_{j}}-1]}\frac{1}{g(r)} d\nu(v,r,w)\\
\end{align*}
Moreover,
\begin{align*}
    f(x)&=c+\int_{\Sd[{d_{j-1}}-1]\times\R\times\Sd[{d_{j}}-1]} w\sigma(\langle v, x\rangle-r)d\nu(v,r,w)\\
    &=c+\int_{\Sd[{d_{j-1}}-1]\times\R} \sigma(\langle v, x\rangle-r)d\mu(v,r).
\end{align*}
\end{itemize}

\end{proof}

\begin{lemma}\label{le:PwellemeasureGEquivPwellemeasure}
\[\PwellemeasureG_j=\Pwellemeasure_j\]

\end{lemma}
\begin{proof}
Analogous arguments as in the proofs of \Cref{le:PmeasurevecEquivPwellemeasureG,le:l2equivl1g} prove the claim as follows.

Let $A:=\Sd[d_{j-1}]\times\Sd[d_{j}-1]$ and $B:=\Sd[d_{j-1}-1]\times\R\times\Sd[d_{j}-1]$. Also, let $A_{0}:=\{((v,r),w)\in A \,|\, v=0\}$ and $A_{\neq0}:=A\setminus A_0=\{((v,r),w)\in A \,|\, v\neq0\}$, and
\begin{align*}
    \Psi:\,A_{\neq0}&\to B,\\
    \left((v,r),w\right)&\mapsto\left(\frac{v}{g\left(-\frac{r}{\twonorm[v]}\right)},\frac{r}{g\left(-\frac{r}{\twonorm[v]}\right)},w\right),
\end{align*}
a measurable and bijective function. Additionally, let $f\in\mathcal{F}_{A}$ and $(\mu,c)\in\mathfrak{C}_{A}^{f}$ be a minimizer of $\Pwellemeasure_j(f)$.\\
Next, we define the restricted measures $\mu_{\neq0}:=\mu|_{A_{\neq0}}$, where $\mu|_{A_{\neq0}}(E):=\mu(E\cap A_{\neq0})$, and $\mu_{0}:=\mu|_{A_{0}}$.\footnote{Note that $\mu(E)=\mu_{\neq0}(E\cap A_{\neq0})+\mu_0(E\cap A_{0})=\mu|_{A_{\neq0}}(E\cap A_{\neq0})+\mu|_{A_0}(E\cap A_{0})$.}
Then, we additionally define the measure
\begin{align*}
    \tilde{\mu}:\,B&\to[0,\infty),\\
    \tilde{\mu}(E)&:=\int_E g\left(-\frac{r}{\twonorm[v]}\right)\,d\Psi_{\#}\mu_{\neq0}(v,r,w),\ E\in\mathfrak{B}(B),
\end{align*}
and the constant vector
\begin{align*}
    \tilde{c}:=c+\int_{A_{0}^-} w\, d\mu_0((v,r),w)
\end{align*}
where $A_0^-:=\{((0,-1),w)\,|\,w\in\Sd[d_{j}-1]\}\subseteq A_0$.
For $x\in\R^{d_{j-1}}$ we then get 
\begin{align*}
f(x)=&c+\int_A\phi_x(v,r,w)\,d\mu((v,r),w)\\
=&c+\int_{A_0}\phi_x(v,r,w)\,d\mu_0((v,r),w)+\int_{A_{\neq0}}\phi_x(v,r,w)\,d\mu_{\neq0}((v,r),w)\\
\stackrel{(a)}{=}&c+\int_{A_0^-}w\,d\mu_0((v,r),w)+\int_{A_{\neq0}}\phi_x(v,r,w)\,d\mu_{\neq0}((v,r),w)\\
=&c+\int_{A_0^-}w\,d\mu_0((v,r),w)+\int_{A_{\neq0}}(\phi_x\circ\Psi^{-1}\circ\Psi)(v,r,w)\,d\mu_{\neq0}((v,r),w)\\
 \stackrel{(b)}{=}&\tilde{c}+\int_B(\phi_x\circ\Psi^{-1})(v,r,w)  \,d(\Psi_{\#}\mu_{\neq0})(v,r,w)\\
  \stackrel{(c)}{=}&\tilde{c}+\int_B(\phi_x\circ\Psi^{-1})(v,r,w)  \frac{1}{g\left(-\frac{r}{\twonorm[v]}\right)}\,d\tilde{\mu}(v,r,w)\\
  \stackrel{(d)}{=}&\tilde{c}+\int_B\phi_x(v,r,w)  \,d\tilde{\mu}(v,r,w)
\end{align*}
where we used
\begin{enumerate}[(a)]
    \item that in $A_0$ it is $r=\pm1$ but in case of $r=1$, we get $\phi_x(0,1,w)=0$, i.e., only the case $r=-1$ is of relevance where we always have $\phi_x(0,-1,w)=w$,
    \item a change-of-variable via the pushforward measure $\Psi_{\#}\mu_{\neq0}$,
    \item the definition of $\tilde{\mu}$ and
    \item the positive homogeneity of $\sigma$.
\end{enumerate}
Overall, we have shown that $(\tilde{\mu},\tilde{c})\in\mathfrak{C}_{B}^{f}$.\\
\newline
Now, first note that
\begin{align}\label{eq:PwellemeasureGregularizer}
\begin{split}
    2\int_B\frac{1}{g(r)}\,d\tilde{\mu}(v,r,w)&=2\int_B\frac{g\left(-\frac{r}{\twonorm[v]}\right)}{g(r)}\,d(\Psi_{\#}\mu_{\neq0})(v,r,w)\\&=2\int_{A_{\neq0}}\frac{g\left(-\frac{r}{\twonorm[v]}\right)}{g\left(-\frac{r}{g\left(-\frac{r}{\twonorm[v]}\right)}\right)}\,d\mu_{\neq0}((v,r),w)\\
    &=2\int_{A_{\neq0}}\,d\mu_{\neq0}((v,r),w)=2\vert\mu_{\neq0}\vert
\end{split}
\end{align}
where the last step stems from $\mu_{\neq0}$ being non-negative.\\
Next, we make a case distinction:
\begin{enumerate}
    \item let $\twonorm[\tilde{c}]\leq1$:
    \begin{align*}
        2\rho(\twonorm[\tilde{c}])=\twonorm[\tilde{c}]^2\leq\twonorm[c]^2+\int_{A_0^-}\twonorm[w]^2\,d\mu_0((v,r),w)\leq\twonorm[c]^2+\vert\mu_0\vert\leq\twonorm[c]^2+2\vert\mu_0\vert,
    \end{align*}
    \item let $\twonorm[\tilde{c}]>1$:
    \begin{align*}
        2\rho(\twonorm[\tilde{c}])=2\twonorm[\tilde{c}]-1\leq2\twonorm[c]+2\vert\mu_0\vert-1\leq\twonorm[c]^2+2\vert\mu_0\vert.
    \end{align*}
\end{enumerate}
Overall, we obtain due to (\ref{eq:PwellemeasureGregularizer}) that
\begin{align*}
    2\int_B\frac{1}{g(r)}\,d\tilde{\mu}(v,r,w)+2\rho(\twonorm[\tilde{c}])\leq2\vert\mu_{\neq0}\vert+2\vert\mu_0\vert+\twonorm[c]^2=2\vert\mu\vert+\twonorm[c]^2,
\end{align*}
i.e., we have $\PwellemeasureG_j(f)\leq\Pwellemeasure_j(f)$.\\
\newline
The case $\PwellemeasureG_j(f)\geq\Pwellemeasure_j(f)$ follows in an analogous manner.

\end{proof}

\begin{lemma}\label{le:PsecondderivequivPwellemeasure}
It holds that
$P_j=\Pwellemeasure_j$, for all $j=1\ldots,\nStacks$, and therefore
\[\Pfunc=\Pwellemeasure.\]
\end{lemma}
\begin{proof}
By \Cref{le:PsecondderivequivPmeasurevec,le:PmeasurevecEquivPwellemeasureG,le:PwellemeasureGEquivPwellemeasure}, we know that 
\[P_j=\Pwellemeasure_j,\quad\forall j=1\ldots,\nStacks.\]
Thus, for every $f\in\cF$
\begin{align}
\Pfunc(f) &= \inf_{\substack{(h_1,\dots, h_{\nStacks})\text{, s.t. }\\ f=\link^{-1}\circ h_{\nStacks}\circ\dots\circ \sigb\circ h_1}} \left(P_1(h_1) + P_2(h_2) + \dots + P_\nStacks(h_\nStacks) \right)\\
&=\inf_{\substack{(h_1,\dots, h_{\nStacks})\text{, s.t. }\\ f=\link^{-1}\circ h_{\nStacks}\circ\dots\circ \sigb\circ h_1}} \left(\Pwellemeasure_1(h_1) + \Pwellemeasure_2(h_2) + \dots + \Pwellemeasure_\nStacks(h_\nStacks) \right)\\
&=\Pwellemeasure(f).
\end{align}
\end{proof}

\subsection{Proof of Existence of Minimizers}\label{subsec:ProofExistenceMinimizer}
In \Cref{le:infinPmeasurewelleismin} we prove that the minimum in eq. \eqref{eq:Pwellemeasure} is actually attained and in \Cref{cor:ExistenceMinimizerF} we prove that the minimum in \eqref{eq:optimalFunction} is attained for the regularization functional $\Pwellemeasure$.
Together with  the equivalences of \Cref{sec:EquivalentPFuncs}, this yields the existence of minimizers in \Cref{thm:PFunc}.

In order to prove \Cref{le:infinPmeasurewelleismin} and \Cref{cor:ExistenceMinimizerF}, we first need some basic definitions and lemmas.

In the following, we will use the \textit{topology of point-wise convergence} or \textit{tptc} for short which we will denote by $\tau_{\text{tptc}}$. For a more detailed introduction of this topology, we refer to \cite[§ 46]{munkres2014topology}. For a topological vector space $\left(V,\tau\right)$, we will also sometimes write $\left(A,\tau\right)$ for a subset $A\subseteq V$ if we want to emphasise the underlying topology.

\begin{lemma}\label{le:ConcatTptcContinuous}
 Let $d_1,\,d_2,\,d_3>1$ and $A:=C(\R^{d_1},\R^{d_2})$, $B:=\operatorname{Lip}_{L_{\text{Lip}}}(\R^{d_2},\R^{d_3})$ the set of Lipschitz continuous functions with a Lipschitz constant smaller than $L_\text{Lip}$ where $\infty>L_\text{Lip}>0$, and $Y^X$, the topological space of all functions from $X:=\R^{d_{1}}$ into $Y:=\R^{d_{3}}$, %
 each be equipped with the tptc. Then, for $h_1\in A$ and $h_2\in B$, each being tptc-continuous, with $h_2$ additionally being Lipschitz continuous, the concatenation
 \begin{align}\label{eq:concatenation}
     \circ:\left(\left(B,\tau_{\text{tptc}}\right),\left(A,\tau_{\text{tptc}}\right)\right)\to \left(Y^X,\tau_{\text{tptc}}\right),\ (h_2\circ h_1)(x):=h_2(h_1(x)),
 \end{align}
 for $x\in \R^{d_1}$, is also continuous.
\end{lemma}
\begin{proof}
First, let $d_2=d_3=1$.\\
Now, let $h:=(h_1,h_2)\in A\times B$ with $h_1$ and $h_2$ being tptc-continuous and $h_2$ additionally being Lipschitz continuous with Lipschitz constant $L_{h_2}<L_{\text{Lip}}$. Also, let $x\in\R^{d_1}$ and $\epsilon>0$. Then, the open set
\begin{align*}%
    S(x,\epsilon):=\Big\{ h \in Y^X \,\Big|\, h(x)\in\left(h_2(h_1(x))-\epsilon,h_2(h_1(x))+\epsilon\right)\Big\}
\end{align*}

forms a subbasis element about $h_2\circ h_1$ at $x$.

Next, we define 
\begin{align}\label{eq:DeltaSet}
\begin{split}
    U(x,\epsilon):=\bigg\{ (\tilde{h}_1,\tilde{h}_2)\in\,& A\times B\, |\, %
    \tilde{h}_1(x)\in\left(h_1(x)-\frac{\epsilon}{2L_\text{Lip}},\,h_1(x)+\frac{\epsilon}{2L_\text{Lip}}\right),\\ & \tilde{h}_2(h_1(x))\in \left(h_2(h_1(x))-\frac{\epsilon}{2},\,h_2(h_1(x))+\frac{\epsilon}{2}\right)\bigg\},
\end{split}
\end{align}
which is an open set in $\left(B,\tau_{\text{tptc}}\right)\times\left(A,\tau_{\text{tptc}}\right)$.\\ \newline
We first note that the set $U(x,\epsilon)$ is not empty since $(h_1,h_2)\in U(x,\epsilon)$.\\
For $(\tilde{h}_2,\tilde{h}_1)\in U(x,\epsilon)$, we get
\begin{align*}
    \tilde{h}_2(\tilde{h}_1(x))\leq\tilde{h}_2(h_1(x))\, +\, L_{\tilde{h}_2}\frac{\epsilon}{2L_\text{Lip}}<h_2(h_1(x))\, +\, \epsilon
\end{align*}
and, analogically,
\begin{align*}
    \tilde{h}_2(\tilde{h}_1(x))>h_2(h_1(x))\, -\, \epsilon,
\end{align*}
i.e., we have $\tilde{h}_2(\tilde{h}_1(x))\in S(x,\epsilon)$.\\
Thus, for every $(h_1,h_2)\in A\times B$ and subbasis element $S$ of $h_2\circ h_1$, we have constructed an open neighborhood $U$ of $(h_1,h_2)$ such that each of its elements are mapped into $S$ under the concatenation, i.e., the mapping in \cref{eq:concatenation} is continuous.
\\ \newline
In case of $d_2,d_3>1$, we just demand each $\tilde{h}_1(x)$ and $\tilde{h}_2(h_1(x))$ in \cref{eq:DeltaSet} to be an element of the open ball around $h_1(x)$ with radius $\frac{\epsilon}{2L_{\text{Lip}}}$ and $h_2(h_1(x))$ with radius $\frac{\epsilon}{2}$, respectively.
\end{proof}

\begin{lemma}\label{le:MeasureToFunctionContinuous}
Let $j\in\{1,\dots,\nStacks\}$, $\mathfrak{M}_j:=\mathfrak{M}(K_j)\times\R^{d_j}$ with $K_j:=\Sd[d_{j-1}]\times\Sd[d_{j}-1]$, and $\phi_x\in C(K_j)$ with $\phi_x((v,r),w):=w\sigma(\langle v, x\rangle-r)$ for $x\in\R^{d_{j-1}}$.
Then, the surjective function
\begin{align*}
    \Psi_j:\left(\mathfrak{M}_j,\tau_{\text{w}}\right)\to\left(\mathcal{F}_{K_j},\tau_{\text{tptc}}\right),\ \Psi_j(\mu,c)(x):=c+\int_{K_j}\phi_{x}\,d\mu,
\end{align*}
where $\tau_{\text{w}}$ denotes the weak topology, is continuous for every $x\in\R^{d_{j-1}}$.
\end{lemma}
\begin{proof}
Let $V\subseteq\mathcal{F}_{K_j}$ be non-empty and open, i.e., for every $f\in V$ there exist finitely many $x_1,\cdots,x_n\in\R^{d_{j-1}}$ and $\epsilon>0$ such that
\begin{align*}
  V(f,n,\epsilon):= \Big\{&g\in\mathcal{F}_{K_j}\,\Big|\, \forall\,\bar{x}\in\{x_1,\cdots,x_n\}:\ \twonorm[f(\bar{x})-g(\bar{x})] <\epsilon\Big\}\\
  =\Big\{&g\in\mathcal{F}_{K_j}\,\Big|\, \forall\, (\mu,c)\in\mathfrak{C}_{K_j}^{f},\,(\nu,\tilde{c})\in\mathfrak{C}_{K_j}^{g},\,\bar{x}\in\{x_1,\cdots,x_n\}:\\ &\twonorm[ c-\tilde{c}+\int_{K_j}\phi_{\bar{x}}\,d(\mu-\nu)]<\epsilon\Big\}
  \subseteq V\,.
\end{align*}
Now, let $(\mu,c)\in \Psi^{-1}_j(V)$, and $\tilde{x}_1,\dots,\tilde{x}_m\in\R^{d_{j-1}}$ and $r>0$ such that $V(\Psi_j(\mu,c),m,r)\subseteq V$. Then, we get %
that
\begin{align*}
    U((\mu,c),m,r)&:=\Big\{(\nu,\tilde{c})\in\mathfrak{M}_j\,\Big|\, \forall\,\bar{x}\in\{\tilde{x}_1,\cdots,\tilde{x}_m\}:\ \twonorm[ c-\tilde{c}+\int_{K_j}\phi_{\bar{x}}\,d(\mu-\nu)]<r\Big\}\\ &\subseteq\Psi_j^{-1}(V(\Psi_j(\mu,c),m,r))\subseteq \Psi_j^{-1}(V),
\end{align*}
 i.e., we have shown that the set $\Psi_j^{-1}(V)$ is open and, therefore, also the continuity of $\Psi_j$.
\end{proof}

\begin{theorem}\label{le:infinPmeasurewelleismin}
The infimum of $\Pwellemeasure(f)$ for every $f\in\cF$ is attained.
\end{theorem}
\begin{proof}
We first focus on a single stack; here, we follow a classical \textit{direct method} approach: Let 
\begin{enumerate}
    \item $K:=\Sd[\din]\times\Sd[\dout-1]$ and $\mathfrak{M}_0:=\mathfrak{M}(K)\times\R^{\dout}$ be equipped with the norm $\vert(\cdot,\cdot)\vert:=\vert\cdot\vert+\twonorm[\cdot]$, where $\mathfrak{M}(K)$ is a closed, convex subset of the Banach space $(\mathfrak{M}_{s}(K),\vert\cdot\vert)$ of all finite signed Radon measures $\mu:\,\mathfrak{B}(K)\to\R$,
    \item the set of functions $\mathcal{F}_{K}$ and the set $ \mathfrak{C}_0:=\mathfrak{C}_{K}^{f}$ as in \Cref{def:FunctionMasureSets}, and
    \item a minimizing sequence $(\mu_k,c_k)$ in $\mathfrak{C}_0$ with $F_0(\mu_k,c_k)\rightarrow\inf\limits_{(\mu,c)\in\mathfrak{C}_0}F_0(\mu,c)$ where
    \begin{align*}
        F_0:\ \mathfrak{M}_0\to\R_{\geq 0},\ (\mu,c)\mapsto 2|\mu|+\twonorm[c]^2.
    \end{align*}
\end{enumerate}
The set $\mathfrak{C}_0$ is obviously convex. Additionally, it is even closed:\\
let $(\tau_k,\tilde{c}_k)$ be a sequence in $\mathfrak{C}_0$ converging to $(\tau,\hat{c})\in\mathfrak{M}_0$. For $\phi_x\in C(K,\R^{\dout})$ with $\phi_x((v,r),w):=w\sigma(\langle v, x\rangle-r)$, we then get
\begin{align*}
    \twonorm[ f(x) -\hat{c}-\int_{K} \phi_x\,d\tau]&=\twonorm[\tilde{c}_k-\hat{c}+\int_{K} \phi_x\,d(\tau_k-\tau)]\\&\leq\twonorm[\tilde{c}_k-\hat{c}] \Vert\phi_x\Vert_{\infty}\vert\tau_k-\tau\vert\ \xrightarrow{k\to\infty}\ 0
\end{align*}
for all $x\in\R^{\din}$, i.e., we have $(\tau,\hat{c})\in \mathfrak{C}_0$ and, therefore, the set $\mathfrak{C}_0$ is closed.\\
It should also be noted that the functional $F_0$ is coercive and, therefore, the sequence $(\mu_k,c_k)$ is bounded.\\
Next, it follows from \textit{Riesz's repsresentation theorem} that there exists an isometric isomorphism  from $\mathfrak{M}(K)$ to the closed subset of all positive linear functionals of the dual space ${C\left(K\right)}^*$ (see \cite[2.26]{elstrodt2018massintegrationstheorie}).
Given that ${C\left(K\right)}^*$ is a separable Banach space, every bounded sequence in ${C\left(K\right)}^*$ has a weak*-convergent subsequence due to the \textit{Banach–Alaoglu theorem}. Therefore, we can say that due to the isometric isomorphism, $(\mu_k,c_k)$ must have a weakly\footnote{The weak topology on $\mathfrak{M}(K)$ is induced by ${C\left(K\right)}^*$; a sequence $(\mu_k)$ converges against a $\mu$ w.r.t. this also called \textit{vague topology} if $\int f\,d\mu_k\to\int f\,\mu$ for all $f\in C\left(K\right)$.\\We also note that the product topology consisting of each of the weak topologies equals the weak topology of the product space.} convergent subsequence that weakly converges against a $(\mu_0,c_0)\in \mathfrak{M}_0$. Since $\mathfrak{C}_0$ is closed and convex, it is even $(\mu_0,c_0)\in \mathfrak{C}_0$ by the \textit{Hahn-Banach separation theorem}, i.e., the set $\mathfrak{C}_0$ is weakly closed.\\ Additionally, because $F_0$ is convex and (lower semi-)continuous, we can conclude via \textit{Mazur's lemma} (see \cite[3.13 Theorem]{rudin1991functionalanalysis}) that $F_0(\mu_0,c_0)=\inf\limits_{(\mu,c)\in \mathfrak{C}_0}F_0(\mu,c)$, i.e., we have $\Pwellemeasure(f)=F_0(\mu,c)$ for at least one $(\mu,c)\in \mathfrak{C}_0$.\\
\newline
In case of a deep stacked NN where we use the notation $h:=(h_1,\dots, h_{\nStacks})$, the minimization problem is
\begin{align}\label{eq:deepPwellemeasure}
\Pwellemeasure(f) := \inf_{h\in \mathcal{D}^f} \sum_{j=1}^{\nStacks}\Pwellemeasure_{j}(h_j)
\end{align}
with
\begin{align*}
    \mathcal{D}^{f}:=\Big\{h\,\Big|\, f=\link^{-1}\circ h_{\nStacks}\circ\dots\circ \sigb\circ h_1,\, h_j:\R^{d_{j-1}}\rightarrow \R^{d_{j}},&\\ \mathfrak{C}_{K_j}^{h_j}\neq\emptyset\ \forall\, j=1,\dots,\nStacks&\Big\}
\end{align*}
where $f\in\cF$ and $K_j:=\Sd[d_{j-1}]\times\Sd[d_{j}-1]$.\\
We will first focus on the following problem in measure space:
\begin{align}\label{eq:measureWelleproblem}
    \inf_{(\mu,c)\in \hat{\mathfrak{C}}} F(\mu,c):= \inf_{(\mu,c)\in \hat{\mathfrak{C}}} \sum_{j=1}^{\nStacks}2\vert\mu_j\vert+\twonorm[c_j]^2
\end{align}
with an appropriate closed and convex set $\hat{\mathfrak{C}}\subseteq \mathfrak{M}^{\nStacks}:=\prod\limits_{j=1}^{\nStacks}\mathfrak{M}_j$ with $\mathfrak{M}_j:=\mathfrak{M}(K_j)\times\R^{d_j}$, that we will define later on, for which we will show that a minimum is being attained; from this, we will then conclude the same for \cref{eq:deepPwellemeasure}:\\
Let $j\in\fromto{\nStacks}$. We first define the set
\begin{align*}
    \mathfrak{S}_j:=\Big\{(\mu,c)\in\mathfrak{M}_j\,\Big|\,\vert(\mu,c)\vert\leq \nStacks+F(\nu,\tilde{c})\Big\}
    \supseteq\Big\{(\mu,c)\in\mathfrak{M}_j\,\Big|\,F(\mu,c)\leq F(\nu,\tilde{c})\Big\}
\end{align*}
with
\begin{align*}
    (\nu,\tilde{c})=((\nu_1,\tilde{c}_1),\dots,(\nu_{\nStacks},\tilde{c}_{\nStacks}))\in\prod\limits_{j=1}^{\nStacks}\mathfrak{C}_{K_j}^{\tilde{h}_j}
\end{align*}
for an arbitrarily chosen $\tilde{h}\in \mathcal{D}^f$ which does exists because $f\in\cF$.
The set $\mathfrak{S}_j$ is obviously closed and convex and, therefore, weakly closed.
Because it is also bounded, any sequence in $\mathfrak{S}_j$ has a subsequence weakly convergent to a point contained in the same set as we have seen in the first part of the proof, i.e., the set is weakly sequentially compact and by the \textit{Eberlein–Šmulian theorem} also weakly compact. The same then also applies to the Cartesian product $\mathfrak{S}:=\prod\limits_{j=1}^{\nStacks} \mathfrak{S}_j$.\\
\newline
Now, given that we can choose $L_{\sigma}:=1$ as a Lipschitz constant for $\sigma$, for a fixed $(\mu,c)\in\mathfrak{C}_{K_j}^{h_j}$ we then get
\begin{align*}%
    \twonorm[h_j(x)-h_j(y)]\leq\vert\mu\vert\twonorm[x-y]
\end{align*}
for all $x,y\in\R^{d_{j-1}}$, i.e., we have $h_j\in \operatorname{Lip}_{\vert\mu\vert}(\R^{d_{j-1}},\R^{d_{j}})$ for every $j=1,\dots,\nStacks$.\\
\newline
Next, let $L_{\tilde{h}}:=\nStacks+F(\nu,\tilde{c})$.
We define the continuous and surjective function
 \begin{align*}%
 \begin{split}
     &\Psi:\,(\mathfrak{M}^{\nStacks},\tau_{w})\to (\mathcal{F}_{K_1}\times\dots\times\mathcal{F}_{K_{\nStacks}},\tau_{\text{tptc}}),\\
     &\Psi(\mu,c):=\left(\Psi_1(\mu_1,c_1),\cdots,\Psi_{\nStacks}(\mu_{\nStacks},c_{\nStacks})\right),
 \end{split}
 \end{align*}
 with $\Psi_j$ from \cref{le:MeasureToFunctionContinuous}. We then have $\Psi(\mathfrak{S})\subseteq \operatorname{Lip}_{L_{\tilde{h}}}^{\nStacks}$ with
 \begin{align*}
    \operatorname{Lip}_{L_{\tilde{h}}}^{\nStacks}:=\operatorname{Lip}_{L_{\tilde{h}}}(\R^{d_{0}},\R^{d_{1}})\times\cdots\times \operatorname{Lip}_{L_{\tilde{h}}}(\R^{d_{\nStacks-1}},\R^{d_{\nStacks}}).
\end{align*}
\newline
Now, for $Y^X$ with $X:=\R^{d_{0}}$ and $Y:=\R^{d_{\nStacks}}$, the function
\begin{align}\label{eq:MappingStacksToShallowNetworks}
\begin{split}
    T:\, &(\operatorname{Lip}_{L_{\tilde{h}}}^{\nStacks},\tau_{\text{tptc}})\to (Y^X,\tau_{\text{tptc}}),\\ &(h_1,\dots,h_{\nStacks})\mapsto \link^{-1}\circ h_{\nStacks}\circ\dots\circ \sigb\circ h_1,
\end{split}
\end{align}
is continuous according to \cref{le:ConcatTptcContinuous} given that $ \link^{-1}$ and $\tilde{\sigma}$ are both Lipschitz continuous. Then, the set $\Psi^{-1}(T^{-1}(\{f\}))$ is weakly closed, i.e., the intersection
\begin{align*}
    \hat{\mathfrak{C}}:=\Psi^{-1}(T^{-1}(\{f\}))\cap \mathfrak{S}\ \subseteq \mathfrak{C}
\end{align*}
is weakly compact. We want to note that due to the domain of $T$, the pre-image $T^{-1}(\{f\})$ is not empty.\\
\newline
Given that the functional $F$ is convex and (lower semi-)continuous and, therefore, weakly lower semi-continuous, the problem in \cref{eq:measureWelleproblem} attains its infimum in the afroementioned weakly compact set $\hat{\mathfrak{C}}$. Let us denote by $\mathfrak{L}^{f}$ the respective solution set.\\
\newline
Now, in \cref{eq:deepPwellemeasure} we can replace $\mathcal{D}^{f}$ with the (tptc-)compact set $\Psi(\hat{\mathfrak{C}})$ because for any $h\in \mathcal{D}^f\backslash\Psi(\hat{\mathfrak{C}})$ we have that for $(\mu,c)\in\Psi^{-1}(h)$ it is $\vert(\mu_j,c_j)\vert>\nStacks+F(\nu,\tilde{c})$ for at least one $j=1,\dots,\nStacks$, 
i.e., it is $F(\mu,c)\geq \vert(\mu_j,c_j)\vert-\nStacks>F(\nu,\tilde{c})$ and, therefore, it would not be in the solution set of \cref{eq:deepPwellemeasure} anyways.\\
\newline
Finally, we can directly conclude that the problem in \cref{eq:deepPwellemeasure} must also attain its infimum only in every $\hat{h}\in\Psi(\mathfrak{L}^{f})\subseteq\Psi(\hat{\mathfrak{C}})$ because if we assume that there exists a minimizer $\breve{h}\in \mathcal{D}^f$ with
\begin{align*}
    \sum_{j=1}^{\nStacks}\Pwellemeasure_j(\breve{h}_j)<\sum_{j=1}^{\nStacks}\Pwellemeasure_j(\hat{h}_j),
\end{align*}
then, following from the proof of the single stack case, there must exist a $(\breve{\mu},\breve{c})\in\Psi^{-1}(\breve{h})$ with
\begin{align*}
    F(\breve{\mu},\breve{c})=\sum_{j=1}^{\nStacks}\Pwellemeasure_j(\breve{h}_j)<\sum_{j=1}^{\nStacks}\Pwellemeasure_j(\hat{h}_j)=F(\hat{\mu},\hat{c})
\end{align*}
for all $(\hat{\mu},\hat{c})\in\Psi^{-1}(\hat{h})$ which contradicts $\Psi^{-1}(\hat{h})\in \mathfrak{L}^{f}$ being a minimizer.
\end{proof}

\begin{corollary}\label{cor:ExistenceMinimizerF}
  The infimum of $L(f)+\lambda \Pwellemeasure(f)$ with $\lambda>0$ over all $f\in\cF$ is attained.
\end{corollary}
\begin{proof}
We will reuse objects and notations defined within the proof of \cref{le:infinPmeasurewelleismin}: We want to show that for $\lambda>0$, the problem
\begin{align}\label{eq:GeneralProblem}
     \inf_{f\in\cF}\ \, G(f):=\inf_{f\in\cF}\ \,L(f)+\lambda \Pwellemeasure(f)
\end{align}
attains its infimum in $\cF$:\\
Let $j\in\fromto{\nStacks}$, $\tilde{h}_j\in\mathcal{F}_{K_j}$ be arbitrarily chosen, and $g:=T(\tilde{h}_1,\dots,\tilde{h}_{\nStacks})$ with $T$ from \cref{eq:MappingStacksToShallowNetworks} (e.g., $\tilde{h}=0$ and, thus, $g=0$). Analogous to the proof of \cref{le:infinPmeasurewelleismin}, we can conclude that the Cartesian product $\tilde{\mathfrak{S}}:=\prod\limits_{j=1}^{\nStacks} \tilde{\mathfrak{S}}_j$ with
\begin{align*}
    \tilde{\mathfrak{S}}_j:=\Big\{(\mu,c)\in\mathfrak{M}(K_j)\times\R^{d_j}\,\Big|\,\vert(\mu,c)\vert\leq \nStacks+G(g)/\lambda\Big\}
\end{align*}
is weakly compact.\\
\newline
Now, let $L_{\tilde{h}}:=\nStacks+G(g)/\lambda$, $x\in\R^{\din}$, and the continuous point evaluation operator
\begin{align*}
    \delta_x:\,(\operatorname{Lip}_{L_{\tilde{f}}}(\R^{\din},\R^{\dout}),\tau_{\text{tptc}})\to\R^{\dout},\ f\mapsto f(x),
\end{align*}
with $L_{\tilde{f}}:=L_{\link^{-1}}\prod\limits_{j=1}^{\nStacks-1}L_{\tilde{\sigma}}\prod\limits_{j=1}^{\nStacks}L_{\tilde{h}}$.\\
Now, given that $x\mapsto x^2$ is continuous, we can conclude that the loss function
\begin{align*}
    L:\,(\operatorname{Lip}_{L_{\tilde{f}}}(\R^{\din},\R^{\dout}),\tau_{\text{tptc}})\to\R_{\geq0},\ \Ltrb f:=\sum_{j=1}^{N}\left(\delta_{\xtr_j}(f)-\ytr_j\right)^2,
\end{align*}
with training data $\{(\xtr_j,\ytr_j)\}_{j=1}^{N}$ is continuous.\\
\newline
Next, we define the functional
\begin{align*}
    \tilde{G}:(\mathfrak{S},\tau_{\text{w}})\to(\R,\twonorm[\cdot]),\, (\mu,c)\mapsto\tilde{G}(\mu,c):=(L\circ T\circ\Phi + \lambda F)(\mu,c),
\end{align*}
which is lower semi-continuous given the continuity of $L$, $T$ and $\Phi$, and the lower semi-continuity of $F$. Then, the problem
\begin{align*}
    \inf_{(\mu,c)\in\tilde{\mathfrak{S}}}\tilde{G}(\mu,c)
\end{align*}
does attain its infimum. Let us denote by $\mathfrak{L}^{\cF}$ the respective solution set.\\
\newline
Finally, we can directly conclude that \cref{eq:GeneralProblem} must also attain its infimum only in every $\hat{f}\in(T\circ\Psi)(\mathfrak{L}^{\cF})$ because if we assume that there exists a minimizer $\breve{f}\in\cF\backslash(T\circ\Psi)(\mathfrak{L}^{\cF})$ with $G(\breve{f})<G(\hat{f})$, then there must exist a $(\breve{\mu},\breve{c})\in(T\circ\Psi)^{-1}(\breve{f})$ with
\begin{align*}
    \tilde{G}(\breve{\mu},\breve{c})=(G\circ T\circ\Psi)(\breve{\mu},\breve{c})<(G\circ T\circ\Psi)(\hat{\mu},\hat{c})=\tilde{G}(\hat{\mu},\hat{c})
\end{align*}
for all $(\hat{\mu},\hat{c})\in(T\circ\Psi)^{-1}(\hat{h})$ which contradicts $(T\circ\Psi)^{-1}(\hat{f})\in \mathfrak{L}^{\cF}$ being a minimizer.
\end{proof}

\subsection{Proof of \Cref{thm:PFunc}}\label{subsec:ProofMainTheorem}
In this \cref{subsec:ProofMainTheorem}, we will first prove multiple Lemmas, that we then use to conclude the proof of \Cref{thm:PFunc}.

For the proof we use the mathematically precise definition of $P_j$ given in \Cref{def:PjDistribution}.

Similar theorems for shallow NNs with one-dimensional output have already been presented in concurrent and previous work, such as \cite{Chizat2020ImplicitBias,NeyshaburImplicitReg2014arXiv1412.6614N,ongie2019function,savarese2019infinite,williams2019gradient,implReg1}. %
More recently, independently and in parallel to us, \citet{parhi2022kinds} have proven a theorem, which is even more similar to \Cref{thm:PFunc} for a slightly different regularization.

To get some intuition for \Cref{thm:PFunc} it is particularly important to understand the equivalence in function space of solutions to different optimization problems on parameter space (leading to different solutions in parameter space). This equivalence is shown in the following \Cref{le:l2equivl1g}.

\begin{lemma}\label{le:l2equivl1g}
Let $\NN_\theta$ be a stack (i.e., $\nStacks=1$) with input dimension $\din$, output dimension $\dout$, number of hidden neurons $n_1$, and parameters $\theta:=(v,b,w,c)$. It holds that the set of solutions
\begin{equation}\label{eq:l2reg}
   \Set{\NN_{\theta^*}| \theta^*\in\argmin_\theta \left( \Ltrb{\NN_{\theta}} + \lw\twonorm[\theta]^2\right)},
\end{equation}
\begin{equation}\label{eq:l1reg}
   \Set{\NN_{\theta^*}| \theta^*\in\argmin_{\theta,\text{ s.t. } {\allIndi{n_1}{k}:\twonorm[(v_k,b_k)]}=1} \left( \Ltrb{\NN_{\theta}} + \lw\left(\|c\|_2^2+\sum_{k=1}^{n_1}2\twonorm[w_k]\right)\right)}
\end{equation}
{\smaller\begin{equation}\label{eq:l1regg}
   \Set{\NN_{\theta^*}| \theta^*\in\argmin_{\theta,\text{ s.t. } \allIndi{n_1}{k}:\twonorm[v_k]=1} \left( \Ltrb{\NN_{\theta}} + \lw\left(\begin{cases}
   2\rho(\|c\|_2) &\text{, if } w_{n_1}=0\\
   \|c\|_2^2 &\text{, else}
   \end{cases}+2\sum_{k=1}^{n_1}\frac{\twonorm[w_k]}{g(-b_k)}\right)\right)},
\end{equation}}
and
\begin{align}\label{eq:Main_Theorem_Subset_helper}
    \argmin_{f\in\cF_n}
    \left(\Ltrb{f}+\lambda\Pwellemeasure(f)\right),
\end{align}
coincide, where $\rho(r)=\begin{cases}r^2/2&\text{, if }|r|\leq1\\
|r|-1/2&\text{, else}
\end{cases}$ is the Huber-loss, and
\begin{align*}
    \begin{split}
         \cF_n:=\Big\{\,&f\in\cF \,\big|\,f=\link^{-1}\circ h_1,\ \exists\,(\mu,c)   \in\mathfrak{M}_{n_1}(\Sd[d_{0}]\times\Sd[d_{1}-1])\times\R^{d_1}:\\ &h_1:\R^{d_{0}}\to\R^{d_1},\ h_1(x):=\int_{\Sd[d_{0}]\times\Sd[d_{1}-1]}w\,\sigma(\langle v, \cdot\rangle-r)\,d\mu((v,r),w)+c\Big\},
    \end{split}
\end{align*}
and $\mathfrak{M}_{n_1}(\Sd[d_{0}]\times\Sd[d_{1}-1])
:=%
\set{\mu\in\mathfrak{M}(\Sd[d_{0}]\times\Sd[d_0-1]): \#\supp(\mu)=n_1}$ is the set of Radon-measures only supported on $n_1$ points.
\end{lemma}
\begin{proof}
For a simpler notation we have formulated the statement for $j=1$, but the proof works analogously for other every stack $j$, therefore we write $j$ instead of 1 within this proof. By the positive homogeneity of the ReLU, for all $x\in\Rdin$ we have
\begin{align*}
    \NN_\theta(x)&=
	\sum_{k=1}^{n_j}w_k\,\relu[b_{k}+\langle v_{k}, x\rangle] + c=	\sum_{k=1}^{n_j}\widetilde{w_k}\,\relu[\widetilde{b_{k}}+\langle \widetilde{v_{k}}, x\rangle] + c,
\end{align*}
where $\widetilde{w_k}:=w_k\sqrt{\frac{\twonorm[(v_k,b_k)]}{\twonorm[w_k]}}$, $\widetilde{v_k}:=v_k\sqrt{\frac{\twonorm[w_k]}{\twonorm[(v_k,b_k)]}}$ and $\widetilde{b_k}:=b_k\sqrt{\frac{\twonorm[w_k]}{\twonorm[(v_k,b_k)]}}$. We define $\widetilde{\theta}:=(\widetilde{v},\widetilde{b},\widetilde{w},c)$. Then,
\begin{enumerate}
    \item \begin{align*}
    \NN_\theta=\NN_{\widetilde{\theta}} \text{ and thus } \Ltr(\NN_\theta)=\Ltr(\NN_{\widetilde{\theta}}),
\end{align*}
\item with the inequality between geometric and arithmetic mean for every $\theta$
\begin{align*}
    \frac{1}{2}\twonorm[{\theta}]^2&=\frac{1}{2}\left(	\sum_{k=1}^{n_j} \twonorm[{w_k}]^2+\twonorm[(v_k,b_k)]^2+\twonorm[c]^2\right)\\
    &\ge\left(\sum_{k=1}^{n_j}\twonorm[{w_k}]\twonorm[(v_k,b_k)]\right)+\frac{1}{2}\twonorm[c]^2 
\end{align*}
\item and equality holds for $\widetilde{\theta}$
\begin{align*}
    \frac{1}{2}\twonorm[\widetilde{\theta}]^2&=\frac{1}{2}\left(	\sum_{k=1}^{n_j} \twonorm[\widetilde{w_k}]^2+\twonorm[(\widetilde{v_k},\widetilde{b_k})]^2+\twonorm[c]^2\right)\\
    &=\frac{1}{2}\left(\sum_{k=1}^{n_j} \twonorm[{w_k}]\twonorm[({v_k},{b_k})]+\twonorm[{w_k}]\twonorm[({v_k},{b_k})]\right)+\frac{1}{2}\twonorm[c]^2\\
    &=\sum_{k=1}^{n_j} \twonorm[{w_k}]\twonorm[({v_k},{b_k})]+\frac{1}{2}\twonorm[c]^2\\
    &=\sum_{k=1}^{n_j} \twonorm[\widetilde{w_k}]\twonorm[(\widetilde{v_k},\widetilde{b_k})]+\frac{1}{2}\twonorm[c]^2\\
    &=\sum_{k=1}^{k^*} \twonorm[\widetilde{w_k}]\twonorm[\widetilde{v_k}]
    \underbrace{\sqrt{1+\left(\frac{\widetilde{b_k}}{\twonorm[\widetilde{v_k}]}\right)^2}}_{\frac{1}{g\left(-\frac{\widetilde{b_k}}{\twonorm[\widetilde{v_k}]}\right)}}
    +\sum_{k=k^*+1}^{n_j} \twonorm[\widetilde{w_k}]
    |\widetilde{b_k}|
    +\frac{1}{2}\twonorm[c]^2,
    \end{align*}
    where we assume {w.l.o.g.}\footnote{We can make this assumption be without loss of generality, since the order of the neurons does not matter.}~that there exists an index $k^*\in\fromto{n_j}$, such that $v_k\neq0$ for all $k\leq k^*$ and $v_k=0$ for all $k\geq k^*+1$. 
    Thus, if $\theta$ is optimal with respect to \meqref{eq:l2reg}, $\widetilde{\theta}$ is optimal with respect to \meqref{eq:l2reg} as well and $\NN_\theta=\NN_{\widetilde{\theta}}$.
    \item Then $\widetilde{\widetilde{\theta}}=(\widetilde{\widetilde{v}},\widetilde{\widetilde{b}},\widetilde{\widetilde{w}},c)$ with $\widetilde{\widetilde{w_k}}:=\widetilde{w_k}\twonorm[(\widetilde{v_k},\widetilde{b_k})]$, $\widetilde{\widetilde{v_k}}:=\widetilde{v_k}\frac{1}{\twonorm[(\widetilde{v_k},\widetilde{b_k})]}$ and $\widetilde{\widetilde{b_k}}:=\widetilde{b_k}\frac{1}{\twonorm[(\widetilde{v_k},\widetilde{b_k})]}$.
    Then $\widetilde{\widetilde{\theta}}$
    is optimal with respect to \meqref{eq:l1reg} and  $\NN_\theta=\NN_{\widetilde{\widetilde{\theta}}}$.
    Thus, \meqref{eq:l2reg}$\subseteq$\meqref{eq:l1reg}.
    Analogously, one can prove that \meqref{eq:l2reg}$\supseteq$\meqref{eq:l1reg}.
    
    \item Analogously, one can show \meqref{eq:l2reg}$=$\meqref{eq:l1regg} by setting
    $\widetilde{\widetilde{\widetilde{\theta}}}=(\widetilde{\widetilde{\widetilde{v}}},\widetilde{\widetilde{\widetilde{b}}},\widetilde{\widetilde{\widetilde{w}}},\widetilde{\widetilde{\widetilde{c}}})$
    with
    $\widetilde{\widetilde{\widetilde{c}}}:=c+\sum_{k=k^*+1}^{n_j} \widetilde{w_k}
    |\widetilde{b_k}|$
    and
    $\forall k \in\fromto[1]{k^*}:$
    $\widetilde{\widetilde{\widetilde{w_k}}}:=\widetilde{\widetilde{w_k}}g\left(-\frac{\widetilde{b_k}}{\twonorm[\widetilde{v_k}]}\right)$,
    $\widetilde{\widetilde{\widetilde{v_k}}}:=\widetilde{\widetilde{v_k}}\frac{1}{g\left(-\frac{\widetilde{b_k}}{\twonorm[\widetilde{v_k}]}\right)}=\widetilde{v_k}\frac{1}{\twonorm[\widetilde{v_k}]}$ and
    $\widetilde{\widetilde{\widetilde{b_k}}}:=\widetilde{\widetilde{b_k}}\frac{1}{g\left(-\frac{\widetilde{b_k}}{\twonorm[\widetilde{v_k}]}\right)}\unimportant{=\widetilde{b_k}\frac{1}{\twonorm[\widetilde{v_k}]}}$
    and $\forall k\in \fromto[k^*+1]{n_j}:$
    $\widetilde{\widetilde{\widetilde{w_k}}}:=0$,
    $\widetilde{\widetilde{\widetilde{v_k}}}:=0$
    and
    $\widetilde{\widetilde{\widetilde{b_k}}}:=0$.
    Then $\widetilde{\widetilde{\widetilde{\theta}}}$ is optimal with respect to \meqref{eq:l1regg} and  $\NN_\theta=\NN_{\widetilde{\widetilde{\widetilde{\theta}}}}$.
    Thus, \meqref{eq:l2reg}$\subseteq$\meqref{eq:l1regg}.
    Analogously, one can prove that \meqref{eq:l2reg}$\supseteq$\meqref{eq:l1regg}.
    
    \item \meqref{eq:l1reg}$=$\meqref{eq:Main_Theorem_Subset_helper}, since we can translate parameters $\widetilde{\widetilde{\theta}}$ into measures $\mu:=\sum_{k=1}^{n_j}\delta_{((\widetilde{\widetilde{v_k}},-\widetilde{\widetilde{b_k}}),\widetilde{\widetilde{w_k}})}\in\mathfrak{M}_{n_j}(\Sd[d_{j-1}]\times\Sd[d_{j}-1])$ and vice versa.

\end{enumerate}

Consequently, the sets \meqref{eq:l2reg}, \meqref{eq:l1reg}, \meqref{eq:l1regg} and \meqref{eq:Main_Theorem_Subset_helper} are equal. 

\end{proof}

\Cref{le:l2equivl1g} has only shown the equivalence for NNs with a finite number of neurons.
The remainder of this \cref{subsec:ProofMainTheorem} will prove the infinite width limit. This proof will rely on \Cref{def:Pwellemeasure,le:PsecondderivequivPwellemeasure,le:DeepRequiredNumberOfNeurons}.

\subsubsection{Proof of \eqref{eq:Main_Theorem_Subset}}

\begin{lemma}\label{le:mainThrmFirstStatement}
It holds that for a sufficiently large\footnote{See \cref{sec:RequiredNumberofNeurons} for explicit bounds how many neurons are sufficient.} number of neurons $n$ every solution $\NN_{\theta^{*,\lw}}$ with
\begin{equation}\label{eq:optimalParametersAppendix}
    \theta^{*,\lw}\in\argmin_\theta \left( \Ltrb{\NN_{\theta}} + \lw\twonorm[\theta]^2\right),
\end{equation}

satisfies
\begin{align}\label{eq:Main_Theorem_SubsetAppendix}
    \NN_{\theta^{*\lw}} \in \argmin_{f\in \cF} \left(\Ltrb{f}+\lambda \Pfunc(f)\right).
\end{align}
\end{lemma}
\begin{proof}
We will first prove the statement for $\Pwellemeasure$ from \Cref{def:Pwellemeasure}.
By the equivalence of \eqref{eq:l2reg} and \eqref{eq:Main_Theorem_Subset_helper} in \Cref{le:l2equivl1g} for every stack we obtain, that 
the neural network $\NN_{\theta^{*,\lw}}$ has optimal parameters according to \meqref{eq:optimalParametersAppendix} if and only if
\begin{align}\label{eq:Main_Theorem_Subset_helper_deep}
    \NN_{\theta^{*\lw}} {\in}
    \argmin_{f\in\cF_n}
    \left(\Ltrb{f}+\lambda\Pwellemeasure(f)\right),
\end{align}
where
\begin{align}
    \begin{split}
         \cF_n:=\Big\{\,&f\in\cF \,\big|\,f=\link^{-1}\circ h_{\nStacks}\circ\dots\circ \sigb\circ h_1,\\&\forall\,j\in\{1,\dots,\nStacks\}:\ \exists\,(\mu_j,c_j)   \in\mathfrak{M}_{n_j}(\Sd[d_{j-1}]\times\Sd[d_{j}-1])\times\R^{d_j}:\\ &h_j:\R^{d_{j-1}}\to\R^{d_j},\ h_j(x):=\int_{\Sd[d_{j-1}]\times\Sd[d_{j}-1]}w\,\sigma(\langle v, \cdot\rangle-r)\,d\mu_j((v,r),w)+c_j\Big\}
    \end{split}
\end{align}
and $\mathfrak{M}_{n_j}(\Sd[d_{j-1}]\times\Sd[d_{j}-1])
:=%
\set{\mu\in\mathfrak{M}(\Sd[d_{j-1}]\times\Sd[d_j-1]): \#\supp(\mu)=n_j}$ is the set of Radon-measures only supported on $n_j$ points.

By \Cref{le:DeepRequiredNumberOfNeurons}, $\argmin_{f\in \cF} \left(\Ltrb{f}+\lambda\Pwellemeasure(f)\right)$ %
contains a function $f$ %
that can be expressed via finite measures $\mu_j\in\mathfrak{M}_{\tilde{n}_j}(\Sd[d_{j-1}-1]\times\Sd[d_j-1])$.
This implies for all $n>\tilde{n}$
\begin{align*}
\min_{f\in\cF_n}
\left(\Ltrb{f}+\lambda\Pwellemeasure(f)\right)
=
\min_{f\in \cF} \left(\Ltrb{f}+\lambda\Pwellemeasure(f)\right),
\end{align*}
since \enquote{$\geq$} obviously holds because removing restrictions cannot increase the minimum and \enquote{$\leq$} again holds because of \Cref{le:DeepRequiredNumberOfNeurons}.
Therefore, \meqref{eq:Main_Theorem_Subset_helper_deep} implies \[\NN_{\theta^{*\lw}} \in \argmin_{f\in \cF} \left(\Ltrb{f}+\lambda \Pwellemeasure(f)\right)
=
\argmin_{f\in \cF} \left(\Ltrb{f}+\lambda\Pfunc(f)\right),\] where the last equality holds because of \Cref{le:PsecondderivequivPwellemeasure}.
\end{proof}

\subsubsection{Proof of \eqref{eq:continousSolutionAlmostSubsetOfDiscrete}}
\begin{lemma}\label{le:mainThrmSecondStatement}
Furthermore, it holds that for every non-empty compact $ K\subset\R^{\din},\ \forall \epsilon\in\Rp:$
\begin{multline}
    \forall f^{*,\lambda} \in \argmin_{f\in \cF} \left(\Ltrb{f}+\lambda \Pfunc(f)\right) : \exists \tilde{n}\in\N^\nStacks:
    \forall n>\tilde{n}:\\
    \exists \theta^{*,\lw}\in\argmin_\theta \left( \Ltrb{\NN_{\theta}} + \lw\twonorm[\theta]^2\right):
    \sup_{x\in K}\|{f^{*,\lambda}(x)-\NN_{\theta^{*,\lw}}}(x)\|_{\infty} <\epsilon.\footnote{We believe this result can be extended to a $W^{1,p}$-Sobolev norm instead of the supremum norm, but the proof would be more involved.}
\end{multline}
\end{lemma}
\begin{proof}%
We prove the statement for $\Pwellemeasure$ from \Cref{def:Pwellemeasure} instead of $\Pfunc$, since \Cref{le:PsecondderivequivPwellemeasure} tells us that $\Pwellemeasure=\Pfunc$.
Let $f^{*,\lambda} \in \argmin_{f\in \cF}\left(\Ltrb{f}+\lambda \Pwellemeasure(f)\right)$.
We define $p^*:=\Pwellemeasure(f^{*,\lambda})$.
Then $f^{*,\lambda}$ is Lipschitz-continuous with Lipschitz-constant at most $C^\text{Lip}\leq \left(\frac{p^*}{2}\right)^\nStacks$. (With a slightly more technical proof one can even show that $C^\text{Lip}\leq \left(\frac{p^*}{2\nStacks}\right)^\nStacks$.)
\newline
Let $\epsilon>0$. Since $K$ is compact, there exists a finite number $N_{K,\delta}\in\N$ of points $\left(\widetilde{\xtr_i}\right)_{i\in\fromto{N_{K,\delta}}}$ such that for every $x\in K$ there exists $\widetilde{\xtr_i}$ with $\twonorm[\widetilde{\xtr_i}-x]<\delta:=\frac{\epsilon}{2C^\text{Lip}}$.
\Cref{le:DeepRequiredNumberOfNeurons} tells us that there exists a network~$\NN_{\theta}$ with (non-bottleneck) widths $\widetilde{n}$, such that $\Ltrb{\NN_{\theta}}=\Ltrb{f^{*,\lambda}}$, $\NN_{\theta}(\widetilde{\xtr_i})=f^{*,\lambda}(\widetilde{\xtr_i})$ for all $i\in\fromto{N_{K,\delta}}$ and $\Pwellemeasure(\NN_{\theta})=\Pwellemeasure(f^{*,\lambda})$. Thus, also $\NN_{\theta}$ has a Lipschitz-constant of at most $C^\text{Lip}\leq \left(\frac{p^*}{2}\right)^\nStacks$.
This implies that $\NN_{\theta}-f^{*,\lambda}$ has a Lipschitz-constant of at most $2C^\text{Lip}$.
Consequently, $\sup_{x\in K}\|\NN_{\theta}-f^{*,\lambda}\|_\infty<\eps$.
\end{proof}

\subsection{Proof of \Cref{cor:wide_dj}}\label{subsec:ProofOfCorollary}
We prove %
a similar statement in \Cref{cor:wide_dj_restated}. Moreover, we give further viewpoints on \Cref{cor:wide_dj,cor:wide_dj_restated} in \Cref{cor:wide_dj_additional_views}. Then we \hyperlink{proof:cor:wide_dj}{prove} \Cref{cor:wide_dj}. In the remainder of \Cref{subsec:ProofOfCorollary} we state and prove the lemmas needed for the proof of \Cref{cor:wide_dj_restated}. %

\begin{corollary}[\Cref{cor:wide_dj} restated]\label{cor:wide_dj_restated}
Let $\sigb$ be ReLU or linear. For every number of training data points $N$, there exist $n^*\in\N^{\nStacks}$ and $d^*\in\N^{\nStacks-1}$, such that for every 
solution 
$
    \theta^{*,\lw}\in\argmin_{\theta\in\Thetandstar} \left( \Ltrb{\NN_{\theta}} + \lw\twonorm[\theta]^2\right)
$
with network-dimensions $n^*$ and $d^*$, the corresponding neural network $\NN_{\theta^{*,\lw}}$
is also a solution of
$%
\argmin_{f\in \cF} \left(\Ltrb{f}+\lambda\Pfunc_{(d)}(f)\right)$ for all bottleneck dimensions $d\geq d^*$.
I.e.,
\begin{multline*}
\forall N\in\N:\exists n^*\in\N^{\nStacks},d^*\in\N^{\nStacks-1}:\forall d>d^*:\\
\forall \theta^{*,\lw}\in\argmin_{\theta\in\Thetandstar} \left( \Ltrb{\NN_{\theta}} + \lw\twonorm[\theta]^2\right):
\NN_{\theta^{*,\lw}}\in\argmin_{f\in \cF} \left(\Ltrb{f}+\lambda\Pfunc_{(d)}(f)\right),
\end{multline*}
where $\NN_{\theta^{*,\lw}}$ has network dimensions $n^*$ and $d^*$ and $P_{(d)}$ has bottleneck dimensions $d$.
\end{corollary}
\begin{proof} According to \Cref{cor:ExistenceMinimizerF}, there exists a minimizer $f^*\in\argmin_{f\in \cF} \left(\Ltrb{f}+\lambda\Pfunc(f)\right)$.
By \Cref{le:PsecondderivequivPwellemeasure}, $f^*$ is also a solution of \[\argmin_{f\in \cF} \left(\Ltrb{f}+\lambda\Pwellemeasure(f)\right).\] Therefore, $f^*$ admits a representation  
\[f=\ell^{-1}\circ h_{\nStacks}\circ\tilde{\sigma}\circ\ldots\circ \tilde{\sigma}\circ h_{1}\]
with stacks
\[\forall x\in\R^{d_{j-1}}: h_{j}(x)=c_j+\int_{\Sd[d_{j-1}]\times\Sd[d_{j}-1]} w\sigma(\langle v, x\rangle-r)d\mu_j((v,r),w),\]
with $c_j\in\R^{d_j}$ and measures
$\mu_j\in\mathfrak{M}(\Om_j)$ and $\Om_j:=\Sd[d_{j-1}]\times\Sd[d_j]$.

Then, by \Cref{le:DeepRequiredNumberOfNeurons}, there exists a NN $\check{f}$ whose corresponding measures $\check{\mu}_j$ are supported on a finite number of nodes $n^*:=(\widetilde{n}_1,\ldots,\widetilde{n}_{\nStacks})$ and $d^*:=(\widetilde{d}_1,\dots,\widetilde{d}_{\nStacks-1})$, s.t. we have
\begin{align}
    \Ltrb{f}&=\Ltrb{\check{f}}\\
   c_j=\check{c}_j \text{ and } \mu_j(\Om_j)&=\check{\mu}_j(\check{\Om}_j),\quad\forall j=1,\ldots,\nStacks.
\end{align}
The latter implies that $\Pwellemeasure(f)=\Pwellemeasure(\check{f})$. Thus in total, \[\check{f}\in\argmin_{f\in \cF} \left(\Ltrb{f}+\lambda\Pwellemeasure(f)\right).\] By the equivalence of \meqref{eq:l2reg} and \meqref{eq:Main_Theorem_Subset_helper} in \Cref{le:l2equivl1g}, there exists $\theta^*\in\argmin_{\theta} \left(\Ltrb{\NN_\theta}+\lambda\twonorm[\theta]^2\right)$ s.t. $\check{f}=\NN_{\theta^*}$.
\end{proof}

\begin{corollary}\label{cor:wide_dj_additional_views}
Let $\sigb$ be ReLU or linear. For every number of training data points $N$, there exist $n^*\in\N^{\nStacks}$ and $d^*\in\N^{\nStacks-1}$,%
\textsuperscript{\ref{footnote:boundsonneurons}} such that for all bottleneck dimensions $d \geq d^*$,\footnote{Note that $\NN_{\theta}$ depends on $d$ and $n$ since it refers to a network with $d_j$ and $n_j$ neurons in the corresponding layers. Moreover, $d\geq d^*$ is always understood component-wise, i.e., $\forall j\in\{1,\dots, \nStacks-1\}:d_j\geq d_j^*$, while $d_0=\din$ and $d_\nStacks=\dout$ are considered constant throughout the paper.}
\begin{enumerate}
\item and for every $n\geq n^*$, 
\begin{subequations}
\begin{align}&\min_{\theta\in\Thetandstar} \left( \Ltrb{\NN_{\theta}} + \lw\twonorm[\theta]^2\right)\\
&=
\min_{\theta\in\Thetand} \left( \Ltrb{\NN_{\theta}} + \lw\twonorm[\theta]^2\right)\\
&=
\min_{f\in \cF} \left(\Ltrb{f}+\lambda\Pfunc_{(d^*)}(f)\right)
\\
&=
\min_{f\in \cF} \left(\Ltrb{f}+\lambda\Pfunc_{(d)}(f)\right).
\end{align}
\end{subequations}
    \item and for every $n\geq n^*$ 
   there exists a solution $\NN_{\theta^{*,\lw}}$ with
\[
    \theta^{*,\lw}\in\argmin_{\theta\in\Thetand} \left( \Ltrb{\NN_{\theta}} + \lw\twonorm[\theta]^2\right)
\]
and network-dimensions $n$ and $d$,
such that the function $\NN_{\theta^{*,\lw}}$ can also be represented by a network with dimensions $n^*$ and $d^*$.
     \item  
there exists a solution \[f^{*,\lw}\in\argmin_{f\in \cF} \left(\Ltrb{f}+\lambda\Pfunc_{(d)}(f)\right)\]
(where $\Pfunc_{(d)}$ denotes $\Pfunc$ with bottleneck-dimensions $d_j$)\footnote{$\Pfunc$ always depends on the bottleneck-dimensions $d$, but by writing $\Pfunc_{(d)}$ we make the dependency on $d$ more explicit at this point.},
s.t. there exist parameters \[
    \theta^{*,\lw}\in\argmin_{\theta\in\Thetandstar} \left( \Ltrb{\NN_{\theta}} + \lw\twonorm[\theta]^2\right)
\] with dimensions $n^*$ and $d^*$ such that $\NN_{\theta^{*,\lw}}=f^{*,\lw}$.

\item\label{itm:nestedNetworkArgmins} and for every $\tilde{n}\geq n \geq n^*$ and for every $\tilde{d} \geq d\geq d^*$ the sets of solutions are nested:
\begin{align*}
\emptyset&\neq%
\left\{\NN_{\theta^{*,\lw}}:\argmin_{\theta\in\Thetand} \left( \Ltrb{\NN_{\theta}} + \lw\twonorm[\theta]^2\right)\right\}\\
&\subseteq
\left\{\NN_{\theta^{*,\lw}}:\argmin_{\theta\in\ThetandGeneral{\tilde{n}}{\tilde{d}}} \left( \Ltrb{\NN_{\theta}} + \lw\twonorm[\theta]^2\right)\right\}.
\end{align*}

\item\label{itm:nestedPfuncArgmins} and for every $\tilde{d} \geq d\geq d^*$ and for every solution \[
    \theta^{*,\lw}\in\argmin_{\theta\in\Thetandstar} \left( \Ltrb{\NN_{\theta}} + \lw\twonorm[\theta]^2\right)%
\] it holds that the sets of solutions are nested:
\begin{multline*}
\NN_{\theta^{*,\lw}}
\in
\argmin_{f\in \cF} \left(\Ltrb{f}+\lambda\Pfunc_{(d^*)}(f)\right)\\
\subseteq
\argmin_{f\in \cF} \left(\Ltrb{f}+\lambda\Pfunc_{(d)}(f)\right)
\subseteq
\argmin_{f\in \cF} \left(\Ltrb{f}+\lambda\Pfunc_{(\tilde{d})}(f)\right).
\end{multline*}

\end{enumerate}

\end{corollary}
\begin{proof}
These statements follow quite directly from \Cref{thm:PFunc,cor:wide_dj_restated}.
\end{proof}

\hypertarget{proof:cor:wide_dj}{}\begin{proof}\textbf{of \Cref{cor:wide_dj}.}
\Cref{itm:nestedNetworkArgmins,itm:nestedPfuncArgmins} in \Cref{cor:wide_dj_additional_views} show that \eqref{subeq:limitAllLayersWideNetworks}$=$\eqref{subeq:unionAllLayersWideNetworks} and \eqref{subeq:limitAllLayersWidePfunc}$=$\eqref{subeq:unionAllLayersWidePfunc} (where both equalities would even hold without the closure).

\Cref{eq:Main_Theorem_Subset} in \Cref{thm:PFunc} implies that \eqref{subeq:unionAllLayersWideNetworks}$\subseteq$\eqref{subeq:unionAllLayersWidePfunc} (without the closure).
\Cref{eq:continousSolutionAlmostSubsetOfDiscrete} in \Cref{thm:PFunc} implies that \eqref{subeq:unionAllLayersWideNetworks}$\supseteq$\eqref{subeq:unionAllLayersWidePfunc}, because of the closure. Thus, \eqref{subeq:unionAllLayersWideNetworks}$=$\eqref{subeq:unionAllLayersWidePfunc}.

Therefore, \eqref{subeq:limitAllLayersWideNetworks}$=$\eqref{subeq:unionAllLayersWideNetworks}$=$ \eqref{subeq:limitAllLayersWidePfunc}$=$\eqref{subeq:unionAllLayersWidePfunc} holds.

The second part of \Cref{cor:wide_dj} follows directly from \Cref{cor:wide_dj_restated}.
\end{proof}

\begin{lemma}\label{le:OneStackNumberOfNeurons}
We define $\Om:=\Sd[\din]\times\Sd[\dout-1]$ and
let $\mu\in\mathfrak{M}(\Om)$ be a Radon-measure. Based on this measure, we define a function $f$ as
\[\forall x\in\R^\din: f(x)=\int_{\Sd[\din]\times\Sd[\dout-1]} w\sigma(\langle v, x\rangle-r)d\mu((v,r),w)\]

Let $I_1\dot{\cup}\dots\dot{\cup}I_\dout=\fromto{N}$.
Then there exists a Radon measure $\check{\mu}\in\mathfrak{M}_{N+1}(\Sd[\din]\times\Sd[\dout-1])$, such that
\[\forall k\in\fromto{\dout}:
\forall i\in I_k:
f_k(\xtr_i)=\check{f}_k(\xtr_i),
\]
where $\check{f}(x)=\int_{\Sd[\din]\times\Sd[\dout-1]} w\sigma(\langle v, x\rangle-r)d\check{\mu}((v,r),w)$ and
\[\mu(\Om)=\check{\mu}(\Om).\]
\end{lemma}
\begin{proof}

\begin{align}
    \forall x\in\R^\din: f(x)&=\mu(\Om)\int_\Om w\, \relu[\langle v, x\rangle-r]\, d\widetilde{\mu}((v,r),w)
\end{align}
for a probability measure $\widetilde{\mu}=\frac{\mu}{\mu(\Om)}\in\mathcal{P}(\Om)$.

We define $k(i)=k\text{, such that }i\in I_k$.
By a change of measure $\widehat{\mu}:=\widetilde{{\mu}}_{\#}\beta \in\mathcal{P}(C)$ with \[\beta: (v,r,w) \mapsto \left(w_{k(i)}\,\relu[\langle v, \xtr_i\rangle-r]\right)_{i\in\fromto{N}}\] and $C:=\beta(\Om)$, we finally obtain
\begin{align}\label{eq:betaPushForward}
   (f_{k(i)}(\xtr_i))_{i\in\fromto{N}}=\mu(\Om)\int_\Om \beta(v,r,w) \,d\widetilde{{\mu}}((v,r),w) = \mu(\Om)\int_C z\, d\widehat{\mu}(z).
\end{align}

Given that $C \subseteq \mathbb{R}^{N}$ is bounded and $supp(\widehat{\mu})\subseteq C$, we obtain by \cite[Lemma 2 on p. 557]{Rossetl1finite10.1007/978-3-540-72927-3_39} that $\int_C z\, d\widehat{\mu}(z) \, \in \, conv\left(C\right)\,$ where $conv\left(C\right)$ is the convex hull of $C$. Then by \textit{Caratheodory’s Convex Hull Theorem}, the integral $\int_C z\, d\widehat{\mu}(z)$ can be expressed as a convex combination of at most $N+1$ points $c_j\in C$, i.e., $\int_C z\, d\widehat{\mu}(z)=\sum_{j=1}^{N+1}t_jc_j$ (a shallow network of finite width).
This convex combination of $N+1$ points defines a measure $\check{\mu}:=\mu(\Om)\sum_{j=1}^{N+1}t_j\delta_{\om_j}$ as the convex combination of $N+1$ Dirac-distributions at points $\om_j\in\beta^{-1}(c_j)$. Thus, we obtain,
\begin{align*}
    \check{f}_{k(i)}(\xtr_i)
    &=\int_\Om w_{k(i)}\sigma(\langle v, \xtr_i\rangle-r)d\check{\mu}((v,r),w)
    =\int_\Om \beta_i(v,r,w) \,d\check{{\mu}}((v,r),w)
    \\
    &=\mu(\Om)\sum_{j=1}^{N+1}t_j\beta_i(\om_j)
    =\mu(\Om)\sum_{j=1}^{N+1}t_j (c_j)_i
    =\left(\mu(\Om)\sum_{j=1}^{N+1}t_j c_j\right)_i
    \\
    &=\left(\mu(\Om)\int_C z\, d\widehat{\mu}(z)\right)_i
    \overset{\text{\eqref{eq:betaPushForward}}}{=} f_{k(i)}(\xtr_i).
\end{align*}
\end{proof}

\begin{lemma}\label{le:DeepRequiredNumberOfNeurons}
We define for $j=1,\ldots,\#$stacks, the sets $\Om_j:=\Sd[d_{j-1}]\times\Sd[d_j-1]$ and
let $\mu_j\in\mathfrak{M}(\Om_j)$ be Radon-measures.
Let $f:\R^{\din}\to\R^{\dout}$ be a function that can then be written as
\[f=\ell^{-1}\circ h_{\nStacks}\circ\tilde{\sigma}\circ\ldots\circ \tilde{\sigma}\circ h_{1}\]
with stacks
\[\forall x\in\R^{d_{j-1}}: h_{j}(x)=\int_{\Sd[d_{j-1}]\times\Sd[d_{j}-1]} w\sigma(\langle v, x\rangle-r)d\mu_j((v,r),w).\]
Let $\tilde{\sigma}$ be ReLU or linear and $I_1\dot{\cup}\dots\dot{\cup}I_\dout=\fromto{N}$.

\begin{enumerate}
    \item Assume that the dimensions of bottleneck layers $d_j$ are kept fixed. Then there exist Radon measures $\check{\mu}_j\in\mathfrak{M}_{n_j}(\Sd[d_{j-1}]\times\Sd[d_{j}-1])$ with finite support on $n_j$ points, such that
\[\forall k\in\fromto{\dout}:
\forall i\in I_k:
f_k(\xtr_i)=\check{f}_k(\xtr_i),\footnote{\label{footnote:lossissame} This implies $\Ltrb{f}=\Ltrb{\check{f}}$.}
\]
where 
\[\check{f}=\ell^{-1}\circ \check{h}_{\nStacks}\circ\tilde{\sigma}\circ\ldots\circ \tilde{\sigma}\circ \check{h}_{1}\]
with stacks
$\check{h}_j(x)=\int_{\Sd[d_{j-1}]\times\Sd[d_{j}-1]} w\sigma(\langle v, x\rangle-r)d\check{\mu}_j((v,r),w)$ and
\[\mu_j(\Om)=\check{\mu}_j(\Om)\quad \forall j=1,\ldots,\nStacks.\] 
The number of required nodes is $n_j=Nd_{j}+1$ for $j=1,\ldots,\nStacks-1$ and $n_{\nStacks}=N+1$ for the terminal stack.
    \item 
    \begin{itemize}
        \item[\textbf{a)}] For $\widetilde{\sigma}=id$, let
        $\widetilde{d}_j= \sum_{k=0}^{\nStacks - j} N^{k}$ for $j=1,\ldots,\nStacks$, and
        \item[\textbf{b)}] for $\widetilde{\sigma}=ReLU$, let 
        $\widetilde{d}_j=\sum_{k=0}^{2(\nStacks - j)} N^{k}$, for $j=1,\ldots,\nStacks-1$.
    \end{itemize}
    Then, for every $d\geq \widetilde{d}$ and $f$ as defined above (i.e., $h_j:\R^{d_{j-1}}\to\R^{d_{j}}$),
    there exist $\check{\mu}_j\in\mathfrak{M}_{\widetilde{n}_j}(\Sd[\widetilde{d}_{j-1}]\times\Sd[\widetilde{d}_{j}-1])$ with finite support on $\widetilde{n}_j$ points, such that
\[\forall k\in\fromto{\dout}:
\forall i\in I_k:
f_k(\xtr_i)=\check{f}_k(\xtr_i),\textsuperscript{\ref{footnote:lossissame}}
\]
where, 
    \[\check{f}=\ell^{-1}\circ \check{h}_{\nStacks}\circ\tilde{\sigma}\circ\ldots\circ \tilde{\sigma}\circ \check{h}_{1}\]
    with stacks
$\check{h}_j(x)=\int_{\Sd[\widetilde{d}_{j-1}]\times\Sd[\widetilde{d}_{j}-1]} w\sigma(\langle v, x\rangle-r)d\check{\mu}_j((v,r),w)$ and
\[\mu_j(\Om)=\check{\mu}_j(\Om)\quad \forall j=1,\ldots,\nStacks.\] 
    
    The number of required nodes in these cases are 
    \begin{itemize}
        \item[\textbf{a)}]  $\widetilde{n}_j= \sum_{k=0}^{\nStacks - j +1} N^{k}$ for $j=1,\ldots,\nStacks$, and
        \item[\textbf{b)}] $\widetilde{n}_j=\sum_{k=0}^{2(\nStacks - j)+1} N^{k}$, $j=1,\ldots,\nStacks$.
    \end{itemize}
\end{enumerate}
\end{lemma}
\begin{proof}
We show that $\check{\mu}_j$ satisfy the even stronger condition:
\[\forall i\in\fromto{N}:
h_j(\xtr_i)=\check{h}_j(\xtr_i)\in\R^{d_j}.
\]
\begin{enumerate}
    \item In the case of a deep stacked neural network with fixed bottleneck dimensions $d_j$ for $j=1,\dots,\nStacks-1$, by \Cref{le:OneStackNumberOfNeurons}, $n_j=N{d_{j}}+1$ is an upper bound for the number of neurons\footnote{By the equivalence of \eqref{eq:l2reg} and \eqref{eq:Main_Theorem_Subset_helper} in \Cref{le:l2equivl1g}, we know every measure $\check{\mu}_j\in\mathfrak{M}_{\widetilde{n}_j}(\Sd[\widetilde{d}_{j-1}]\times\Sd[\widetilde{d}_{j}-1])$ corresponds to a stack $\NNj{j}$ with $\widetilde{n}_j$ hidden neurons.} in the hidden layer of stack $j$ for $j=1,\dots,\nStacks$.
    
    \item
Next, we assume the bottleneck dimensions not to be fixed and $\widetilde{\sigma}$ to be the identity function. Again, by \Cref{le:OneStackNumberOfNeurons}, an upper bound for the number of neurons of the last hidden layer is $\widetilde{n}_{\nStacks}:=N+1$. Additionally, \Cref{le:OneStackNumberOfNeurons} applied to the second-to-last stack gives a representation of the $\nStacks-1$-th stack by a measure with finite support on $n_{\nStacks-1}=Nd_{\nStacks-1}+1$, s.t. the mapping from the hidden layer of the second-to-last to the last hidden layer is $M:=V^{\nStacks}W^{\nStacks-1} \in \mathbb{R}^{\widetilde{n}_{\nStacks}\times n_{\nStacks-1}}$ where
\begin{align}
    W^{\nStacks-1}&:=
    \begin{pmatrix}
    w^{(\nStacks-1)}_1
    \hdots\
    w^{(\nStacks-1)}_{n_{\nStacks-1}}
  \end{pmatrix}^\intercal \in \mathbb{R}^{d_{\nStacks-1}\times n_{\nStacks-1}}\\
   V^{\nStacks}&:=
    \begin{pmatrix}
    v^{(\nStacks)}_1
    \hdots\
    v^{(\nStacks)}_{n_{\nStacks}}
  \end{pmatrix}^\intercal \in \mathbb{R}^{\widetilde{n}_{\nStacks}\times d_{\nStacks-1}}\, .
\end{align}\\
Now, knowing that we can write $M$ as its single value decomposition, i.e., $M=U\Sigma V$ with $p:=min\{\widetilde{n}_{\nStacks},n_{\nStacks-1}\}$ being the number of singular values, we define $A:=\sqrt{\Sigma_p}V_{p_1}\in\mathbb{R}^{p\times n_{\nStacks-1}}$ and $B:=U_{p_2}\sqrt{\Sigma_p}\in \mathbb{R}^{\widetilde{n}_{\nStacks}\times p}$ where the subscripts $p$, $p_1$ and $p_2$ refer to the upper left submatrix with dimension $\mathbb{R}^{p\times p}$, the submatrix with all rows and the submatrix with all columns removed after dimension $p$, respectively, such that we get the additional decomposition $M=BA$. This allows us to limit the maximum number of necessary neurons of the last \emph{bottleneck layer} to $\widetilde{d}_{\nStacks-1}:=N+1$ (since $\widetilde{n}_{\nStacks}=N+1<Nd_{\nStacks-1}+1=n_{\nStacks-1}$) and of the second-to-last hidden layer, following again from \Cref{le:OneStackNumberOfNeurons}, to $\widetilde{n}_{\nStacks-1}:=N(\widetilde{d}_{\nStacks-1})+1=N(N+1)+1$. Inductively, an upper limit of necessary neurons of the j-th bottleneck layer can be set at $\widetilde{d}_j=\widetilde{n}_{j+1}$ and for the j-th hidden layer at $\widetilde{n}_j= \sum_{k=0}^{\nStacks +1 - j} N^{k}$.\\ \\
In case of $\widetilde{\sigma}$ being ReLU, the architecture reduces to a regular deep neural network, i.e., respective upper bounds for the number of required neurons can be set at $\widetilde{d}_j=\sum_{k=0}^{2(\nStacks - j)} N^{k}, j=1,\ldots,\nStacks-1$ and $\widetilde{n}_j=\sum_{k=0}^{2(\nStacks - j)+1} N^{k}, j=1,\ldots,\nStacks$. 

\end{enumerate}

\end{proof}
\clearpage{}%

\end{document}